\pdfoutput=1
\documentclass[11pt]{article}
\usepackage{hyphenat}
\usepackage{CJKutf8}
\newcommand{\ko}[1]{\begin{CJK}{UTF8}{mj}#1\end{CJK}}
\newcommand{\zh}[1]{\begin{CJK}{UTF8}{gbsn}#1\end{CJK}}
\usepackage[final]{acl}
\usepackage{times}
\usepackage{xltabular}
\usepackage[T1, T2A]{fontenc}
\usepackage[russian, english]{babel}
\usepackage{booktabs}
\usepackage{longtable}
\usepackage{latexsym}
\usepackage{makecell}
\usepackage{stfloats}
\usepackage{amsmath}
\usepackage{amssymb}
\usepackage{amsfonts}
\usepackage{multirow}
\usepackage{adjustbox}
\usepackage{pifont}
\usepackage{silence}
\usepackage{hyperref}
\usepackage{subcaption}
\usepackage[ruled,vlined,linesnumbered]{algorithm2e}
\usepackage{tcolorbox}
\usepackage{colortbl}
\usepackage{xcolor}
\usepackage[table]{xcolor}
\usepackage{arydshln}
\usepackage{dsfont}
\usepackage[utf8]{inputenc}
\DeclareUnicodeCharacter{2212}{-}

\definecolor{rowgray}{gray}{0.75} 
\definecolor{gold}{RGB}{150,130,20}
\definecolor{statusNormal}{HTML}{2E7D32} 
\definecolor{statusTrigger}{HTML}{EF6C00} 
\definecolor{statusBroken}{HTML}{C62828} 
\definecolor{statusRecover}{HTML}{1565C0}

\usepackage{tabularx}
\usepackage{fontawesome5}

\usepackage[T1]{fontenc}

\usepackage[utf8]{inputenc}

\usepackage{microtype}

\usepackage{inconsolata}

\title{\textsc{Doc2FRC}: Length-Consistent Document-Level Machine Translation via Fixed-Range Chunking}

\author{Xiaotian Wang$^{1, 3}$ \ \ \ \ Youyuan Lin$^2$ \ \ \ \ Zhan Shen$^1$ \ \ \ \ Hitomi Yanaka$^{1, 3, 4}$ \\
       $^1$The University of Tokyo \ \ $^2$Kyoto University \ \ $^3$Riken \ \ $^4$ Tohoku University \\
       $^1$\texttt{\{eternaleden0321, shenzhan, hyanaka\}@g.ecc.u-tokyo.ac.jp}\\ 
       $^2$\texttt{lin.youyuan.73v@st.kyoto-u.ac.jp}\\ 
       }

\usepackage{amsmath}
\begin{document}
\maketitle
\begin{abstract}
Advanced large language models (LLMs) with long context windows can substantially reduce input truncation in document-level machine translation (DocMT).
However, direct \textit{Doc2Doc} translation remains prone to n-gram repetition
and progressive quality degradation.
A common remedy is to segment the document into finer-grained chunks.
Nonetheless, conventional rule-based chunking approaches
fail to handle the length distribution mismatch between training and inference.
To address this,
we introduce \textbf{F}ixed-\textbf{R}ange \textbf{C}hunking~(\textbf{FRC}),
utilizing dynamic programming to partition documents into chunks within a predefined length interval.
By consistently applying FRC during training and inference, 
the input documents of any length are mapped to the same length distribution,
substantially reducing train-test length mismatch.
Centered on FRC,
we propose a lightweight dual-boundary matching algorithm for chunk alignment,
alongside four distinct training strategies. 
Experimental results show that FRC-based fine-tuning substantially improves 7B LLMs over direct \textit{Doc2Doc} fine-tuning and outperforms existing DocMT methods on IWSLT2017. 
We further construct \textsc{GlobVDoc}, 
a 10-language test set independent of mainstream DocMT training sources, 
and show that FRC improves out-of-distribution document translation.\footnote{
The code, dataset, and fine-tuned models are publicly available at: 
\url{https://github.com/ynklab/Doc2FRC}.
}
\end{abstract}

\section{Introduction}
\label{sec:intro}
Document-level machine translation (DocMT) has been continuously investigated for a long time~\citep{miculicich-etal-2018-document, voita-etal-2018-context, maruf-haffari-2018-document, lyu-etal-2021-encouraging}. 
Current large language models~(LLMs) with long context windows have greatly alleviated the challenge of input length constraints for DocMT.
However, 
direct \textit{Doc2Doc} translation is still plagued by various issues, 
such as susceptibility to n-gram repetition during decoding~\citep{jin2024chaptertochaptercontextawareliterarytranslation,peng-etal-2025-investigating},
and progressive quality degradation as sequence lengths increase~\citep{yang2025hallucinatelongresponsegeneration}.
Accordingly, dominant strategies such as \textit{Doc2Sent} and \textit{Doc2Chunk} emerged,
which translate documents sequentially at the level of sentences or chunks.
While these methods can also mitigate the challenges associated with variable document input lengths,
the design of optimal training paradigms based on them is still under active exploration.

Regarding the \textit{Doc2Sent} strategy, 
early studies~\citep{wu2024adaptinglargelanguagemodels, li-etal-2024-towards-demonstration} utilized a fixed number of sentence pairs as contexts to facilitate training. 
While this approach can ensure input length alignment between training and inference, 
it remains rooted in sentence-level processing and thus lacks sufficient in-context information.
More recently, \citet{ramos2025multilingual} introduced a chunk-level training paradigm that partitions documents into a fixed number of sub-documents,
and employs a hybrid training strategy incorporating both context-free and context-aware settings,
emphasizing length-agnostic generalization.
However, it cannot guarantee length distributional alignment between training and inference, 
which has been empirically observed to potentially lead to performance degradation~\citep{varis-bojar-2021-sequence, wan-etal-2022-challenges, pitorro-etal-2024-effective}.
While \citet{peng-etal-2025-investigating} proposed a solution using varying position embeddings to improve length extrapolation during training,
they also prioritized enhancing translation capabilities across diverse lengths.

Driven by similar concerns regarding length distribution matching,
instead of improving generalization across diverse lengths,
we focus on achieving chunk-level alignment by mapping both training and inference documents into a unified length interval,
thereby training exclusively on these consistent distributions.
Nonetheless, conventional strategies,
such as fixed-number and fixed-length chunking,
cannot fully reconcile length discrepancies, 
frequently leaving short residual fragments at the end of a document.
To address this limitation,
we propose \textbf{F}ixed-\textbf{R}ange \textbf{C}hunking~(\textbf{FRC}),
which employs a dynamic programming algorithm for document chunking,
ensuring all document segments fall within a target length interval.
Consequently, the length distribution of the training and inference data maintains consistently well-aligned.

In addition to introducing the method of FRC for DocMT,
during the training data curation stage, 
we propose a lightweight chunk alignment algorithm based on dual-boundary matching, 
which can be applied to all document-level parallel corpora that are not fully sentence-aligned. 
Then, we use FRC pairs to construct five different data formats to train four types of models~(as shown in Figure~\ref{fig:train_format}),
enabling a systematic model-wise investigation.

Furthermore, 
aiming to probe out-of-distribution generalization and mitigate the scarcity of long-form evaluation data in the news and social domains, 
we manually curate a high-quality, 
10-language test set named \textsc{GlobVDoc},
based on the Global Voices\footnote{\url{https://globalvoices.org}}.
This dataset is guaranteed to be strictly sentence-aligned
and is designed to remain independent of the data distributions used in current mainstream DocMT training sets.

Our main contributions are as follows:
\begin{itemize}
\vspace{-4pt}
\item

We propose Fixed-Range Chunking and Dual-Boundary Matching-based Chunk Alignment algorithms to ensure length distribution consistency across training and inference with efficient chunk synchronization.

\vspace{-4pt}
\item
We train four types of models
across five training data formats.
Our results show that FRC-based training strategies outperform the \textit{Doc2Doc} baseline. 
On IWSLT2017 test data, 
our method achieves a d-BLEU improvement that exceeds the existing DocMT methods.

\vspace{-4pt}
\item
We present GlobVDoc, 
a high-quality multilingual test dataset specifically designed for document translation,
and establish a multilingual long-form DocMT benchmark.

\vspace{-4pt}
\item
We introduce a variant of the Lexical Translation Consistency Ratio (LTCR) that estimates translation consistency based on the word alignments between the reference and the translation hypothesis.
\end{itemize}

\section{Related Work}
\label{sec:related}

\paragraph{Training-based DocMT}
Following the emergence of various LLM-based sentence-level translation paradigms~\citep{xu2024a, alves2024tower, guo-etal-2024-novel},
research focus has increasingly pivoted toward LLM-based DocMT.
Early investigations primarily focused on \textit{Doc2Sent} methodologies.
\citet{wu2024adaptinglargelanguagemodels} leveraged historical context from previous translations,
whereas \citet{li-etal-2024-towards-demonstration} introduced retrieval-based sentence pairs to provide auxiliary contextual information.
Subsequent research focused on dataset development,
utilizing training to establish performance benchmarks~\citep{wicks-etal-2024-recovering, alabi-etal-2025-afridoc}.
\citet{jin2024chaptertochaptercontextawareliterarytranslation} extended the translation object to the chapter level for literary translation, 
employing direct chapter-to-chapter fine-tuning using only inter-chapter context.
Compared to training exclusively on long document pairs,
\citet{ramos2025multilingual} introduced a training paradigm that segments documents into \{1,2,4\} parts and fine-tuned the model with MRD2D~\citep{sun-etal-2022-rethinking} and CAPT~\citep{wang-etal-2023-document-level} techniques,
achieving performance comparable to large-scale LLMs while ensuring generalized translation ability across diverse input lengths. 
In contrast, our FRC approach focuses on specializing within a targeted length range.

\vspace{-5pt}
\paragraph{Training-free DocMT}
Early DocMT works explored prompting formats~\citep{hendy2023goodgptmodelsmachine, wang-etal-2023-document-level, karpinska-iyyer-2023-large, wu2024adaptinglargelanguagemodels}. 
Subsequent studies enriched prompts with document-derived signals: \citet{liu2025improvingllmbaseddocumentlevelmachine} extracted summarization and entity translation knowledge from documents to construct diverse prompts and reranked candidate translations. \citet{hu-etal-2025-source} investigated DocMT within a multi-turn conversation framework and showed that providing the entire source document as prior context can improve iterative refinement.

Beyond prompting, agentic systems explicitly separate roles to support DocMT:
TransAgent~\citep{Wu2024PerhapsBH} assigns specialized agents (e.g., translator and reviewer) to achieve iterative refinement. 
Sent2Sent++~\citep{guo2025docguidedsent2sentsent2sentagent},
GRAFT~\citep{dutta-etal-2025-graft}, and DELTA~\citep{wang2025delta} utilize translation memory to retrieve contextual cues and perform step-by-step translation without debate. 
Finally, TransGraph~\citep{pham2025discoursegraphguideddocument} constructs a discourse-based knowledge graph from document chunks to guide the translation.

Since this work focuses on FRC as a solution for ensuring distributional consistency between training and inference,
we do not discuss its application to training-free DocMT in detail.

\vspace{-5pt}
\paragraph{Analysis}
\citet{peng-etal-2025-investigating} showed that translation quality degrades when inference-time inputs follow length distributions unseen during training, 
and mitigated this shift by modifying positional embeddings.
Our proposed FRC approach addresses the same length-mismatch issue by directly synchronizing the length distributions between the training and inference phases.
Moreover, while many studies incorporate partial translations as context~\citep{wang-etal-2023-document-level, wu2024adaptinglargelanguagemodels, pham2025discoursegraphguideddocument, ramos2025multilingual}, 
\citet{choudhary-etal-2025-exploring} demonstrate that source-only context can also effectively capture discourse phenomena. 
Therefore, to ensure length regulation in FRC and efficient decoding,
we use source-only context by default (see Appendix~\ref{sec:con_src_vs_mt} for further discussion).

\section{Training Data Curation}
\label{sec:train_curation}
\subsection{Fixed-Range Chunking}
\label{sec:algo1}
We propose a Two-Stage Dynamic Programming Algorithm~(see Algorithm~\ref{alg:detailed_frc} in the Appendix for details) 
designed to optimize the partitioning of a set of pre-segmented basic units
$U=\{u_i\}_{i=1}^n$
into chunks that strictly fall within a fixed range $[m,M]$, 
where $m$ and $M$ denote the minimum and maximum token limits, respectively.

We also design a piecewise cost function $C(l)$.
It encourages lengths to cluster within $[m,M]$ when valid, 
and minimizes the magnitude of deviation when boundary violations are inevitable.

\vspace{-5pt}
\paragraph{Pre-processing}
Any single unit $u_i$ exceeding $M$ tokens is isolated prior to the DP process.

\vspace{-5pt}
\paragraph{Stage 1: Constraint Optimization}
We search for a segmentation in which every chunk's length $l$ strictly satisfies $m\leq l \leq M$.
Let $f[i]$ denote the minimum cost to pack the first $i$ units.
If $f[n]<\infty$, the result is returned.

\vspace{-5pt}
\paragraph{Stage 2: Relaxed Optimization}
If strict segmentation is infeasible,
we relax the constraints. 
We allow at most $K$ chunks to violate the boundary $[m,M]$
and extend the DP state to $f[v][i]$, representing the minimum cost for the first $i$ units using exactly $v$ violations.
The final solution is chosen by minimizing the cost over all $v\in[1,K]$.

\subsection{Chunk Alignment}
\label{sec:algo2}
While existing sentence alignment methods~\citep{sennrich-volk-2011-iterative,thompson-koehn-2019-vecalign} 
can provide precise correspondences and achieve global optimality for chunk alignment,
the intensive preprocessing required for large-scale corpora is often computationally prohibitive.
Additionally,
perfect sentence alignment is often unattainable in real-world scenarios~\citep{obrien-etal-2025-dochplt}.

Therefore, we obviate strict sentence alignment 
by proposing a lightweight Dual-Boundary Matching based Chunk Alignment method that leverages the two boundary units of a source chunk to match high-scoring target units located in proximal relative positions within the target document; 
the span delimited by these two target units is designated as the corresponding target chunk.
Although this heuristic may lead to marginal data loss due to algorithmic constraints
(see Table~\ref{tab:frc_outliers} in the Appendix for details),
it offers a highly scalable and efficient alternative for processing massive datasets.

Algorithm \ref{alg:chunk_align} details the proposed chunk alignment strategy.
We first define \textsc{SimwRP}, a hybrid scoring function 
that aggregates similarity and relative position information 
with hyperparameters $\lambda$ and $\sigma$. 
For each source chunk $c_i$, we calculate an initial match score against all target units $t_j\in \mathcal{T}$ based on the chunk's end boundary $c_i^{end}$
and identify the top-$k$ candidates.
To reduce matching ambiguity, 
we perform dual verification,
which refines their scores by incorporating the corresponding match of the subsequent chunk's start boundary unit $c_{i+1}^{first}$.
Finally, a DP approach searches for the optimal path that maximizes the total score. 
If the boundary constraints remain unsatisfied, the candidate window size $k$ is incrementally expanded until a feasible alignment is identified or the search space is exhausted.

\subsection{FRC-based Training Formats}
\label{sec:train_strategy}
\begin{figure*}[htbp]
    \centering
    \includegraphics[width=0.90\textwidth]{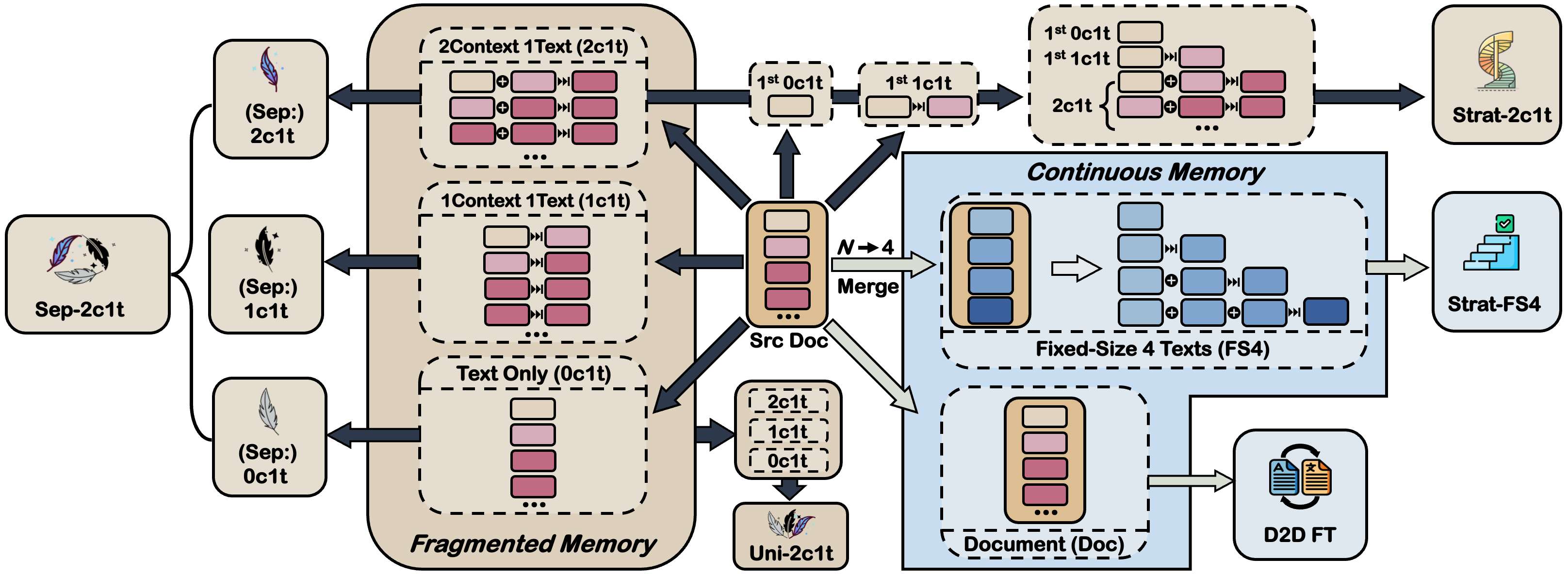}
    \caption{Construction scheme of training data for four model configurations derived from a single document.}
    \label{fig:train_format}
    \vspace{-1em}
\end{figure*}

Source documents are partitioned into fixed-range chunks using the method outlined in Section~\ref{sec:algo1}, 
and aligned via the algorithm described in Section~\ref{sec:algo2} to generate FRC pairs. 
Utilizing these pairs, we develop various training strategies.
As shown in Figure~\ref{fig:train_format},
we define five distinct data formats:
(1) \textbf{Doc}: The original, full-length document;
(2) \textbf{0c1t}: A standalone chunk pair without additional context;
(3) \textbf{1c1t}: Uses the preceding chunk as context~(the $1^{\text{st}}$ chunk serves exclusively as the contextual input);
(4) \textbf{2c1t}: Uses the two preceding chunks as context~(the $1^{\text{st}}$ and $2^{\text{nd}}$ chunks serve exclusively as contextual inputs);
(5) \textbf{FS4} (Fixed-Size 4): Chunks are merged into four segments, 
utilizing all previous chunks as context in an incremental manner. 
Leveraging these training data formats, we train four different types of models:

\vspace{-5pt}
\paragraph{\textsc{d2dFT}} A baseline model fine-tuned directly on document pairs.

\vspace{-5pt}
\paragraph{\textsc{Sep}} Independent models trained separately using 0c1t, 1c1t, and 2c1t formats. 
Their combined implementation is referred to as the \textsc{Sep}-2c1t.

\vspace{-5pt}
\paragraph{\textsc{Stair}} This category includes \textsc{Stair}-FS4, 
which utilizes the incrementally stratified data, and \textsc{Stair}-2c1t, 
which employs a ``spiral'' staircase approach that begins with 0c1t and 1c1t samples
before maintaining the 2c1t training format.

\vspace{-5pt}
\paragraph{\textsc{Uni}} A single model trained on a comprehensive mixture of 0c1t, 1c1t, and 2c1t data, 
where the three formats exhibit a clustered distribution, 
collectively forming three phases during training.

\vspace{0.4em}
Regarding data formats, 
we categorize the training data into \textit{Fragmented Memory} and \textit{Continuous Memory} types:
While $*$c1t formats utilize only partial contextual information, 
both Doc and FS4 provide access to the full document contexts.  
Regarding training methods, 
while chunks typically serve as a translation target once, 
the \textsc{Uni} model may target a single chunk up to three times.
Moreover, while \textsc{Stair} and \textsc{Uni} models undergo joint training on mixed formats, 
\textsc{Sep} models focus on ``expert'' training specialized for individual formats.

\section{GlobVDoc Dataset}
Existing test sets for DocMT in the news and social domains typically consist of relatively short documents~(e.g., WMT2022~\citep{kocmi-etal-2022-findings} and News Commentary).
To address the scarcity of long-form document pairs within this domain,
we introduce \textsc{GlobVDoc}, a 10-language document-level test set sourced from Global Voices, a multilingual news site previously included in OPUS~\citep{tiedemann-2012-parallel}.
This dataset is designed to be independent of mainstream DocMT training sources, including
\textbf{\textit{IWSLT}}~\citep{cettolo-etal-2012-wit3, cettolo-etal-2017-overview}, \textbf{\textit{BWB}}~\citep{jiang-etal-2022-blonde},
\textbf{\textit{GuoFeng}}~\citep{xu-etal-2022-guofeng},
\textbf{\textit{News Commentary}},
\textbf{\textit{OpenSubtitles}}~\citep{lison-tiedemann-2016-opensubtitles2016,lison-etal-2018-opensubtitles2018},
and \textbf{\textit{Europarl}}~\citep{koehn-2005-europarl},
thereby enabling the evaluation of out-of-distribution translation performance.

\textsc{GlobVDoc} was constructed by a team of eight individuals, 
comprising several authors and additional volunteers.
We manually selected 10 document pairs for each en $\rightarrow$ \{de, es, fr, it, ko, nl, pt, ru, zh\}.
Considering document translation methods derived from a sentence-level translation system, 
the test data was required to be strictly sentence-aligned.
To ensure consistency and quality, 
the construction process adhered to standardized guidelines,
including Source Selection and Diversity, Temporal Relevance,
Alignment Quality, Structural Consistency, Manual Segmentation,
Minimal Adjustments, and Mapping Priority~(see Appendix~\ref{sec:appen_globvdoc} for details).
The documents comprise various formats, including interviews, biographies, real-time news reports, and podcasts.
Appendix~\ref{sec:appen_globvdoc} provides detailed statistics for all language pairs in the GlobVDoc dataset (Table~\ref{tab:globvdoc_stat}), 
together with a comparison against other commonly used datasets in Table~\ref{tab:dataset_statistics}.

In addition,
we introduce a quality estimation function in Appendix~\ref{sec:qe_test} to evaluate the autocorrelation within each test dataset.

\section{Experiment}
\label{sec:exp}

\subsection{Datasets}
\label{sec:data}
We utilize the \textbf{\textit{DocBlocks}} dataset\footnote{\url{https://huggingface.co/datasets/sardinelab/DocBlocks}}~\citep{ramos2025multilingual} as our training data,
which is composed of five distinct corpora:
IWSLT, BWB, GuoFeng, News Commentary, and Europarl.

For evaluation\footnote{
In the Limitations Section, we present the reasons for not incorporating the GuoFeng test data into our study.},
we use IWSLT2017~\citep{cettolo-etal-2017-overview} and BWB test data,
as well as our self-established GlobVDoc dataset.
Initially, we utilize the IWSLT2017 test data~(TED talk documents),
consistent with~\citet{ramos2025multilingual}, 
to conduct experiments on the six language pairs
en $\leftrightarrow$ \{de, fr, it, ko, nl, zh\}.
The BWB test data was derived from continuous chapters extracted from six web novels.
Due to the relatively short length of individual chapters, 
we concatenate two adjacent chapters from the same book for zh $\rightarrow$ en evaluation.

\subsubsection{Experimental Setups}
\label{sec:main_setup}
We employ \textsc{Qwen2.5-7B-Instruct}~\citep{qwen2.5} 
and \textsc{TowerInstruct-Mistral-7B}~\citep{alves2024tower}~(hereinafter referred to as \textsc{Tower-7B-Instruct}) 
as our backbone models
to facilitate a comparison of fine-tuning performance on the shared DocBlocks dataset with \citet{ramos2025multilingual}.
Both models support 32k-token contexts, 
with \textsc{Tower-7B-Instruct} model having been post-trained for sentence-level MT.
Due to the significant discrepancy between their tokenizers, 
we construct two distinct sets of fixed-range $[256,512]$ chunk pairs by utilizing the respective tokenizer of each model for length measurement,
while the Stanza tokenizer~\citep{qi2020stanza} is jointly used to delineate the basic units,
and cosine similarity based on LaBSE~\citep{feng-etal-2022-language} embeddings is used for chunk alignment.

For reference, we also include results of some widely used commercial LLMs,
 GPT-4.1~\citep{openai2024gpt4technicalreport},
 DeepSeek-v3.2~\citep{deepseekai2025deepseekv3technicalreport}, 
 and Gemini-2.5-Pro~\citep{comanici2025gemini25pushingfrontier}.

We provide a comparison of the train-test length distributions across all the training formats in Appendix~\ref{sec:data_stat}.
The training data curation procedure, 
sample statistics, 
and setups for both training and decoding are detailed in Appendix~\ref{sec:train_setup}.

\subsubsection{Evaluation Methods}

\paragraph{d-BLEU, ds-BLEU} Classical document-level BLEU~(d-BLEU;~\citealp{papineni-etal-2002-bleu, liu-etal-2020-multilingual-denoising}) 
and ds-BLEU defined by~\citet{peng-etal-2025-investigating}, 
calculating the equivalent of sentence-level BLEU scores~\citep{lin-och-2004-automatic}.

\vspace{-5pt}
\paragraph{d-COMET}
We compute d-COMET\footnote{\url{https://huggingface.co/Unbabel/wmt22-comet-da}}
at the chunk-level using SLIDE~\citep{raunak-etal-2024-slide}
in the same manner as \citet{ramos2025multilingual}. 
However, we introduce variations to the chunking procedure~(see Appendix~\ref{sec:eval_details} for details).



\vspace{-5pt}
\paragraph{LTCR}
The Lexical Translation Consistency Ratio~(LTCR;~\citealp{lyu-etal-2021-encouraging}) measures terminology consistency in DocMT.
The conventional implementation relies on cross-lingual word alignment between the source and hypothesis, making the metric sensitive to alignment errors.
We propose a LTCR variant that performs monolingual alignment between the hypothesis and reference using the Needleman–Wunsch algorithm~\citep{Needleman1970AGM}.

Specifically, for each interest word $w$ in the reference (see Appendix~\ref{sec:ltcr-interest-words}), we collect its aligned translations in the hypothesis across all sentence pairs.
The word-level consistency is:
\begin{equation}
C(w) = \frac{\max_v \text{count}_w(v)}{K_w}, \quad K_w \geq 2
\end{equation}
where $K_w$ is the number of aligned translations and $\text{count}_w(v)$ denotes the frequency of the translation form $v$.
The document-level score is the occurrence-weighted average:
\begin{equation}
\text{LTCR} = \frac{\sum_{w \in W'} C(w) \cdot K_w}{\sum_{w \in W'} K_w}
\end{equation}
where $W' = \{w \mid K_w \geq 2\}$.

\vspace{-5pt}
\paragraph{GEMBA-DA}
While GEMBA-MQM~\citep{kocmi-federmann-2023-gemba} has demonstrated efficacy in detecting error spans at the sentence or segment level,
its applicability to document-level evaluation remains under-explored. 
Therefore, following \citet{mrozinski2025qualityestimationrerankingdocumentlevel} and \citet{Wu2024PerhapsBH},
we utilize GEMBA-DA\footnote{\url{https://github.com/MicrosoftTranslator/GEMBA}} for our evaluation.

\begin{table*}[h!]
\centering
\small
\resizebox{\textwidth}{!}{
\begin{tabular}{llcccccccccccc}
\toprule
\multirow{2}{*}{\bf Models} &
\multirow{2}{*}{\makecell{\bf Decoding\\ \bf Format}} &
\multicolumn{4}{c}{\bf IWSLT2017 en-xx} & 
\multicolumn{4}{c}{\bf IWSLT2017 xx-en} &
\multicolumn{4}{c}{\bf BWB zh-en} \\
\cmidrule(lr){3-6}\cmidrule(lr){7-10}\cmidrule(lr){11-14}
& & \makecell{\bf d-BLEU} & \makecell{\bf ds-BLEU} & \makecell{\bf d-COM.} & \makecell{\bf GEM.-DA}
&   \makecell{\bf d-BLEU} & \makecell{\bf ds-BLEU} & \makecell{\bf d-COM.} &
\makecell{\bf GEM.-DA}
&   \makecell{\bf d-BLEU} & \makecell{\bf ds-BLEU} & \makecell{\bf d-COM.} & \makecell{\bf GEM.-DA}\\
\midrule
\rowcolor{gray!15} \multicolumn{14}{c}{Large-Scale LLMs} \\
\addlinespace[1pt]
\multirow{1}{*}{\bf GPT-4.1} 
& d2d & 36.53 & 34.91 & 86.23 & 94.72 & 39.19 & 38.71 & 85.79 & 91.93 & 22.85 & 23.09 & 81.06 & 94.26  \\
\multirow{1}{*}{\bf Deepseek-v3.2} 
& d2d & 32.44 & 31.60 & 84.86 & 92.16 & 45.73 & 44.15 & 85.60 & 92.64 & 21.73 & 21.71 & 81.25 & 93.92 \\
\multirow{1}{*}{\bf Gemini-2.5-Pro} 
& d2d & 40.05 & 38.00 & 86.38 & 94.79 & 49.67 & 48.38 & 86.22 & 94.38 & 21.53 & 21.23 & 80.72 & 94.74 \\
\rowcolor{gray!15} \multicolumn{14}{c}{Qwen2.5-7B-Instruct} \\
\addlinespace[1pt]
\multirow{5}{*}{\bf Orig Model} 
& d2d & 25.88 & 27.52 & 80.00 & 61.16 & 35.54 & \underline{35.07} & 84.54 & 84.46 & 18.57 & 18.47 & 80.20 & \underline{81.21} \\
& 0c1t    & 30.00 & \underline{29.23} & 81.83 & 63.73 & 35.42 & 34.79 & \underline{85.03} & 84.34 & \underline{19.38} & \underline{19.18} & \underline{80.72} & 79.00 \\
& 1c1t    & 29.80 & 28.79 & 82.05 & 63.72 & 35.29 & 34.69 & 84.94 & \underline{84.64} & 19.07 & 18.85 & 80.47 & 79.00 \\
& 2c1t    & \underline{30.07} & 29.00 & \underline{82.42} & \underline{64.33} & 35.37 & 34.72 & 84.88 & 84.60 & 18.75 & 18.52 & 80.59 & 79.03 \\
& FS4     & 28.29 & 28.42 & 81.06 & 63.14 & \underline{35.56} & 34.79  & 83.33 & 84.57 & 18.81 & 18.66 & 80.27 & 79.56 \\
\hdashline
\multirow{2}{*}{\quad \bf w/ \textsc{d2dFT}}
& \underline{\bf d2d} & 29.17 & \underline{34.50} & 82.38 & \underline{74.31} & 22.46 & \underline{38.18} & 82.78 & \underline{80.07} & \underline{26.41} & \underline{26.27} & \underline{80.63} & \underline{84.03} \\
& 0c1t    & \underline{32.07} & 33.09 & \underline{82.76} & 74.03 & \underline{35.78} & 36.16 & \underline{83.99} & 76.24 & 25.09 & 25.31 & 79.92 & 82.36\\
\hline
\hline
\bf \multirow{5}{*}{\bf \textsc{Stair}-FS4}
& d2d & 31.42 & \underline{35.77} & \underline{83.52} & \underline{78.57} & 39.20 & 41.22 & 85.36 & 84.56 & 26.07 & 25.83 & \underline{80.80} & \underline{85.46} \\
& 0c1t    & 30.43 & 29.52 & 78.51 & 66.29 & 42.00 & 40.81 & 85.06 & 82.96 & 26.20 & 26.05 & 80.71 & 83.95 \\
& 1c1t    & 32.70 & 32.07 & 80.82 & 71.79 & 43.58 & 42.26 & 85.65 & 84.65 & 26.40 & 26.24 & 80.71 & 83.46 \\
& 2c1t    & \underline{33.05} & 32.46 & 81.16 & 73.02 & \underline{43.75} & \underline{42.43} & \bf \underline{85.73} & \underline{84.82} & \underline{26.42} & \underline{26.27} & 80.70 & 84.13 \\
& \underline{\bf FS4} & 30.68 & 33.05 & 81.57 & 75.62 & 37.30 & 42.18 & 83.94 & 84.49 & 26.40 & 26.22 & 80.49 & 84.18 \\
\hline
\bf \multirow{2}{*}{\bf \textsc{Stair}-1c1t}
& 0c1t    & 32.44 & 32.14 & 80.61 & 71.67 & 44.40 & 42.95 & 85.25 & 81.98 & \bf \underline{27.14} & 27.06 & 80.94 & 85.23 \\
& \underline{\bf 1c1t}    & \underline{36.72} & \underline{35.59} & \underline{83.25} & \underline{79.03} & \underline{46.87} & \underline{45.15} & \underline{85.62} & \underline{84.06} & \bf \underline{27.14} & \bf \underline{27.09} & \bf \underline{81.10}  & \bf \underline{86.38} \\
\hdashline
\bf \multirow{3}{*}{\bf \textsc{Stair}-2c1t}
& 0c1t    & 32.06 & 31.14 & 79.79 & 69.74 & 43.51 & 41.88 & 85.27 & 80.95 & 26.84 & \underline{26.74} & 80.76 & 84.33 \\
& 1c1t    & 36.74 & 35.34 & 83.00 & \underline{78.38} & 45.65 & 43.91 & 85.62 & 84.01 & \underline{26.77} & 26.66 & \underline{80.89} & 84.97\\
& \underline{\bf 2c1t}    & \underline{36.97} & \underline{35.48} & \underline{83.08} & 78.01 & \underline{46.24} & \underline{44.37} & \underline{85.70} & \underline{84.19} & 26.71 & 26.61 & 80.88 & \underline{85.28} \\
\hline
\bf \multirow{3}{*}{\bf \textsc{Sep}-2c1t}
& \underline{\bf 0c1t} & 38.21 & 36.82 & 84.13 & 81.67 & 45.21 & 43.79 & 84.83 & 82.89 & 26.28 & 25.76 & 80.47 & 84.15\\
& \underline{\bf 1c1t} & \bf \underline{38.51} & \bf \underline{37.00} & \bf \underline{84.70} & \bf \underline{82.42} & \bf \underline{47.07} & 45.19 & \underline{85.57} & \bf \underline{84.90} & \underline{26.58} & \underline{26.41} & \underline{80.72} & 84.74\\
& \underline{\bf 2c1t} & 38.32 & 36.54 & 84.59 & 82.40 & 46.80 & \bf \underline{45.22} & 85.47 & 84.86 & 26.44 & 26.34 & 80.64 & 84.97 \\
\toprule
\rowcolor{gray!15} \multicolumn{14}{c}{Tower-7B-Instruct-Mistral} \\
\addlinespace[1pt]
\multirow{5}{*}{\bf Orig Model} 
& d2d & 6.54  & 11.77 & 38.58 & -     & 10.36 & 15.40 & 56.49 & -     & 8.79  & 8.45  & 69.15 & - \\
& 0c1t    & \underline{32.24} & \underline{30.91} & \underline{83.61} & \underline{77.21} & 33.56 & 33.03 & \underline{84.09} & 79.12 & 17.35 & \underline{17.18} & \underline{79.51} & \underline{50.95} \\
& 1c1t    & 31.55 & 29.96 & 82.68 & 74.56 & \underline{36.83} & 34.67 & 82.84 & 79.19 & 17.48 & 16.94 & 79.26 & 50.21 \\
& 2c1t    & 31.66 & 30.00 & 82.73 & 74.96 & 36.45 & \underline{35.05} & 83.27 & \underline{80.24} & \underline{17.63} & 17.17 & 79.16 & 49.77 \\
& FS4     & 20.16 & 24.34 & 77.83 & 65.20 & 19.80 & 25.75 & 75.53 & 62.37 & 15.56 & 15.73 & 77.21 & 48.92 \\
\hdashline
\multirow{2}{*}{\quad \bf w/ \textsc{d2dFT}}
& \underline{\bf d2d}     & 34.36 & \underline{35.13} & \underline{82.51} & \underline{73.38} & 17.20 & \underline{37.91} & 75.52 & \underline{69.91} & \underline{25.14} & \underline{25.10} & \underline{80.24} & \underline{70.26} \\
& 0c1t    & \underline{34.51} & 33.67 & 82.15 & 73.05 & \underline{31.59} & 32.74 & \underline{78.18} & 68.99 & 21.99 & 21.31 & 80.20 & 53.15\\
\hline
\hline
\bf \multirow{5}{*}{\bf \textsc{Stair}-FS4}
& d2d & 21.54 & 32.12 & 80.49 & 73.76 & 31.88 & 44.81 & 80.96 & 71.28 & 26.52 & 26.64 & 80.90 & 80.13 \\
& 0c1t    & 31.35 & 30.04 & 78.48 & 66.12 & 52.34 & 51.21 & 85.33 & 79.57 & 27.43 & 27.38 & 80.96 & 82.77 \\
& 1c1t    & 34.55 & 33.37 & 81.56 & 75.09 & 56.35 & 55.40 & \underline{86.01} & 83.06 & 27.50 & 27.36 & 81.02 & \underline{83.44} \\
& 2c1t    & 34.96 & 33.90 & 81.84 & 76.10 & 56.79 & 56.15 & 85.87 & 83.32 & 27.47 & 27.38 & \underline{81.03} & 83.38 \\
& \underline{\bf FS4} & \underline{35.81} & \underline{35.32} & \underline{82.29} & \underline{81.59} & \underline{59.36} & \underline{57.98} & 84.07 & \underline{83.86} & \bf \underline{27.61} & \bf \underline{27.62} & 80.56 & 83.28 \\
\hline
\multirow{2}{*}{\bf \textsc{Stair}-1c1t}
& 0c1t    & 34.34 & 32.98 & 80.58 & 71.97 & 61.81 & 59.87 & 85.46 & 81.53 & 27.04 & 26.97 & 81.03 & 81.92 \\
& \underline{\bf 1c1t}    & \underline{38.91} & \underline{37.10} & \underline{84.09} & \underline{83.02} & \underline{66.14} & \underline{64.61} & \underline{86.29} & \underline{85.07} & \underline{27.53} & \underline{27.46} & \bf \underline{81.16} & \underline{83.33} \\
\hdashline
\multirow{3}{*}{\bf \textsc{Stair}-2c1t}
& 0c1t    & 33.45 & 31.97 & 79.39 & 70.36 & 60.46 & 58.88 & 85.37 & 80.89 & 27.32 & 27.33 & 80.96 & 81.21 \\
& 1c1t    & 38.26 & 36.38 & 84.13 & 81.55 & 65.32 & 63.90 & 86.17 & 84.54 & \underline{27.55} & \underline{27.57} & 81.06 & 83.26 \\
& \underline{\bf 2c1t}    & \underline{38.29} & \underline{36.40} & \underline{84.14} & \underline{81.70} & \underline{66.51} & \underline{64.86} & \underline{86.26} & \underline{84.93} & 27.53 & 27.50 & \underline{81.10} & \underline{83.72} \\
\hline
\multirow{3}{*}{\bf \textsc{Uni}-2c1t}
& \underline{\bf 0c1t} & 36.32 & 34.51 & 84.00 & 76.36 & 63.71 & 61.21 & 85.85 & 77.86 & 25.43 & 25.26 & 80.79 & 78.15 \\ 
& \underline{\bf 1c1t} & \underline{36.65} & \underline{34.75} & \underline{84.40} & 77.15 & 65.82 & 63.32 & \underline{86.15} & 79.44 & \underline{25.68} & \underline{25.51} & 80.97  & 78.85 \\
& \underline{\bf 2c1t} & 36.53 & 34.72 & 84.33 & \underline{77.32} & \underline{66.06} & \underline{63.55} & \underline{86.15} & \underline{80.68} & 25.62 & 25.38 & \underline{80.98} & \underline{79.41} \\
\hline
\bf \multirow{3}{*}{\bf \textsc{Sep}-2c1t}
& \underline{\bf 0c1t} & \bf \underline{39.61} &\bf \underline{37.87} & 84.69 & 86.29 & 63.05 & 61.75 & 86.09 & 83.56 & \underline{27.38} & \underline{27.42} & \underline{81.11} & \bf \underline{83.95} \\
& \underline{\bf 1c1t} & 39.51 & 37.86 & \bf \underline{84.83} & \bf \underline{86.34} & 66.84 & \bf \underline{65.29} & 86.49 & 85.44 & 27.16 & 27.14 & 81.03 & 83.31 \\
& \underline{\bf 2c1t} & 38.78 & 37.54 & 84.80 & 85.80 & \bf \underline{67.20} & 65.12 & \bf \underline{86.53} & \bf \underline{85.73} & 26.82 & 26.77 & 81.03 & 83.77\\
\bottomrule
\end{tabular}
}
\caption{Translation results of different models under various decoding formats. 
Decoding formats that are consistent with the model training setup are highlighted in \underline{\textbf{bold}}.
For the four evaluation metrics, the best performance achieved under the same backbone model is also shown in \textbf{bold}.
For each individual model, comparisons across different decoding formats are indicated by \underline{underlining} the better-performing results.}
\label{tab:main_result}
\vspace{-1em}
\end{table*}

\begin{table*}[t]
\centering
\small
\resizebox{\textwidth}{!}{
\begin{tabular}{lcccccccccc}
\toprule
\multirow{2}{*}{\bf Models} &
\multicolumn{10}{c}{\bf GlobVDoc en-xx (d-BLEU / d-COMET)} \\
\cmidrule(lr){2-11}
& \bf de & \bf es & \bf fr & \bf it & \bf ko & \bf nl & \bf pt & \bf ru & \bf zh & \bf all\\
\midrule
\rowcolor{gray!15} \multicolumn{11}{c}{Large-Scale LLMs} \\
\addlinespace[1pt]
\multirow{1}{*}{\bf GPT-4.1} 
& 35.58 / 88.63 & 58.32 / 89.43 & 47.38 / 88.82 & 44.95 / 89.45 & 30.88 / 91.33 & 43.13 / 89.75 & 60.39 / 89.92 & 35.07 / 92.00 & 54.82 / 90.97 & 45.61 / 90.03\\
\multirow{1}{*}{\bf GPT-4.1-mini} 
& 34.80 / 88.53 & 57.29 / 89.17 & 46.63 / 88.98 & 43.65 / 89.02 & 30.93 / 90.88 & 41.76 / 89.63 & 60.19 / 89.91 & 34.57 / 91.71 & 52.12 / 90.34 & 44.65 / 89.80\\ 
\multirow{1}{*}{\bf Deepseek-v3.2} 
& 33.32 / 88.19 & 53.48 / 88.86 & 49.38 / 88.52 & 44.84 / 89.20 & 32.14 / 91.34 & 41.13 / 89.35 & 57.08 / 89.81 & 31.89 / 91.66 & 45.95 / 89.60 & 43.24 / 89.61\\
\multirow{1}{*}{\bf Gemini-2.5-Pro} 
& 37.28 / 88.70 & 54.08 / 89.23 & 53.03 / 89.06 & 47.57 / 89.69 & 35.39 / 91.67 & 43.98 / 90.12 & 56.37 / 89.08 & 34.32 / 91.99 & 56.80 / 91.18 & 46.54 / 90.08 \\
\rowcolor{gray!15} \multicolumn{11}{c}{Agent-based Methods} \\
\multirow{1}{*}{\bf GRAFT-Q} 
& 29.20 / 87.54 & 49.83 / 88.21 & 43.93 / 87.96 & 39.11 / 88.72 & 28.68 / 90.22 & 32.59 / 87.66 & 53.53 / 89.48 & 26.55 / 90.46 & 51.74 / 90.29 & 39.46 / 88.96\\
\multirow{1}{*}{\bf GRAFT-G} 
& 34.65 / 88.36 & 54.31 / 88.73 & 47.81 / 88.57 & 44.81 / 88.71 & 33.73 / 91.42 & 40.06 / 89.44 & 59.79 / 89.73 & 31.98 / 91.21 & 55.60 / 90.92 & 44.75 / 89.68\\
\multirow{1}{*}{\bf DELTA-Q} 
& 29.37 / 87.24 & 50.69 / 88.17 & 44.87 / 87.57 & 40.24 / 88.33 & 29.18 / 90.29 & 32.94 / 87.60 & 53.54 / 89.42 & 25.78 / 89.90 & 54.27 / 90.62 & 40.10 / 88.79 \\
\multirow{1}{*}{\bf DELTA-G} 
& 34.76 / 88.70 & 55.34 / 88.99 & 47.28 / 88.94 & 42.77 / 89.41 & 35.22 / 91.48 & 40.62 / 89.73 & 59.41 / 90.11 & 30.86 / 91.46 & 54.13 / 90.52 & 44.49 / 89.93 \\
\rowcolor{gray!15} \multicolumn{11}{c}{Qwen2.5-7B-Instruct} \\
\addlinespace[1pt]
\multirow{1}{*}{\bf Orig Model} 
& 23.50 / 84.92 & 47.35 / 88.02 & 40.50 / 85.80 & 35.01 / 86.70 & 15.67 / 81.03 & 27.28 / 84.44 & 48.76 / 88.50 & 21.92 / 84.76 & 52.85 / \textbf{90.65} & 34.76 / 86.09 \\
\multirow{1}{*}{\quad \bf w/ \textsc{d2dFT}}
& 31.56 / 87.97 & \textbf{52.91} / 88.77 & 46.74 / 88.01 & \textbf{44.64} / 88.35 & 18.48 / 82.86 & 35.90 / 89.06 & \textbf{50.33} / 88.28 & 28.14 / 89.30 & \textbf{54.84} / 90.17 & 40.39 / 88.08 \\
\hline
\multirow{1}{*}{\bf \textsc{Stair}-FS4}
& 31.76 / 87.24 & 36.57 / 84.66 & 47.02 / 87.47 & 43.53 / 87.97 & 11.59 / 87.91 & 36.28 / 88.29 & 50.26 / 87.22 & 28.11 / 89.98 & 54.22 / 88.36 & 37.70 / 87.68 \\
\multirow{1}{*}{\bf \textsc{Stair}-2c1t}
& \textbf{31.84} / 87.99 & 51.95 / \textbf{88.79} & \textbf{47.91} / 88.29 & 43.24 / 88.41 & \textbf{27.52} / \textbf{89.82} & 36.66 / \textbf{89.22} & 49.20 / \textbf{88.46} & 27.53 / \textbf{90.05} & 53.91 / 90.26 & \textbf{41.08} / \textbf{89.05} \\
\multirow{1}{*}{\bf \textsc{Sep}-2c1t}
& 31.39 / \textbf{88.09} & 52.62 / 88.78 & 46.64 / \textbf{88.32} & 43.20 / \textbf{88.46} & 16.89 / 81.16 & \textbf{36.71} / 89.09 & 48.54 / 88.26 & \textbf{28.47} / 89.66 & 46.96 / 83.19 & 39.05 / 87.49 \\
\toprule
\rowcolor{gray!15} \multicolumn{11}{c}{Tower-7B-Instruct-Mistral} \\
\addlinespace[1pt]
\multirow{1}{*}{\bf Orig Model} 
& 31.79 / 87.53 & 50.13 / 87.69 & 41.84 / 87.30 & 36.64 / 86.50 & 28.33 / 87.84 & 37.01 / 89.01 & 44.19 / 85.74 & 25.55 / 88.96 & 42.76 / 86.85 & 37.58 / 87.49\\
\multirow{1}{*}{\quad \bf w/ \textsc{d2dFT}}
& 30.76 / 84.60 & 50.20 / 87.81 & 46.31 / 87.35 & 43.72 / 86.16 & 25.10 / 89.62 & 15.05 / 79.57 & 48.06 / 86.96 & 26.17 / 82.98 & 53.95 / 90.11 & 37.70 / 85.79\\
\hline
\bf \multirow{1}{*}{\bf \textsc{Stair}-FS4}
& 33.74 / 87.51 & \textbf{51.75} / 87.49 & 47.75 / 87.66 & 43.93 / 87.87 & \textbf{30.25} / 89.81 & 39.03 / 88.49 & \textbf{49.83} / 87.22 & 28.40 / 90.33 & 52.93 / 88.40 & 41.96 / 88.31\\
\multirow{1}{*}{\bf \textsc{Stair}-2c1t}
& \textbf{34.06} / \textbf{88.30} & 51.31 / \textbf{88.76} & 48.29 / \textbf{88.15} & \textbf{44.69} / \textbf{88.99} & 27.94 / \textbf{90.45} & \textbf{39.59} / 89.56 & 49.36 / \textbf{88.54} & \textbf{29.28} / \textbf{91.18} & \textbf{54.01} / \textbf{90.23} & \textbf{42.06} / \textbf{89.35} \\
\bf \multirow{1}{*}{\bf \textsc{Sep}-2c1t}
& 33.08 / 88.16 & 51.64 / 88.75 & \textbf{48.62} / 88.03 & 44.66 / 88.74 & 29.37 / 90.37 & 38.97 / \textbf{89.61} & 49.29 / 88.53 & 29.20 / 91.09 & 52.97 / 90.04 & 41.98 / 89.26\\
\bottomrule
\end{tabular}
}
\caption{Translation Benchmark on GlobVDoc.
For the two evaluation metrics, the best performance achieved under the same backbone model is shown in \textbf{bold}.
Large-scale LLMs utilize d2d decoding, 
while original model follows a 0c1t format.
Other fine-tuned models utilize decoding format corresponds to the training configuration.
\{GRAFT, DELTA\}-\{Q,G\} denote using \textsc{Qwen3-30B-A3B-Instruct} and GPT-4.1-mini as backbone models, respectively.
}
\label{tab:bench_globvdoc}
\vspace{-1em}
\end{table*}

\subsection{Results on IWSLT2017 and BWB}
\label{sec:rslt_iwslt_bwb}

\paragraph{Overview}
The performance of models trained with two different backbone architectures is reported in Table~\ref{tab:main_result}.
First, chunking generally outperforms the d2d strategy for both original models, 
a benefit especially pronounced for \textsc{Tower} due to its prior sentence-level fine-tuning.
While all fine-tuned models naturally exceed the baselines on in-domain data, 
\textsc{Tower}-based models show more substantial gains than \textsc{Qwen2.5}.

Since our evaluated d-BLEU scores for the original models differ from those reported by \citet{ramos2025multilingual}, 
we compare the relative gains over the original model in the d2d mode.
On IWSLT2017, 
their best results via chunking reported improvements of approximately 6/26 d-BLEU points on en-xx and 13.5/47 on xx-en for \textsc{Qwen2.5} / \textsc{Tower}, 
whereas our \textsc{Sep}-2c1t model achieves gains of 12.44~(+6.44), 32.24~(+6.24), 11.26~(-2.24), and 56.84~(+9.84), respectively. 
Consequently, our approach outperforms theirs in most instances,
except for the \textsc{Qwen2.5}-based IWSLT xx–en case.

\vspace{-3pt}
\paragraph{Model-Wise Analysis} 
From here on, we will describe the model-wise analysis.

\textbf{\textsc{d2dFT}}: 
Although the d2d decoding strategy consistently yields higher ds-BLEU scores than 0c1t,
its d-BLEU scores on IWSLT2017 are markedly lower. 
Given that d-BLEU assigns greater weight to longer documents than ds-BLEU, 
this indicates that while the overall translation quality of the \textsc{d2dFT} model under d2d decoding is higher than 0c1t, 
chunking remains more effective for translating longer documents.

\textbf{\textsc{Stair}-FS4}:
Instead of rigidly dividing into four equal-sized chunks,
this approach applies FRC followed by merging,
which results in a highly discretized chunk-length distribution, 
increasing the coverage of test-time sequence lengths during training.
Across all evaluation metrics, 
\textsc{Stair}-FS4 outperforms \textsc{d2dFT} on IWSLT2017 and achieves comparable results on BWB.
However, a latent issue remains regarding the mismatch between the FS4 training data format and the optimal decoding format, 
particularly in the \textsc{Qwen2.5} variant.

\textbf{\textsc{Stair}-1c1t / 2c1t}: 
As 1c1t and 2c1t samples dominate training, 
using their corresponding formats during decoding naturally yields optimal performance.
Moreover, in terms of d(s)-BLEU,
both models consistently exceed \textsc{Stair}-FS4 on IWSLT2017, 
despite marginal differences on BWB.
A direct comparison between \textsc{Stair}-1c1t and \textsc{Stair}-2c1t reveals a narrow performance gap, 
with the former maintaining a slight advantage.

\textbf{\textsc{Sep} / \textsc{Uni}-2c1t}:
Leveraging three independently trained sub-models, 
the \textsc{Sep} strategy achieves superior performance across all $*$c1t decoding modes and yields the best overall results on IWSLT2017,
attaining d(s)-BLEU and d-COMET parity with large-scale LLMs. 
Conversely, joint training with mixed formats~(\textsc{Uni}-2c1t) significantly degrades accuracy. 
This decline likely stems from training confusion, 
as the same target chunk is mapped to three source texts with varying contextual depth
(see Appendix~\ref{sec:uni_training} for further discussion).

\subsection{Benchmark on GlobVDoc}
\label{sec:globv_bench}
Based on the proposed GlobVDoc test data, 
we establish a translation benchmark in Table~\ref{tab:bench_globvdoc}.
In the benchmark, we include two agent-based methods,
GRAFT~\citep{dutta-etal-2025-graft} and DELTA~\citep{wang2025delta},
with \textsc{Qwen3-30B-A3B-Instruct} and GPT-4.1-mini as the underlying models.

For the \textsc{Qwen2.5}-based models,
performance differences are mainly seen in en $\rightarrow$ \{es, ko, zh\}.
Notably, \textsc{Stair}-FS4 underperforms on es and ko,
while \textsc{Sep}-2c1t lags on ko and zh. 
Therefore, \textsc{Stair}-2c1t achieves the highest overall accuracy. 
While most \textsc{Tower}-based models demonstrate consistent performance across various language pairs,
the \textsc{d2dFT} model exhibits an obvious degradation in quality for en $\rightarrow$ \{de, nl, ru\} translations.

Unlike results on IWSLT2017 and BWB, 
performance on the out-of-distribution GlobVDoc dataset exhibits a gap compared to large LLMs, 
despite achieving parity in certain language pairs.
This deficiency is most pronounced for pt, nl, and ko, 
likely due to limited samples in the DocBlocks dataset.
Furthermore, 
while most of our models outperform \{GRAFT,DELTA\}-Q, 
they still trail GPT-4.1-mini. 

\begin{figure*}[htbp]
    \centering
    \includegraphics[width=0.9\textwidth]{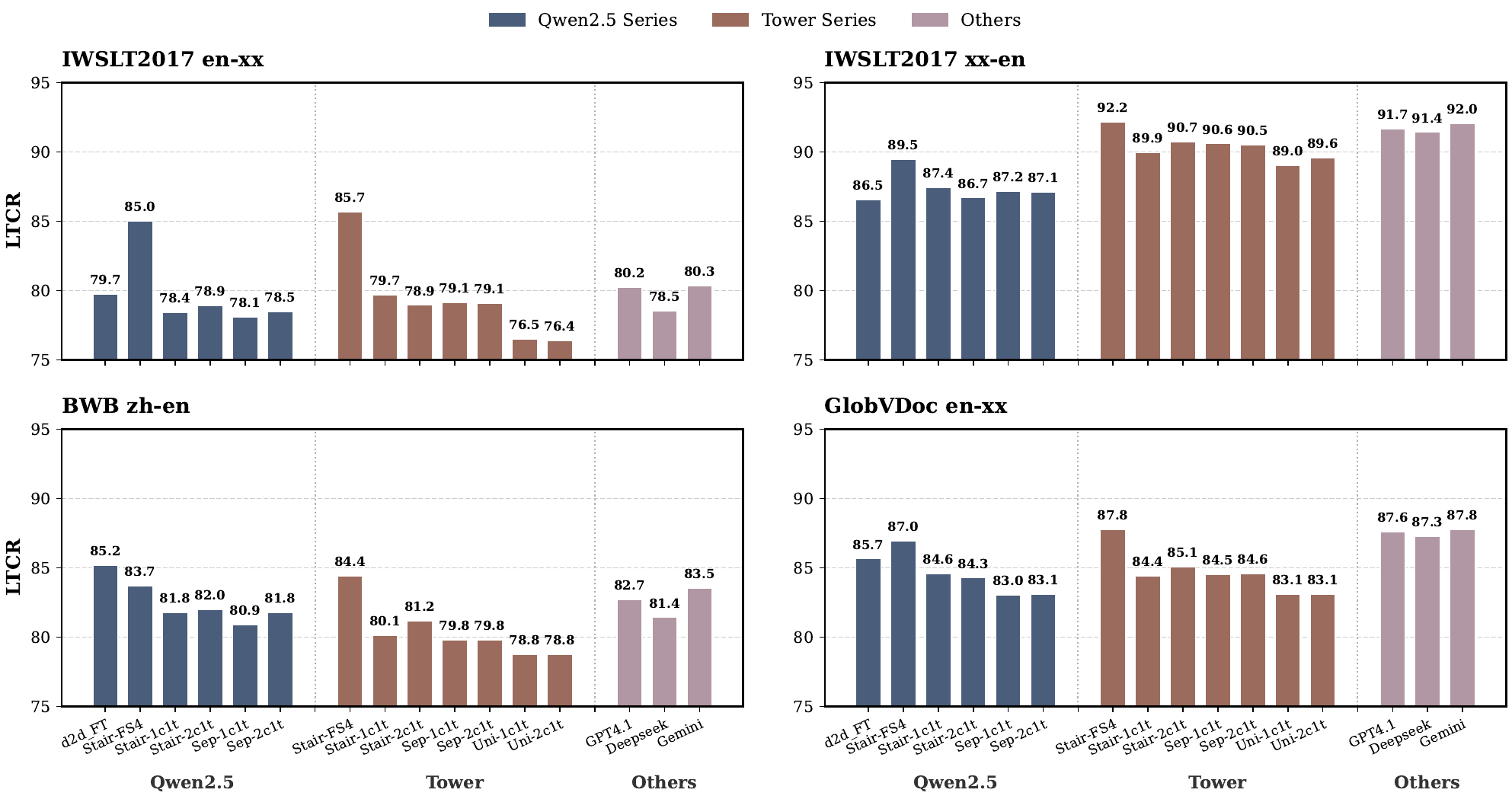}
    \caption{LTCR scores of various models on each dataset.}
    \label{fig:ltcr_results}
    \vspace{-1em}
\end{figure*}

In addition, Figure~\ref{fig:ltcr_results} reports the LTCR results for different models.
The \textsc{d2dFT} and \textsc{Stair-FS4} models, 
leveraged by \textit{Continuous Memory} training~(Section~\ref{sec:train_strategy}),
yield superior LTCR results,
with the latter matching or surpassing large-scale LLMs.
In contrast, while other models may achieve high accuracy scores, 
their limited contextual scope results in diminished terminology consistency.

\section{Analysis}
\label{sec:aba}
Beyond the experiments discussed in this section,
Appendix~\ref{sec:con_src_vs_mt} compares Source-Only and Source + MT context strategies, 
and Appendix~\ref{sec:ttc} reports results of length-centric analysis and FRC unit selection during test time.

\subsection{Extrapolating Position Embedding}
Prior work has extensively explored Position Embedding~(PE) to mitigate the discrepancy between sequence length distributions during training and inference. 
In this paper, we evaluate two primary strategies for comparison:
(1) Position Offset Augmentation, which enhances model generalization 
across diverse length distributions by introducing an offset to position indices.
Specifically, we implement two variants: SHAPE~\citep{kiyono-etal-2021-shape,ruoss-etal-2023-randomized},
which samples offsets uniformly from the gap between the sample length and the model’s maximum window size $M$;
and Uniform SHAPE~(UnifPE;~\citealp{peng-etal-2025-investigating}),
which makes each position within $[0,M−1]$ equally likely during training.
(2) Position Interpolation (PI;~\citealp{chen2023extendingcontextwindowlarge}),
which compresses or interpolates the position indices of long sequences into a pre-defined range. 
We implement Linear PI, 
which downscales position indices in both training and test data by a factor of 4 into $[0,8192]$.

The results\footnote{
Since \textsc{Tower} uses Sliding Window Attention, 
applying Linear PI would not remove the local-attention constraint. 
We therefore evaluate Linear PI on \textsc{Qwen2.5} with full attention.
} 
are summarized in Table~\ref{tab:PE}.
All evaluated PE-based strategies successfully enhance the \textsc{d2dFT} baseline. 
This improvement is especially evident on the longer sequences of the IWSLT test set, 
where d-BLEU scores increased by 5.22, 9.25, and 10.58 across the backbone models. 
Because standard d2d decoding is prone to instability and n-gram repetition~\citep{jin2024chaptertochaptercontextawareliterarytranslation,hiraoka-inui-2025-repetition}, 
PE-based interventions effectively alleviate these hallucinations and stabilize text generation~\citep{peng-etal-2025-investigating}. 
\textsc{Sep}-1c1t model achieves the best performance compared with these baselines,
showing clear improvements in both accuracy and n-gram repetition rate.
This advantage is particularly pronounced on the IWSLT dataset.

\begin{table}[h]
\centering
\small
\renewcommand{\arraystretch}{1.2} 

\begin{adjustbox}{width=\columnwidth}
\begin{tabular}{lllccc}
\toprule
\multirow{2}{*}{\bf Models} &
\multirow{2}{*}{\bf PE} &
\multirow{2}{*}{\bf FT} &
\multicolumn{3}{c}{\bf d-BLEU$\uparrow$ (NRR$\downarrow$)} \\
\cmidrule(lr){4-6}
& & &
\bf IWSLT & \bf BWB  & \bf GlobVDoc \\
\midrule
\bf \multirow{3}{*}{\bf \textsc{Qwen}}
& R.    & \textsc{d2d} & 25.82 (5.75\%) & 26.27 (0.00\%) & \textbf{40.39} (\textbf{0.00\%})\\
& R.+L. & \textsc{d2d} & 36.40 (1.34\%) & 26.12 (0.00\%) & 39.76 (1.11\%)\\
\cdashline{2-6}
& R.    & \textsc{Sep} & \textbf{42.79} (\textbf{1.07\%}) & \textbf{26.58} (0.00\%) & 39.16 (2.22\%)\\
\hline
\hline
\bf \multirow{4}{*}{\bf \textsc{Tower}}
& R.    & \textsc{d2d} & 25.78 (13.64\%) & 25.14 (0.00\%) & 37.70 (3.33\%) \\
& R.+S.& \textsc{d2d} & 31.00 (8.83\%) & 26.51 (0.00\%) & 41.11 (2.22\%) \\
& R.+U. & \textsc{d2d} & 35.03 (7.36\%) & 26.31 (0.00\%) & 41.52 (\textbf{0.00\%}) \\
\cdashline{2-6}
& R.    & \textsc{Sep} & \textbf{53.18} (\textbf{1.07\%}) & \textbf{27.16} (0.00\%) & \textbf{41.87} (\textbf{0.00\%})\\
\bottomrule
\end{tabular}
\end{adjustbox}
\caption{\textsc{Sep}-1c1t v.s. PE-based methods.
IWSLT values denote the average of en-xx and xx-en.
Abbreviations: R.~(RoPE), S.~(SHAPE), U.~(UnifPE) and L.~(Linear PI).
NRR denotes the n-gram repetition rate.
}
\label{tab:PE}
\vspace{-1.5em}
\end{table}

\subsection{Chunking Strategy}
\label{sec:chunk_strategy}

In addition to our proposed FRC strategy and the FS4 strategy followed \citet{ramos2025multilingual},
we further compare our method with two conventional chunking strategies in this section:

\paragraph{Fixed Window Chunking (FW).} 
This method greedily traverses the document with a fixed-length sliding window, 
concatenating consecutive sentences into a chunk (aka., a \textit{blob}; \citealp{finkelstein-etal-2024-introducing, wang-etal-2025-bimax})
as long as their combined length does not exceed the window size. 
Since most chunks are close to the window size,
FW is the baseline most similar to FRC. 
However, it suffers from the short-tail problem, 
where the final chunk of each document can have an arbitrary length ranging from one sentence up to the window size. 
FRC can be viewed as an enhanced version of FW by additionally enforcing a lower bound on chunk length, 
resulting in a narrower chunk-length distribution.
We set the window size to 512 tokens.

\vspace{-4pt}

\paragraph{Fixed Number Chunking (FN).}
This method greedily concatenates a fixed number of consecutive sentences into each chunk.
Similar to FW,
the final chunk of a document may contain fewer sentences than the predefined number.
Moreover, unlike FW, 
the resulting chunk length is not directly controlled,
as it depends on the lengths of the sentences.
Following \citet{alabi-etal-2025-afridoc},
we set the fixed number to 10 for our experiments.

For both chunking strategies,
we employed the same training data generation pipeline as used for FRC. 
All models were trained on \textsc{Tower}-7B under the \textsc{Sep}-2c1t format.

As shown in Table~\ref{tab:chunking_results},
the proposed FRC strategy consistently outperforms the conventional FW and FN chunking strategies across almost evaluation settings.
The exception is the OOD GlobVDoc dataset,
where all chunking strategies achieve similar performance. 
In terms of d-BLEU, 
FW and FN remain competitive with FRC on IWSLT2017 en-xx and BWB, 
although a noticeable gap persists on the IWSLT2017 xx-en.
In contrast, GEMBA-DA reveals a clearer advantage for FRC, 
which surpasses other chunking methods by approximately 1.0 point or more on IWSLT2017 and BWB.

\begin{table}[htbp]
\centering
\small
\renewcommand{\arraystretch}{1.2} 
\begin{adjustbox}{width=\columnwidth}
\begin{tabular}{lllccc}
\toprule
\multirow{2}{*}{\bf Models} &
\multirow{2}{*}{\bf Decode} &
\multicolumn{4}{c}{\bf d-BLEU / Gemba-DA} \\
\cmidrule(lr){3-6}
& & 
\bf IWS. en-xx & \bf IWS. xx-en & \bf BWB  & \bf GlobV. \\
\midrule
\bf FS: \textsc{Stair}
& FS4 & 35.81 / 81.59 & 59.36 / 83.86 & \textbf{27.61} / 83.28 & 41.96 / \textbf{86.73}\\
\midrule
\bf \multirow{3}{*}{\bf FRC: \textsc{Sep}}
& 0c1t & \textbf{\underline{39.61}} / 86.29 & 63.05 / 83.56 & \underline{27.38} / \textbf{\underline{83.95}} & 41.92 / \underline{86.58}\\
& 1c1t & 39.51 / \textbf{\underline{86.34}} & 66.84 / 85.44 & 27.16 / 83.31 & 41.87 / 86.53\\
& 2c1t & 38.78 / 85.80 & \textbf{\underline{67.20}} / \textbf{\underline{85.73}} & 26.82 / 83.77 & \underline{41.98} / 86.42\\
\midrule
\bf \multirow{3}{*}{\bf FW: \textsc{Sep}}
& 0c1t & \underline{39.02} / 84.85 & 62.03 / 82.59 & \underline{27.15} / \underline{83.21} & \textbf{\underline{42.15}} / 86.29\\
& 1c1t & 38.67 / \underline{85.41} & 65.04 / 83.97 & 26.67 / 83.10 & 41.95 / 86.38\\
& 2c1t & 38.45 / 85.39 & \underline{65.39} / \underline{84.32} & 26.51 / 82.64 & 41.91 / \underline{86.41}\\
\midrule
\bf \multirow{3}{*}{\bf FN: \textsc{Sep}}
& 0c1t & \underline{39.19} / \underline{84.77} & 61.10 / 83.02 & 27.12 / 82.26 & \underline{42.11} / \underline{86.19}\\
& 1c1t & 39.09 / 84.61 & 65.64 / 84.45 & \underline{27.17} / \underline{82.69} & 41.92 / 86.10\\
& 2c1t & 38.98 / 84.45 & \underline{66.17} / \underline{84.82} & 26.79 / 82.38 & 41.90 / 86.00\\
\bottomrule
\end{tabular}
\end{adjustbox}
\caption{
Comparison of various chunking strategies.
\textbf{Bold} denotes the highest score achieved on each dataset,
while \underline{underline} indicates the best performance for each \textsc{Sep} model on each dataset.
}
\label{tab:chunking_results}
\vspace{-1.0em}
\end{table}

\subsection{Fixed Range}
Following the same data curation pipeline, 
we constructed two additional datasets based on fixed ranges of
$[0,256]$ and $[512,768]$
(see Appendix~\ref{sec:data_stat} for data statistics).
We fine-tuned the \textsc{Tower} model using the \textsc{Sep}-1c1t strategy,
with the results shown in Table~\ref{tab:fl_ranges}.
Across the three fixed ranges, 
the performance discrepancies in the en-xx direction are relatively marginal,
with the $[0,256]$ interval performing slightly worse than the other two.
Conversely, in the xx-en direction, 
the $[256,512]$ range shows a clear advantage. 
Furthermore, 
the results consistently substantiate a strong dependency on 
contextual information for IWSLT xx-en.

\begin{figure}[t]
  \centering
  \includegraphics[width=\linewidth,trim=1.4cm 3.5cm 1.8cm 3cm, clip]{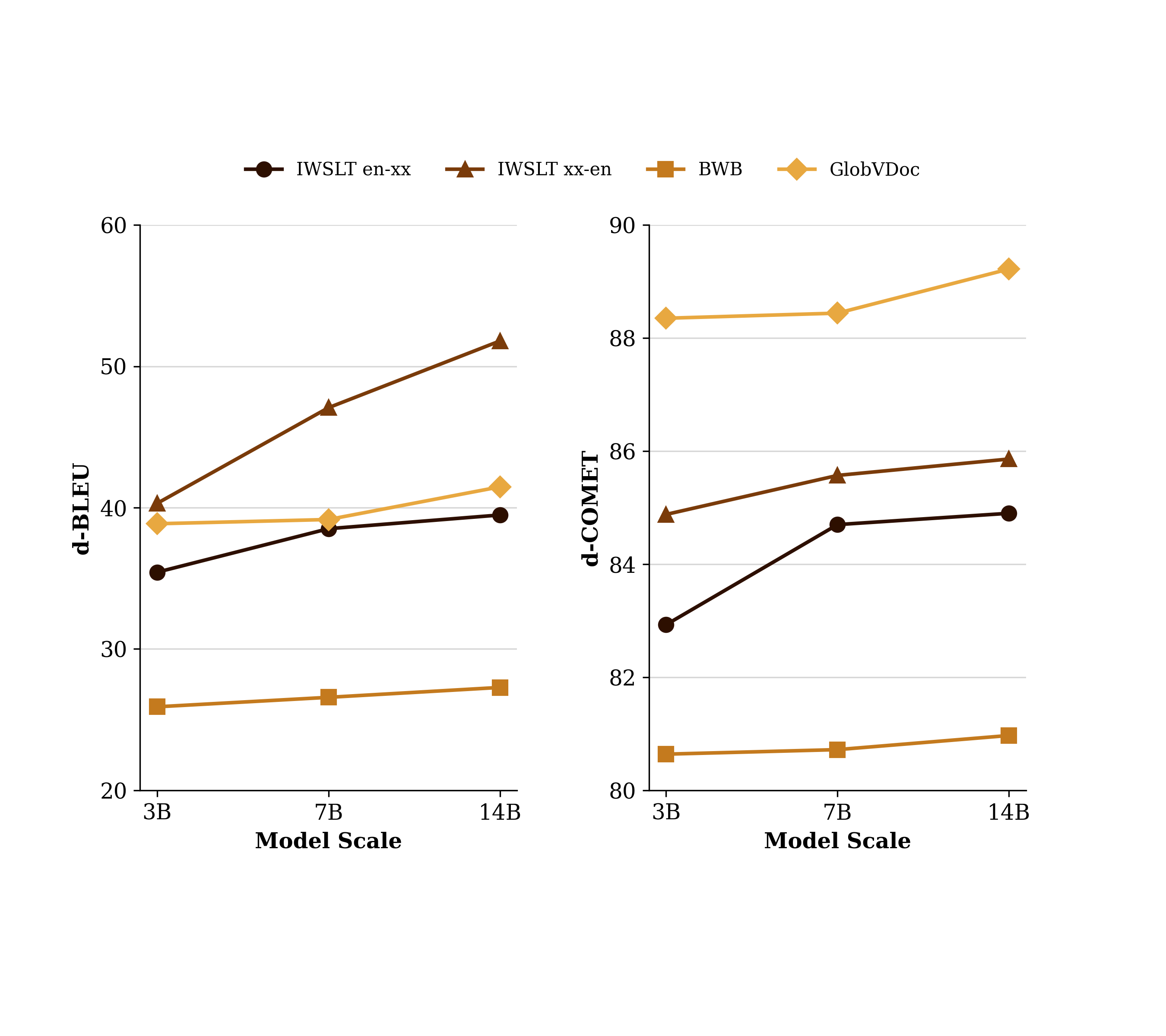}
  \caption{Performance of the \textsc{Sep}-1c1t training strategy under the $[256, 512]$ FRC setting across different model scales in the \textsc{Qwen2.5} series.}
  \label{fig:qwen2.5_scale}
\end{figure}

\begin{table}[htbp]
\centering
\small
\renewcommand{\arraystretch}{1.2} 
\begin{adjustbox}{width=\columnwidth}
\begin{tabular}{lccccc}
\toprule
\multirow{2}{*}{\bf Models} &
\multicolumn{2}{c}{\bf en-xx} & &
\multicolumn{2}{c}{\bf xx-en} \\
\cmidrule(lr){2-3} \cmidrule{5-6}
& \bf IWSLT & \bf GlobVDoc & &
\bf IWSLT & \bf BWB \\
\midrule
\bf \textsc{S-0} & 39.37 / 78.6 & 41.79 / 84.2 & & 59.12 / 88.3 & 26.83 / 80.1\\
\bf \textsc{S-1} & 39.50 / 78.7 & 41.66 / 83.9 & & 62.83 / 89.4 & 26.76 / 78.7 \\
\bf \textsc{M-0} & 39.61 / 79.2 & 41.92 / \textbf{85.5} & & 61.75 / 89.9 & \textbf{27.38} / 80.8\\
\bf \textsc{M-1} & 39.51 / 79.1 & 41.87 / 84.5 & & \textbf{66.84} / \textbf{90.6} & 27.16 / 79.8\\
\bf \textsc{L-0} & \textbf{39.66} / \textbf{79.4} & \textbf{42.15} / 85.1 & & 61.70 / 89.0 & 26.61 / \textbf{81.7}\\
\bf \textsc{L-1} & 39.61 / 79.1 & 41.67 / 84.8 & & 63.62 / 89.5 & 27.03 / 80.9\\
\bottomrule
\end{tabular}
\end{adjustbox}
\caption{
Results of \textsc{Tower}-based \textsc{Sep} models trained on fixed ranges S, M, and L, corresponding to $[0,256]$, $[256,512]$, and $[512,768]$, respectively.
Results are reported as d-BLEU / LTCR.
$*$-0 / $*$-1 denote 0c1t / 1c1t. 
}
\label{tab:fl_ranges}
\vspace{-1.5em}
\end{table}

\begin{figure*}[htbp]
    \centering
    \includegraphics[width=1.0\textwidth]{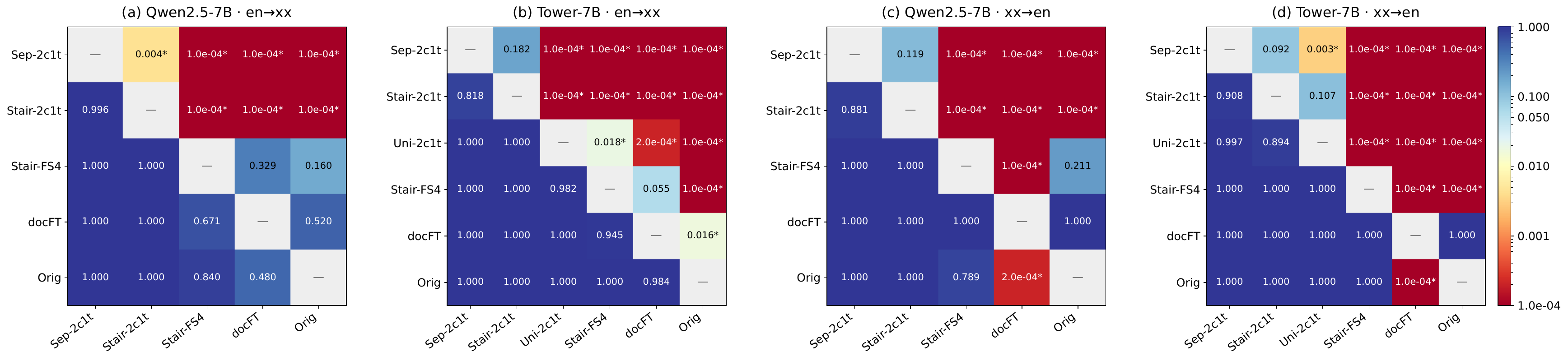}
    \caption{Pairwise one-sided significance tests on IWSLT2017 for \textsc{Qwen2.5}-7B and \textsc{Tower}-7B.
    Each cell reports the paired document-bootstrap p-value for the hypothesis that the row system outperforms the column system in d-BLEU.
    Asterisks indicate \(p<0.05\).}
    \label{fig:iwslt_stat}
\end{figure*}

\begin{table*}[t]
\centering
\setlength{\tabcolsep}{5.5pt}
\renewcommand{\arraystretch}{1.20}

\resizebox{\textwidth}{!}{
\begin{tabular}
{
>{\bfseries}l
*{3}{c}
@{\hspace{10pt}}
*{3}{c}
@{\hspace{10pt}}
*{3}{c}
}
\toprule
\multirow{2}{*}{\bf Models}
& \multicolumn{3}{c}{\textbf{\textsc{Sep}-2c1t (LoRA)}}
& \multicolumn{3}{c}{\textbf{\textsc{Stair}-2c1t (Full)}}
& \multicolumn{3}{c}{\textbf{\textsc{Sep}-2c1t (Full)}} \\

\cmidrule(lr){2-4}
\cmidrule(lr){5-7}
\cmidrule(lr){8-10}

& \textbf{0c1t}
& \textbf{1c1t}
& \textbf{2c1t}
& \textbf{0c1t}
& \textbf{1c1t}
& \textbf{2c1t}
& \textbf{0c1t}
& \textbf{1c1t}
& \textbf{2c1t} \\

\midrule

Training
& \multicolumn{3}{c}{3 LoRA}
& \multicolumn{3}{c}{1 Full}
& \multicolumn{3}{c}{3 Full} \\

Deployment
& \multicolumn{3}{c}{1 model + 3 adapters}
& \multicolumn{3}{c}{1 model}
& \multicolumn{3}{c}{3 models} \\

\midrule

IWSLT en-xx
& 38.61 / 84.04
& 38.65 / 84.25
& 37.85 / 84.26
& 33.45 / 79.39
& 38.26 / 84.13
& 38.29 / 84.14
& \textbf{39.61} /84.69
& 39.51 / \textbf{84.83}
& 38.78 / 84.80\\

IWSLT xx-en
& 43.04 / 85.06
& 43.56 / 86.15
& 43.88 / 86.30
& 60.46 / 85.37
& 65.32 / 86.17
& 66.51 / 86.26
& 63.05 / 86.09
& 66.84 / 86.49
& \textbf{67.20 /86.53} \\

BWB
& 25.99 / 80.42
& 25.96 / 80.52
& 25.74 / 80.51
& 27.32 / 80.96
& \textbf{27.55} / 81.06
& 27.53 / 81.10
& 27.38 / \textbf{81.11}
& 27.16 / 81.03
& 26.82 / 81.03 \\

GlobVDoc
& 42.78 / 89.02
& \textbf{43.28 / 89.52}
& 42.76 / 89.39
& 41.57 / 88.98
& 42.16 / 89.34
& 42.06 / 89.35
& 41.92 / 89.11
& 41.87 / 89.25
& 41.98 / 89.26\\

\bottomrule
\end{tabular}
}
\caption{
Comparison of different training and deployment configurations.
The values $*$ / $*$ denote d-BLEU and d-COMET, respectively. 
Within each dataset, the highest score is highlighted in \textbf{bold}.}
\label{tab:trade_off}
\vspace{-1.0em}
\end{table*}

\subsection{Model Scale}
\label{sec:model_scale}
We further examined the \textsc{Sep}-1c1t training strategy, 
which achieved the best performance on most test datasets, 
across different model scales using the \textsc{Qwen2.5} series.
The results are presented in Figure~\ref{fig:qwen2.5_scale}. 
As the model size increases from 3B to 14B, 
both d-BLEU and d-COMET exhibit a strictly increasing trend, 
indicating that our training strategy is broadly applicable to models of various sizes.

\subsection{Trade-off: Accuracy, Training Cost and Deployment Efficiency}
\label{sec:trade_off}
Although \textsc{Sep}-2c1t achieves the best overall accuracy~(Section~\ref{sec:exp}),
it requires three times of full-parameter fine-tuning and the deployment of three separate models,
resulting in substantial training and deployment overhead.
To investigate a more efficient alternative,
we replace full-parameter fine-tuning with LoRA 
\citep{hu2022lora} 
and analyze the trade-offs among translation quality,
training efficiency, and deployment cost.

We adopt LoRA\footnote{
We additionally evaluate a higher-rank configuration $r=64$.
Although a larger rank slightly improves in-distribution performance,
it leads to degradation on the OOD GlobVDoc benchmark.
Detailed results are provided in Appendix~\ref{sec:lora_results}.
} with $r=16$, $\alpha=32$, and a dropout rate of 0.05 on the \textsc{Tower}-7B model.
The results are shown in Table~\ref{tab:trade_off}.
While LoRA performs slightly worse than full-parameter fine-tuning overall, 
the performance gap remains small.
In particular, it yields substantially lower d-BLEU scores on IWSLT xx-en while maintaining competitive d-COMET, 
suggesting comparable translation quality despite reduced lexical overlap. 
Furthermore, LoRA surpasses full-parameter fine-tuning on the OOD GlobVDoc benchmark,
indicating improved robustness under distribution shifts.

Overall, the three configurations represent different points on the efficiency–performance trade-off. 
\textsc{Sep}-2c1t (LoRA) offers the best balance between effectiveness and efficiency, 
\textsc{Stair}-2c1t (Full) minimizes deployment complexity while maintaining strong performance, 
and \textsc{Sep}-2c1t (Full) achieves the highest accuracy at the cost of increased training and deployment overhead.

\subsection{Statistical Significance Analysis}
\label{sec:statistical-significance}

We assess the reliability of the d-BLEU differences using one-sided paired bootstrap resampling \citep{koehn-2004-statistical}, 
with documents as the sampling unit and 10,000 bootstrap samples. 
For each system pair, the alternative hypothesis is that the row system outperforms the column system. 
We adopt a significance threshold of $p<0.05$.

The results on IWSLT2017 are shown in Figure~\ref{fig:iwslt_stat}.
\textsc{Sep}-2c1t and \textsc{Stair}-2c1t significantly outperform \textsc{Stair}-FS4, \textsc{docFT}, 
and original model across both backbones and translation directions ($p\leq10^{-4}$) on IWSLT2017. 
The difference between \textsc{Sep}-2c1t and \textsc{Stair}-2c1t is significant only for
\textsc{Qwen2.5}-7B on en-xx ($p=0.004$), 
indicating that their relative advantage is not consistent across settings.
For \textsc{Tower}-7B, \textsc{Uni}-2c1t also significantly outperforms \textsc{Stair}-FS4 in both directions.
Overall, the results suggest the advantage of the 2c1t training strategies over the document-level and fixed-sized systems.
The results of the statistical significance tests on BWB and GlobVDoc 
are provided in Figure~\ref{fig:bwb_globv_stat} in the Appendix.

\section{Conclusion}
\label{sec:con}
In this study, 
to handle training-inference length discrepancies caused by document length variability, 
we introduced FRC, 
a DP-based strategy that unifies length intervals across training and testing.
Building upon FRC,
we conducted a systematic analysis of various training paradigms across diverse data formats,
finding that separate training exhibited superior stability on in-domain test data,
while offering a trade-off analysis among translation quality, training efficiency, and deployment cost.
Furthermore, we benchmarked our fine-tuned models against commercial LLMs
and agent-based methods on our self-established GlobVDoc dataset.
The results showed that our models surpassed \textsc{Qwen3-30B-A3B-Instruct} based agent methods in out-of-distribution performance.

\newpage
\section*{Limitations}
Although the GuoFeng dataset is widely used in the current DocMT task, 
we discovered only upon completing all training tasks that the DocBlocks dataset included GuoFeng's Test 1, Test 2, Valid 1, and Valid 2 data~\citep{xu-etal-2022-guofeng}.
However, some chapters of them overlap entirely with Test 3,
which was used in the WMT23 and WMT24 Discourse-Level Literary Translation shared tasks~\citep{wang-etal-2023-findings,wang-etal-2024-findings}.
While this overlap constitutes only a small portion, 
we elected to exclude the GuoFeng test data from our evaluation to ensure the integrity and stringency of our results.

Since \citet{ramos2025multilingual} did not release the parameters of their trained model,
faithfully reproducing their method is challenging. 
Consequently, in Section~\ref{sec:rslt_iwslt_bwb},
we compare our approach with theirs only in terms of the relative performance improvement over the corresponding baseline. 
However, we cannot guarantee that this comparison is fully equivalent.

In Section~\ref{sec:chunk_strategy},
we compare FRC with both FW and FN under the \textsc{Sep}-2c1t setting.
Although we controlled the experimental variables as much as possible by using an identical data construction pipeline that differed only in the chunking method,
and by training all models with the same hyperparameters and random seed, 
the resulting training information may still differ because intermediate processing steps,
such as chunk alignment and filtering. 
Furthermore, the \textsc{Sep}-2c1t setting may encourage homogeneous representations, 
thereby reducing the observable performance differences among the compared methods.


Prior research~\citep{kim-etal-2019-document,peng-etal-2025-self,obrien-etal-2025-dochplt,li2026crosspreferencelearningsentencelevelcontextaware} suggests that context selection can significantly influence model precision. 
Consequently, we leave the analysis of how to leverage FRC for optimized context selection as a subject for future work.

To ensure a fair comparison with existing methods, 
all the models undergo full-parameter fine-tuning, 
which incurs substantial computational costs.
Consequently, most of our analysis and comparative experiments are conducted on a single backbone model under a specific training setting~(e.g., \textsc{Tower} under \textsc{d2dFT}, \textsc{Sep}-1c1t).

\section*{Ethical Statement}
All models, datasets, and tools utilized in this study are publicly accessible and intended for research purposes. 

We ensure that all source material for the GlobVDoc Dataset,
sourced from Global Voices, 
carries a Creative Commons~(CC) license.
Comprehensive metadata, including titles, authors, and translators,
is recorded in Table~\ref{tab:globvdoc_docs} in the Appendix.

Regarding agent-based methods,
we employed GRAFT\footnote{\url{https://anonymous.4open.science/r/graft-docmt/README.md}}~\citep{dutta-etal-2025-graft}
and DELTA\footnote{\url{https://github.com/YutongWang1216/DocMTAgent}}~\citep{wang2025delta} in their original states without modification to the source code.

Furthermore, while the formulation of our proposed reference-based LTCR metric is informed by~\citet{lyu-etal-2021-encouraging}, 
the software implementation was developed independently by the authors.


\section*{Acknowledgements}
We would like to thank all of our co-authors and the team members who voluntarily contributed to the construction of the GlobVDoc dataset.

We are grateful to all the anonymous reviewers for their constructive feedback, which greatly improved this paper.
We also sincerely thank Adam Nohejl,
Daiki Matsuoka, Koki Ryu, Ryoma Kumon, Rongzhi Li, Taisei Yamamoto, and Tomoki Doi 
for their valuable discussions, suggestions, and insightful comments.

This work was supported by JST CREST Grant Number JPMJCR2565, Japan. 

\bibliography{custom, anthology}

@misc{sturua2024jinaembeddingsv3multilingualembeddingstask,
      title={jina-embeddings-v3: Multilingual Embeddings With Task LoRA}, 
      author={Saba Sturua and Isabelle Mohr and Mohammad Kalim Akram and Michael Günther and Bo Wang and Markus Krimmel and Feng Wang and Georgios Mastrapas and Andreas Koukounas and Andreas Koukounas and Nan Wang and Han Xiao},
      year={2024},
      eprint={2409.10173},
      archivePrefix={arXiv},
      primaryClass={cs.CL},
      url={https://arxiv.org/abs/2409.10173}, 
}

@inproceedings{finkelstein-etal-2024-introducing,
    title = "Introducing the {N}ews{P}a{LM} {MBR} and {QE} Dataset: {LLM}-Generated High-Quality Parallel Data Outperforms Traditional Web-Crawled Data",
    author = "Finkelstein, Mara  and
      Vilar, David  and
      Freitag, Markus",
    editor = "Haddow, Barry  and
      Kocmi, Tom  and
      Koehn, Philipp  and
      Monz, Christof",
    booktitle = "Proceedings of the Ninth Conference on Machine Translation",
    month = nov,
    year = "2024",
    address = "Miami, Florida, USA",
    publisher = "Association for Computational Linguistics",
    url = "https://aclanthology.org/2024.wmt-1.126/",
    doi = "10.18653/v1/2024.wmt-1.126",
    pages = "1355--1372"
}

@inproceedings{wang-etal-2023-document-level,
    title = "Document-Level Machine Translation with Large Language Models",
    author = "Wang, Longyue  and
      Lyu, Chenyang  and
      Ji, Tianbo  and
      Zhang, Zhirui  and
      Yu, Dian  and
      Shi, Shuming  and
      Tu, Zhaopeng",
    editor = "Bouamor, Houda  and
      Pino, Juan  and
      Bali, Kalika",
    booktitle = "Proceedings of the 2023 Conference on Empirical Methods in Natural Language Processing",
    month = dec,
    year = "2023",
    address = "Singapore",
    publisher = "Association for Computational Linguistics",
    url = "https://aclanthology.org/2023.emnlp-main.1036/",
    doi = "10.18653/v1/2023.emnlp-main.1036",
    pages = "16646--16661"
}

@inproceedings{peng-etal-2025-investigating,
    title = "Investigating Length Issues in Document-level Machine Translation",
    author = "Peng, Ziqian  and
      Bawden, Rachel  and
      Yvon, Fran{\c{c}}ois",
    editor = "Bouillon, Pierrette  and
      Gerlach, Johanna  and
      Girletti, Sabrina  and
      Volkart, Lise  and
      Rubino, Raphael  and
      Sennrich, Rico  and
      Farinha, Ana C.  and
      Gaido, Marco  and
      Daems, Joke  and
      Kenny, Dorothy  and
      Moniz, Helena  and
      Szoc, Sara",
    booktitle = "Proceedings of Machine Translation Summit XX: Volume 1",
    month = jun,
    year = "2025",
    address = "Geneva, Switzerland",
    publisher = "European Association for Machine Translation",
    url = "https://aclanthology.org/2025.mtsummit-1.3/",
    pages = "4--23",
    ISBN = "978-2-9701897-0-1"
}

@inproceedings{choudhary-etal-2025-exploring,
    title = "Exploring Context Strategies in {LLM}s for Discourse-Aware Machine Translation",
    author = "Choudhary, Ritvik  and
      Hida, Rem  and
      Hamada, Masaki  and
      Futami, Hayato  and
      Sekiya, Toshiyuki",
    editor = "Christodoulopoulos, Christos  and
      Chakraborty, Tanmoy  and
      Rose, Carolyn  and
      Peng, Violet",
    booktitle = "Findings of the Association for Computational Linguistics: EMNLP 2025",
    month = nov,
    year = "2025",
    address = "Suzhou, China",
    publisher = "Association for Computational Linguistics",
    url = "https://aclanthology.org/2025.findings-emnlp.1324/",
    doi = "10.18653/v1/2025.findings-emnlp.1324",
    pages = "24382--24391",
    ISBN = "979-8-89176-335-7"
}

@inproceedings{hu-etal-2025-source,
    title = "Source-primed Multi-turn Conversation Helps Large Language Models Translate Documents",
    author = "Hu, Hanxu  and
      Vamvas, Jannis  and
      Sennrich, Rico",
    editor = "Christodoulopoulos, Christos  and
      Chakraborty, Tanmoy  and
      Rose, Carolyn  and
      Peng, Violet",
    booktitle = "Findings of the Association for Computational Linguistics: EMNLP 2025",
    month = nov,
    year = "2025",
    address = "Suzhou, China",
    publisher = "Association for Computational Linguistics",
    url = "https://aclanthology.org/2025.findings-emnlp.1289/",
    doi = "10.18653/v1/2025.findings-emnlp.1289",
    pages = "23702--23712",
    ISBN = "979-8-89176-335-7"
}

@misc{wu2024adaptinglargelanguagemodels,
      title={Adapting Large Language Models for Document-Level Machine Translation}, 
      author={Minghao Wu and Thuy-Trang Vu and Lizhen Qu and George Foster and Gholamreza Haffari},
      year={2024},
      eprint={2401.06468},
      archivePrefix={arXiv},
      primaryClass={cs.CL},
      url={https://arxiv.org/abs/2401.06468}, 
}

@misc{nllbteam2022languageleftbehindscaling,
      title={No Language Left Behind: Scaling Human-Centered Machine Translation}, 
      author={{NLLB Team} and Marta R. Costa-jussà and James Cross and Onur Çelebi and Maha Elbayad and Kenneth Heafield and Kevin Heffernan and Elahe Kalbassi and Janice Lam and Daniel Licht and Jean Maillard and Anna Sun and Skyler Wang and Guillaume Wenzek and Al Youngblood and Bapi Akula and Loic Barrault and Gabriel Mejia Gonzalez and Prangthip Hansanti and John Hoffman and Semarley Jarrett and Kaushik Ram Sadagopan and Dirk Rowe and Shannon Spruit and Chau Tran and Pierre Andrews and Necip Fazil Ayan and Shruti Bhosale and Sergey Edunov and Angela Fan and Cynthia Gao and Vedanuj Goswami and Francisco Guzmán and Philipp Koehn and Alexandre Mourachko and Christophe Ropers and Safiyyah Saleem and Holger Schwenk and Jeff Wang},
      year={2022},
      eprint={2207.04672},
      archivePrefix={arXiv},
      primaryClass={cs.CL},
      url={https://arxiv.org/abs/2207.04672}, 
}

@misc{liu2025improvingllmbaseddocumentlevelmachine,
      title={Improving LLM-based Document-level Machine Translation with Multi-Knowledge Fusion}, 
      author={Bin Liu and Xinglin Lyu and Junhui Li and Daimeng Wei and Min Zhang and Shimin Tao and Hao Yang},
      year={2025},
      eprint={2503.12152},
      archivePrefix={arXiv},
      primaryClass={cs.CL},
      url={https://arxiv.org/abs/2503.12152}, 
}

@inproceedings{
wang2025delta,
title={Del{TA}: An Online Document-Level Translation Agent Based on Multi-Level Memory},
author={Yutong Wang and Jiali Zeng and Xuebo Liu and Derek F. Wong and Fandong Meng and Jie Zhou and Min Zhang},
booktitle={The Thirteenth International Conference on Learning Representations},
year={2025},
url={https://openreview.net/forum?id=hoYFLRNbhc}
}

@inproceedings{peng-etal-2025-self,
    title = "Self-Retrieval from Distant Contexts for Document-Level Machine Translation",
    author = "Peng, Ziqian  and
      Bawden, Rachel  and
      Yvon, Fran{\c{c}}ois",
    editor = "Haddow, Barry  and
      Kocmi, Tom  and
      Koehn, Philipp  and
      Monz, Christof",
    booktitle = "Proceedings of the Tenth Conference on Machine Translation",
    month = nov,
    year = "2025",
    address = "Suzhou, China",
    publisher = "Association for Computational Linguistics",
    url = "https://aclanthology.org/2025.wmt-1.13/",
    doi = "10.18653/v1/2025.wmt-1.13",
    pages = "220--240",
    ISBN = "979-8-89176-341-8"
}

@misc{pham2025discoursegraphguideddocument,
      title={Discourse Graph Guided Document Translation with Large Language Models}, 
      author={Viet-Thanh Pham and Minghan Wang and Hao-Han Liao and Thuy-Trang Vu},
      year={2025},
      eprint={2511.07230},
      archivePrefix={arXiv},
      primaryClass={cs.CL},
      url={https://arxiv.org/abs/2511.07230}, 
}

@article{Wu2024PerhapsBH,
  title={(Perhaps) Beyond Human Translation: Harnessing Multi-Agent Collaboration for Translating Ultra-Long Literary Texts},
  author={Minghao Wu and Yulin Yuan and Gholamreza Haffari and Longyue Wang},
  journal={Trans. Assoc. Comput. Linguistics},
  year={2024},
  volume={13},
  pages={901-922},
  url={https://api.semanticscholar.org/CorpusID:269921643}
}

@inproceedings{
ramos2025multilingual,
title={Multilingual Contextualization of Large Language Models for Document-Level Machine Translation},
author={Miguel Moura Ramos and Patrick Fernandes and Sweta Agrawal and Andre Martins},
booktitle={Second Conference on Language Modeling},
year={2025},
url={https://openreview.net/forum?id=Ah0U1r5Ldq}
}

@inproceedings{raunak-etal-2024-slide,
    title = "{SLIDE}: Reference-free Evaluation for Machine Translation using a Sliding Document Window",
    author = "Raunak, Vikas  and
      Kocmi, Tom  and
      Post, Matt",
    editor = "Duh, Kevin  and
      Gomez, Helena  and
      Bethard, Steven",
    booktitle = "Proceedings of the 2024 Conference of the North American Chapter of the Association for Computational Linguistics: Human Language Technologies (Volume 2: Short Papers)",
    month = jun,
    year = "2024",
    address = "Mexico City, Mexico",
    publisher = "Association for Computational Linguistics",
    url = "https://aclanthology.org/2024.naacl-short.18/",
    doi = "10.18653/v1/2024.naacl-short.18",
    pages = "205--211"
}

@misc{gu2025surveyllmasajudge,
      title={A Survey on LLM-as-a-Judge}, 
      author={Jiawei Gu and Xuhui Jiang and Zhichao Shi and Hexiang Tan and Xuehao Zhai and Chengjin Xu and Wei Li and Yinghan Shen and Shengjie Ma and Honghao Liu and Saizhuo Wang and Kun Zhang and Yuanzhuo Wang and Wen Gao and Lionel Ni and Jian Guo},
      year={2025},
      eprint={2411.15594},
      archivePrefix={arXiv},
      primaryClass={cs.CL},
      url={https://arxiv.org/abs/2411.15594}, 
}

@misc{
zhu2024judgelm,
title={Judge{LM} : Fine-tuned Large Language Models are Scalable Judges},
author={Lianghui Zhu and Xinggang Wang and Xinlong Wang},
year={2024},
url={https://openreview.net/forum?id=87YOFayjcG}
}

@inproceedings{kocmi-federmann-2023-gemba,
    title = "{GEMBA}-{MQM}: Detecting Translation Quality Error Spans with {GPT}-4",
    author = "Kocmi, Tom  and
      Federmann, Christian",
    editor = "Koehn, Philipp  and
      Haddow, Barry  and
      Kocmi, Tom  and
      Monz, Christof",
    booktitle = "Proceedings of the Eighth Conference on Machine Translation",
    month = dec,
    year = "2023",
    address = "Singapore",
    publisher = "Association for Computational Linguistics",
    url = "https://aclanthology.org/2023.wmt-1.64/",
    doi = "10.18653/v1/2023.wmt-1.64",
    pages = "768--775"
}

@misc{mrozinski2025qualityestimationrerankingdocumentlevel,
      title={Quality Estimation Reranking for Document-Level Translation}, 
      author={Krzysztof Mrozinski and Minji Kang and Ahmed Khota and Vincent Michael Sutanto and Giovanni Gatti De Giacomo},
      year={2025},
      eprint={2510.08870},
      archivePrefix={arXiv},
      primaryClass={cs.CL},
      url={https://arxiv.org/abs/2510.08870}, 
}

@article{Needleman1970AGM,
  title={A general method applicable to the search for similarities in the amino acid sequence of two proteins.},
  author={Saul B. Needleman and Christian D. Wunsch},
  journal={Journal of molecular biology},
  year={1970},
  volume={48 3},
  pages={
          443-53
        },
  url={https://api.semanticscholar.org/CorpusID:17406543}
}

@inproceedings{qi2020stanza,
    title={Stanza: A {Python} Natural Language Processing Toolkit for Many Human Languages},
    author={Qi, Peng and Zhang, Yuhao and Zhang, Yuhui and Bolton, Jason and Manning, Christopher D.},
    booktitle = "Proceedings of the 58th Annual Meeting of the Association for Computational Linguistics: System Demonstrations",
    year={2020}
}

@inproceedings{wang-etal-2023-findings,
    title = "Findings of the {WMT} 2023 Shared Task on Discourse-Level Literary Translation: A Fresh Orb in the Cosmos of {LLM}s",
    author = "Wang, Longyue  and
      Tu, Zhaopeng  and
      Gu, Yan  and
      Liu, Siyou  and
      Yu, Dian  and
      Ma, Qingsong  and
      Lyu, Chenyang  and
      Zhou, Liting  and
      Liu, Chao-Hong  and
      Ma, Yufeng  and
      Chen, Weiyu  and
      Graham, Yvette  and
      Webber, Bonnie  and
      Koehn, Philipp  and
      Way, Andy  and
      Yuan, Yulin  and
      Shi, Shuming",
    editor = "Koehn, Philipp  and
      Haddow, Barry  and
      Kocmi, Tom  and
      Monz, Christof",
    booktitle = "Proceedings of the Eighth Conference on Machine Translation",
    month = dec,
    year = "2023",
    address = "Singapore",
    publisher = "Association for Computational Linguistics",
    url = "https://aclanthology.org/2023.wmt-1.3/",
    doi = "10.18653/v1/2023.wmt-1.3",
    pages = "55--67"
}

@inproceedings{alves2024tower,
title={Tower: An Open Multilingual Large Language Model for Translation-Related Tasks},
author={Duarte Miguel Alves and Jos{\'e} Pombal and Nuno M Guerreiro and Pedro Henrique Martins and Jo{\~a}o Alves and Amin Farajian and Ben Peters and Ricardo Rei and Patrick Fernandes and Sweta Agrawal and Pierre Colombo and Jos{\'e} G. C. de Souza and Andre Martins},
booktitle={First Conference on Language Modeling},
year={2024},
url={https://openreview.net/forum?id=EHPns3hVkj}
}

@inproceedings{rei-etal-2023-scaling,
    title = "Scaling up {C}omet{K}iwi: Unbabel-{IST} 2023 Submission for the Quality Estimation Shared Task",
    author = "Rei, Ricardo  and
      Guerreiro, Nuno M.  and
      Pombal, Jos{\~A}{\textcopyright}  and
      van Stigt, Daan  and
      Treviso, Marcos  and
      Coheur, Luisa  and
      C. de Souza, Jos{\'e} G.  and
      Martins, Andr{\'e}",
    editor = "Koehn, Philipp  and
      Haddow, Barry  and
      Kocmi, Tom  and
      Monz, Christof",
    booktitle = "Proceedings of the Eighth Conference on Machine Translation",
    month = dec,
    year = "2023",
    address = "Singapore",
    publisher = "Association for Computational Linguistics",
    url = "https://aclanthology.org/2023.wmt-1.73/",
    doi = "10.18653/v1/2023.wmt-1.73",
    pages = "841--848"
}

@inproceedings{wang-etal-2024-findings,
    title = "Findings of the {WMT} 2024 Shared Task on Discourse-Level Literary Translation",
    author = "Wang, Longyue  and
      Liu, Siyou  and
      Lyu, Chenyang  and
      Jiao, Wenxiang  and
      Wang, Xing  and
      Xu, Jiahao  and
      Tu, Zhaopeng  and
      Gu, Yan  and
      Chen, Weiyu  and
      Wu, Minghao  and
      Zhou, Liting  and
      Koehn, Philipp  and
      Way, Andy  and
      Yuan, Yulin",
    editor = "Haddow, Barry  and
      Kocmi, Tom  and
      Koehn, Philipp  and
      Monz, Christof",
    booktitle = "Proceedings of the Ninth Conference on Machine Translation",
    month = nov,
    year = "2024",
    address = "Miami, Florida, USA",
    publisher = "Association for Computational Linguistics",
    url = "https://aclanthology.org/2024.wmt-1.58/",
    doi = "10.18653/v1/2024.wmt-1.58",
    pages = "699--700"
}

@inproceedings{
loshchilov2018decoupled,
title={Decoupled Weight Decay Regularization},
author={Ilya Loshchilov and Frank Hutter},
booktitle={International Conference on Learning Representations},
year={2019},
url={https://openreview.net/forum?id=Bkg6RiCqY7},
}

@misc{openai2024gpt4technicalreport,
      title={GPT-4 Technical Report}, 
      author={{OpenAI Team}},
      year={2024},
      eprint={2303.08774},
      archivePrefix={arXiv},
      primaryClass={cs.CL},
      url={https://arxiv.org/abs/2303.08774}, 
}

@misc{comanici2025gemini25pushingfrontier,
      title={Gemini 2.5: Pushing the Frontier with Advanced Reasoning, Multimodality, Long Context, and Next Generation Agentic Capabilities}, 
      author={{Gemini Team}},
      year={2025},
      eprint={2507.06261},
      archivePrefix={arXiv},
      primaryClass={cs.CL},
      url={https://arxiv.org/abs/2507.06261}, 
}

@misc{deepseekai2025deepseekv3technicalreport,
      title={DeepSeek-V3 Technical Report}, 
      author={DeepSeek-AI and Aixin Liu and  Bei Feng and Bing Xue and Bingxuan Wang and Bochao Wu and Chengda Lu and Chenggang Zhao and Chengqi Deng and Chenyu
Zhang and Chong Ruan and others},
      year={2025},
      eprint={2412.19437},
      archivePrefix={arXiv},
      primaryClass={cs.CL},
      url={https://arxiv.org/abs/2412.19437}, 
}

@misc{zhao2025metachunkinglearningtextsegmentation,
      title={Meta-Chunking: Learning Text Segmentation and Semantic Completion via Logical Perception}, 
      author={Jihao Zhao and Zhiyuan Ji and Yuchen Feng and Pengnian Qi and Simin Niu and Bo Tang and Feiyu Xiong and Zhiyu Li},
      year={2025},
      eprint={2410.12788},
      archivePrefix={arXiv},
      primaryClass={cs.CL},
      url={https://arxiv.org/abs/2410.12788}, 
}

@misc{wang2024multilinguale5textembeddings,
      title={Multilingual E5 Text Embeddings: A Technical Report}, 
      author={Liang Wang and Nan Yang and Xiaolong Huang and Linjun Yang and Rangan Majumder and Furu Wei},
      year={2024},
      eprint={2402.05672},
      archivePrefix={arXiv},
      primaryClass={cs.CL},
      url={https://arxiv.org/abs/2402.05672}, 
}

@inproceedings{dutta-etal-2025-graft,
    title = "{GRAFT}: A Graph-based Flow-aware Agentic Framework for Document-level Machine Translation",
    author = "Dutta, Himanshu  and
      Manchanda, Sunny  and
      Bapat, Prakhar  and
      Gurjar, Meva Ram  and
      Bhattacharyya, Pushpak",
    editor = "Potdar, Saloni  and
      Rojas-Barahona, Lina  and
      Montella, Sebastien",
    booktitle = "Proceedings of the 2025 Conference on Empirical Methods in Natural Language Processing: Industry Track",
    month = nov,
    year = "2025",
    address = "Suzhou (China)",
    publisher = "Association for Computational Linguistics",
    url = "https://aclanthology.org/2025.emnlp-industry.166/",
    doi = "10.18653/v1/2025.emnlp-industry.166",
    pages = "2405--2428",
    ISBN = "979-8-89176-333-3"
}

@misc{mikhaylovskiy2023autocorrelationsdecaytextsapplicability,
      title={Autocorrelations Decay in Texts and Applicability Limits of Language Models}, 
      author={Nikolay Mikhaylovskiy and Ilya Churilov},
      year={2023},
      eprint={2305.06615},
      archivePrefix={arXiv},
      primaryClass={cs.CL},
      url={https://arxiv.org/abs/2305.06615}, 
}

@inproceedings{kwon2023efficient,
  title={Efficient Memory Management for Large Language Model Serving with PagedAttention},
  author={Woosuk Kwon and Zhuohan Li and Siyuan Zhuang and Ying Sheng and Lianmin Zheng and Cody Hao Yu and Joseph E. Gonzalez and Hao Zhang and Ion Stoica},
  booktitle={Proceedings of the ACM SIGOPS 29th Symposium on Operating Systems Principles},
  year={2023}
}

@inproceedings{obrien-etal-2025-dochplt,
    title = "{D}oc{HPLT}: A Massively Multilingual Document-Level Translation Dataset",
    author = {O{'}Brien, Dayy{\'a}n  and
      Malik, Bhavitvya  and
      de Gibert, Ona  and
      Chen, Pinzhen  and
      Haddow, Barry  and
      Tiedemann, J{\"o}rg},
    editor = "Haddow, Barry  and
      Kocmi, Tom  and
      Koehn, Philipp  and
      Monz, Christof",
    booktitle = "Proceedings of the Tenth Conference on Machine Translation",
    month = nov,
    year = "2025",
    address = "Suzhou, China",
    publisher = "Association for Computational Linguistics",
    url = "https://aclanthology.org/2025.wmt-1.17/",
    doi = "10.18653/v1/2025.wmt-1.17",
    pages = "286--300",
    ISBN = "979-8-89176-341-8"
}

@inproceedings{wicks-etal-2024-recovering,
    title = "Recovering document annotations for sentence-level bitext",
    author = "Wicks, Rachel  and
      Post, Matt  and
      Koehn, Philipp",
    editor = "Ku, Lun-Wei  and
      Martins, Andre  and
      Srikumar, Vivek",
    booktitle = "Findings of the Association for Computational Linguistics: ACL 2024",
    month = aug,
    year = "2024",
    address = "Bangkok, Thailand",
    publisher = "Association for Computational Linguistics",
    url = "https://aclanthology.org/2024.findings-acl.589/",
    doi = "10.18653/v1/2024.findings-acl.589",
    pages = "9876--9890"
}

@misc{jin2024chaptertochaptercontextawareliterarytranslation,
      title={Towards Chapter-to-Chapter Context-Aware Literary Translation via Large Language Models}, 
      author={Linghao Jin and Li An and Xuezhe Ma},
      year={2024},
      eprint={2407.08978},
      archivePrefix={arXiv},
      primaryClass={cs.CL},
      url={https://arxiv.org/abs/2407.08978}, 
}

@inproceedings{alabi-etal-2025-afridoc,
    title = "{AFRIDOC}-{MT}: Document-level {MT} Corpus for {A}frican Languages",
    author = "Alabi, Jesujoba Oluwadara  and
      Azime, Israel Abebe  and
      Zhang, Miaoran  and
      Espa{\~n}a-Bonet, Cristina  and
      Bawden, Rachel  and
      Zhu, Dawei  and
      Adelani, David Ifeoluwa  and
      Odoje, Clement Oyeleke  and
      Akinade, Idris  and
      Maab, Iffat  and
      David, Davis  and
      Muhammad, Shamsuddeen Hassan  and
      Putini, Neo  and
      Ademuyiwa, David O.  and
      Caines, Andrew  and
      Klakow, Dietrich",
    editor = "Christodoulopoulos, Christos  and
      Chakraborty, Tanmoy  and
      Rose, Carolyn  and
      Peng, Violet",
    booktitle = "Proceedings of the 2025 Conference on Empirical Methods in Natural Language Processing",
    month = nov,
    year = "2025",
    address = "Suzhou, China",
    publisher = "Association for Computational Linguistics",
    url = "https://aclanthology.org/2025.emnlp-main.1413/",
    doi = "10.18653/v1/2025.emnlp-main.1413",
    pages = "27758--27794",
    ISBN = "979-8-89176-332-6"
}

@misc{guo2025docguidedsent2sentsent2sentagent,
      title={Doc-Guided Sent2Sent++: A Sent2Sent++ Agent with Doc-Guided memory for Document-level Machine Translation}, 
      author={Jiaxin Guo and Yuanchang Luo and Daimeng Wei and Ling Zhang and Zongyao Li and Hengchao Shang and Zhiqiang Rao and Shaojun Li and Jinlong Yang and Zhanglin Wu and Hao Yang},
      year={2025},
      eprint={2501.08523},
      archivePrefix={arXiv},
      primaryClass={cs.CL},
      url={https://arxiv.org/abs/2501.08523}, 
}

@inproceedings{
xu2024a,
title={A Paradigm Shift in Machine Translation: Boosting Translation Performance of Large Language Models},
author={Haoran Xu and Young Jin Kim and Amr Sharaf and Hany Hassan Awadalla},
booktitle={The Twelfth International Conference on Learning Representations},
year={2024},
url={https://openreview.net/forum?id=farT6XXntP}
}

@inproceedings{guo-etal-2024-novel,
    title = "A Novel Paradigm Boosting Translation Capabilities of Large Language Models",
    author = "Guo, Jiaxin  and
      Yang, Hao  and
      Li, Zongyao  and
      Wei, Daimeng  and
      Shang, Hengchao  and
      Chen, Xiaoyu",
    editor = "Duh, Kevin  and
      Gomez, Helena  and
      Bethard, Steven",
    booktitle = "Findings of the Association for Computational Linguistics: NAACL 2024",
    month = jun,
    year = "2024",
    address = "Mexico City, Mexico",
    publisher = "Association for Computational Linguistics",
    url = "https://aclanthology.org/2024.findings-naacl.42/",
    doi = "10.18653/v1/2024.findings-naacl.42",
    pages = "639--649"
}

@inproceedings{karpinska-iyyer-2023-large,
    title = "Large Language Models Effectively Leverage Document-level Context for Literary Translation, but Critical Errors Persist",
    author = "Karpinska, Marzena  and
      Iyyer, Mohit",
    editor = "Koehn, Philipp  and
      Haddow, Barry  and
      Kocmi, Tom  and
      Monz, Christof",
    booktitle = "Proceedings of the Eighth Conference on Machine Translation",
    month = dec,
    year = "2023",
    address = "Singapore",
    publisher = "Association for Computational Linguistics",
    url = "https://aclanthology.org/2023.wmt-1.41/",
    doi = "10.18653/v1/2023.wmt-1.41",
    pages = "419--451"
}

@inproceedings{li-etal-2024-towards-demonstration,
    title = "Towards Demonstration-Aware Large Language Models for Machine Translation",
    author = "Li, Chen  and
      Zhang, Meishan  and
      Liu, Xuebo  and
      Li, Zhaocong  and
      Wong, Derek  and
      Zhang, Min",
    editor = "Ku, Lun-Wei  and
      Martins, Andre  and
      Srikumar, Vivek",
    booktitle = "Findings of the Association for Computational Linguistics: ACL 2024",
    month = aug,
    year = "2024",
    address = "Bangkok, Thailand",
    publisher = "Association for Computational Linguistics",
    url = "https://aclanthology.org/2024.findings-acl.824/",
    doi = "10.18653/v1/2024.findings-acl.824",
    pages = "13868--13881"
}

@misc{hendy2023goodgptmodelsmachine,
      title={How Good Are GPT Models at Machine Translation? A Comprehensive Evaluation}, 
      author={Amr Hendy and Mohamed Abdelrehim and Amr Sharaf and Vikas Raunak and Mohamed Gabr and Hitokazu Matsushita and Young Jin Kim and Mohamed Afify and Hany Hassan Awadalla},
      year={2023},
      eprint={2302.09210},
      archivePrefix={arXiv},
      primaryClass={cs.CL},
      url={https://arxiv.org/abs/2302.09210}, 
}

@misc{yang2025hallucinatelongresponsegeneration,
      title={Hallucinate at the Last in Long Response Generation: A Case Study on Long Document Summarization}, 
      author={Joonho Yang and Seunghyun Yoon and Hwan Chang and Byeongjeong Kim and Hwanhee Lee},
      year={2025},
      eprint={2505.15291},
      archivePrefix={arXiv},
      primaryClass={cs.CL},
      url={https://arxiv.org/abs/2505.15291}, 
}

@misc{chen2023extendingcontextwindowlarge,
      title={Extending Context Window of Large Language Models via Positional Interpolation}, 
      author={Shouyuan Chen and Sherman Wong and Liangjian Chen and Yuandong Tian},
      year={2023},
      eprint={2306.15595},
      archivePrefix={arXiv},
      primaryClass={cs.CL},
      url={https://arxiv.org/abs/2306.15595}, 
}

@misc{li2026crosspreferencelearningsentencelevelcontextaware,
      title={Cross-Preference Learning for Sentence-Level and Context-Aware Machine Translation}, 
      author={Ying Li and Xinglin Lyu and Junhui Li and Jinlong Yang and Hengchao Shang and Min Zhang and Shimin Tao and Daimeng Wei},
      year={2026},
      eprint={2603.25183},
      archivePrefix={arXiv},
      primaryClass={cs.CL},
      url={https://arxiv.org/abs/2603.25183}, 
}

@inproceedings{hiraoka-inui-2025-repetition,
    title = "Repetition Neurons: How Do Language Models Produce Repetitions?",
    author = "Hiraoka, Tatsuya  and
      Inui, Kentaro",
    editor = "Chiruzzo, Luis  and
      Ritter, Alan  and
      Wang, Lu",
    booktitle = "Proceedings of the 2025 Conference of the Nations of the Americas Chapter of the Association for Computational Linguistics: Human Language Technologies (Volume 2: Short Papers)",
    month = apr,
    year = "2025",
    address = "Albuquerque, New Mexico",
    publisher = "Association for Computational Linguistics",
    url = "https://aclanthology.org/2025.naacl-short.41/",
    doi = "10.18653/v1/2025.naacl-short.41",
    pages = "483--495",
    ISBN = "979-8-89176-190-2"
}

@inproceedings{pitorro-etal-2024-effective,
    title = "How Effective Are State Space Models for Machine Translation?",
    author = "Pitorro, Hugo  and
      Vasylenko, Pavlo  and
      Treviso, Marcos  and
      Martins, Andr{\'e}",
    editor = "Haddow, Barry  and
      Kocmi, Tom  and
      Koehn, Philipp  and
      Monz, Christof",
    booktitle = "Proceedings of the Ninth Conference on Machine Translation",
    month = nov,
    year = "2024",
    address = "Miami, Florida, USA",
    publisher = "Association for Computational Linguistics",
    url = "https://aclanthology.org/2024.wmt-1.111/",
    doi = "10.18653/v1/2024.wmt-1.111",
    pages = "1107--1124"
}

@inproceedings{
hu2022lora,
title={Lo{RA}: Low-Rank Adaptation of Large Language Models},
author={Edward J Hu and yelong shen and Phillip Wallis and Zeyuan Allen-Zhu and Yuanzhi Li and Shean Wang and Lu Wang and Weizhu Chen},
booktitle={International Conference on Learning Representations},
year={2022},
url={https://openreview.net/forum?id=nZeVKeeFYf9}
}

@inproceedings{wang-etal-2025-bimax,
    title = "{B}i{M}ax: Bidirectional {M}ax{S}im Score for Document-Level Alignment",
    author = "Wang, Xiaotian  and
      Utsuro, Takehito  and
      Nagata, Masaaki",
    editor = "Christodoulopoulos, Christos  and
      Chakraborty, Tanmoy  and
      Rose, Carolyn  and
      Peng, Violet",
    booktitle = "Findings of the Association for Computational Linguistics: EMNLP 2025",
    month = nov,
    year = "2025",
    address = "Suzhou, China",
    publisher = "Association for Computational Linguistics",
    url = "https://aclanthology.org/2025.findings-emnlp.704/",
    doi = "10.18653/v1/2025.findings-emnlp.704",
    pages = "13095--13116",
    ISBN = "979-8-89176-335-7"
}

@misc{qwen2.5,
    title = {Qwen2.5: A Party of Foundation Models},
    url = {https://qwenlm.github.io/blog/qwen2.5/},
    author = {{Qwen Team}},
    month = {September},
    year = {2024}
}

\appendix

\section{Evaluation Procedures}
\label{sec:eval_details}
\paragraph{d-BLEU} The computation of the d-BLEU score is contingent upon the translation direction. For the ``xx-en'' direction, 
the sacreBLEU~\citep{post-2018-call} ``13a'' tokenization scheme is uniformly applied across the entire corpus.
In contrast, for the ``en-xx'' direction, 
d-BLEU scores are first computed individually for each target language and subsequently averaged. 
The tokenization method for the target language is specified as ``zh'' for Chinese, ``ko-mecab'' for Korean, and ``13a'' for all other languages.

\paragraph{d-COMET} 
Although many studies on DocMT use the contextual-embedding COMET~\citep{rei-etal-2022-comet, vernikos-etal-2022-embarrassingly} as their evaluation method,
the metric essentially remains sentence-level and incorporates additional contextual information. 
Therefore, we follow \citet{ramos2025multilingual} 
and compute d-COMET\footnote{\url{https://huggingface.co/Unbabel/wmt22-comet-da}}
at the chunk-level using SLIDE~\citep{raunak-etal-2024-slide}.
Since \citet{ramos2025multilingual} relied solely on overlapping 512-token chunks to compute chunk-level scores independently,
which may result in misalignment among the source, hypothesis, and reference,
we construct three types of alignment-oriented chunk triplets $(src, hyp, ref)$,
and across all strategies, we maintain a 50\% overlap between consecutive chunks:
\begin{itemize}
    \item \textbf{Src–Ref Aligned Stacking}: Source segments and their corresponding reference segments are concatenated until one sequence approaches the 512-token limit. Subsequently, the hypothesis is segmented and stacked to approximate the reference length, ensuring that the total count remains under 512 tokens.
    \item \textbf{Segment Independent Stacking}: Source, hypothesis, and reference segments are concatenated independently until each sequence reaches its maximum capacity of 512 tokens, disregarding cross-sequence alignment.
    \item \textbf{Token Independent Stacking}: A 512-token sliding window is applied separately to the source, hypothesis, and reference sequences, ignoring segment boundaries and alignment.
\end{itemize}
The final d-COMET score is computed as the average of the scores derived from these three distinct segmentation methods.

\paragraph{GEMBA-DA} 
GEMBA-DA~\citep{kocmi-federmann-2023-large} is an LLM-as-a-judge method~\citep{zhu2024judgelm, gu2025surveyllmasajudge}. 
We use GPT-4.1 as the judge model and set the temperature to 0 for deterministic evaluation.
By conditioning on the source, hypothesis, and reference, the model conducts a direct assessment of translation quality at the document level,
and the final score is obtained by averaging scores over the test set.
The prompt used for GEMBA-DA evaluation is provided in Appendix~\ref{sec:prompts}.

\section{Interest Word Selection for LTCR}
\label{sec:ltcr-interest-words}

Interest words are defined as named entities extracted from the reference text using spaCy\footnote{\url{https://spacy.io/}}.
We retain entities corresponding to the following labels: \textsc{Person}, \textsc{Per}, \textsc{Ps}, \textsc{Org}, \textsc{Og}, \textsc{Gpe}, \textsc{Norp}, \textsc{Fac}, \textsc{Loc}, \textsc{Lc}, \textsc{Misc}, \textsc{Product}, \textsc{Event}, \textsc{Work\_of\_Art}, \textsc{Law}, and \textsc{Language}.
Only entities appearing at least twice in the reference document are considered for consistency evaluation.

\begin{table*}[hbpt]
\centering
\small
\setlength{\tabcolsep}{10pt}
\begin{tabular}{ccccccr}
\toprule
\textbf{Domain} & \textbf{Source} & \textbf{Language Pair} & \textbf{|D|} & \textbf{|S|} & \textbf{|W|} & \textbf{|W|/|D|} \\
\midrule
\multirow{10}{*}{\makecell[c]{Global \\ Voices}} 
& \multirow{10}{*}{GlobVDoc}
& En $\rightarrow$ De & 10 & 458 & 10.4K / 10.8K & 1,038 / 1,080 \\
&  & En $\rightarrow$ Es & 10 & 596 & 11.6K / 12.9K & 1,162 / 1,292 \\
&  & En $\rightarrow$ Fr & 10 & 489 & 11.2K / 13.4K & 1,118 / 1,342 \\
&  & En $\rightarrow$ It & 10 & 584 & 11.9K / 13.2K & 1,195 / 1,316 \\
&  & En $\rightarrow$ Ko & 10 & 487 & 10.8K / 21.7K & 1,082 / 2,172 \\
&  & En $\rightarrow$ Nl & 10 & 492 & 9.6K / 9.9K & 963 / 989 \\
&  & En $\rightarrow$ Pt & 10 & 529 & 10.9K / 11.6K & 1,088 / 1,164 \\
&  & En $\rightarrow$ Ru & 10 & 492 & 9.4K / 7.8K & 942 / 780 \\
&  & En $\rightarrow$ Zh & 10 & 522 & 12.3K / 19.4K & 1,233 / 1,944 \\
\cmidrule(lr){3-7}
&  & En $\rightarrow$ Xx & 90 & 4,649 & 98.2K / 120.8K & 1,091 / 1,342 \\

\bottomrule
\end{tabular}
\caption{Statistics of GlobVDoc Dataset.
|D| represents the document count, 
|S| represents the total sentence count, 
and |W| indicates the total number of words in both the source language~(English) 
and the corresponding target language~(the second language listed in the Language Pair).}
\label{tab:globvdoc_stat}
\end{table*}

\begin{table*}[t]
\centering
\small
\setlength{\tabcolsep}{8pt}
\begin{tabular}{ccccccc}
\toprule
\textbf{Domain} & \textbf{Source} & \textbf{Language Pair} & \textbf{|D|} & \textbf{|S|} & \textbf{|W|} & \textbf{|W|/|D|} \\
\midrule

News
& \multirow{2}{*}{WMT2022}
& Zh $\rightarrow$ En
& 38
& 505
& 18.5K
& 487 \\

Social
& 
& Zh $\rightarrow$ En
& 25
& 478
& 13.3K
& 532 \\

\midrule

TED
& IWSLT2017
& En $\leftrightarrow$ Xx
& 374
& 37,082
& 615.8K
& 1,646 \\

\midrule

News
& News Commentary v11
& En $\rightarrow$ De
& 155
& 2,999
& 56.8K
& 366 \\

\midrule

Europarl
& Europarl v7
& En $\rightarrow$ De
& 360
& 5,134
& 130.1K
& 361 \\

\midrule

\multirow{5}{*}{\makecell[c]{Web\\Novel}}
& GuoFeng Valid1
& Zh $\rightarrow$ En
& 22
& 755
& 18.3K
& 832 \\

&
GuoFeng Test1
& Zh $\rightarrow$ En
& 22
& 697
& 19.5K
& 884 \\

&
GuoFeng Valid2
& Zh $\rightarrow$ En
& 10
& 853
& 16.0K
& 1.6K \\

&
GuoFeng Test2
& Zh $\rightarrow$ En
& 12
& 917
& 16.7K
& 1.4K \\

&
BWB
& Zh $\rightarrow$ En
& 80
& 2,633
& 58.5K
& 731 \\

\midrule

Global Voices
& GlobVDoc (Our)
& En $\rightarrow$ Xx
& 90
& 4,649
& 98.2K
& 1,092 \\

\bottomrule
\end{tabular}

\caption{Statistics of commonly used document-level test datasets.
|D| denotes the number of documents,
|S| denotes the number of sentences,
|W| denotes the total number of words in English,
and |W|/|D| denotes the average number of words per document.
For IWSLT2017, the statistics reported in the table only include en $\leftrightarrow$ \{de, fr, it, ko, nl, zh\} language pairs used in this work.}
\label{tab:dataset_statistics}
\end{table*}

\section{Details of GlobVDoc}
\label{sec:appen_globvdoc}
All contributors hold at least a Master’s degree and are engaged in research related to natural language processing,
specializing in areas such as machine translation, linguistics, and interpretability.
The team consisted of native speakers of English, Chinese, Spanish, and Dutch.
For sentence alignment involving non-native languages,
we conducted cross-verification using multiple high-precision translation tools and multilingual dictionaries.
The team members are instructed to follow a standardized formatting protocol: 
the article title is placed in the first row, 
the introduction in the second, and the main body from the third row.
Moreover, Global Voices articles that relied heavily on multimedia elements, 
such as videos, images, or audio,
are excluded during the document compilation process to ensure textual consistency.
In addition, 
while Global Voices features interview-based articles that may contain personal details, 
such information is publicly available and does not constitute sensitive or privacy-infringing data.
Each document pair is checked by at least two volunteers, 
and disagreements are discussed and resolved collectively. 

The detailed guidelines for data construction are provided below:
\begin{enumerate}
\item
\textbf{Source Selection and Diversity}: English was designated as the source language. 
To ensure the diversity of the test data and avoid redundancy, each unique English document was utilized only once.
\item
\textbf{Temporal Relevance}: Priority was given to the most recent documents to ensure the timeliness and contemporary relevance of the data.
\item
\textbf{Alignment Quality}: We selected documents where sentence-level alignment was as complete as possible. 
Texts exhibiting significant alignment gaps or structural discrepancies were excluded in favor of more consistent alternatives.
\item
\textbf{Structural Consistency}: Preference was given to translations that maintained a structural sequence consistent with the source text.
This avoided the alignment complexities often introduced by translations that radically reorder sentences within a paragraph.
\item
\textbf{Manual Segmentation}: Sentence segmentation was performed manually.
Rather than strictly decomposing texts into the smallest possible fragments, 
we merged related short segments to preserve discourse integrity.
\item
\textbf{Minimal Adjustments}: In cases of minor content misalignment~(e.g., where a translator added brief explanatory notes), 
slight adjustments were made while strictly respecting the original intent. 
No other modifications to the content were permitted.
\item
\textbf{Mapping Priority}: In instances of many-to-many or many-to-one sentence mappings,
priority was given to establishing complete semantic correspondence.
\end{enumerate}

Detailed counts of documents, sentences, and words in the GlobVDoc dataset are summarized across different language pairs in Table~\ref{tab:globvdoc_stat}.
The specifications of the data utilized to construct the dataset,
including document IDs, source and translated titles, source and target languages, 
and the names of authors and translators,
are presented in Table~\ref{tab:globvdoc_docs}.

Table~\ref{tab:dataset_statistics} compares the statistics of GlobVDoc with those of other commonly used document-level machine translation test datasets.
We compute the statistics for the IWSLT2017 and BWB datasets used in this work, 
while the statistics for the remaining datasets refer to \citet{wang-etal-2023-document-level, wang-etal-2024-findings}. 
We do not include GuoFeng Test~3, 
which was used in the WMT23 and WMT24 Discourse-Level Literary Translation shared tasks, 
because its reference translations are not publicly available at the time of this work.
As shown in the table, 
the scale of GlobVDoc is comparable to existing commonly used document-level evaluation datasets.

\section{Quality Estimation of Test Data}
\label{sec:qe_test}
The intrinsic quality of the test data warrants careful examination.
Following the word-level definition of autocorrelations using distributional semantics proposed by~\citet{mikhaylovskiy2023autocorrelationsdecaytextsapplicability},
we define a document-level non-centered autocorrelation function and further extend it to the corpus level. 

Given a document $d\in \mathcal{D}$ consisting of units $\{u_1, \ldots, u_{T_d}\}$,
the corpus autocorrelation function at lag $k$, denoted as $R[k]$, is defined as: 

\begin{equation}
R[k] =
\frac{
\sum_{\mathrm{d}} \sum_{t=k}^{T_d} \operatorname{sim}(u_t, u_{t-k}) \cdot \left(1 - \frac{k}{T_d}\right)
}{
\sum_{\mathrm{d}} (T_d - k) \cdot \left(1 - \frac{k}{T_d}\right)
}
\end{equation}
where the term $(1 - k / T_d)$ serves as a length-penalty weight to mitigate the impact of large variations in document length within the corpus 
(e.g., in some documents, $u_1$ and $u_4$ are still located near the beginning of the document, whereas in others $u_4$ already appears at the end).
We then take a weighted average of the first four lags as the autocorrelation coefficient of the data, referred to as d-ACF:
\begin{equation}
\text{d-}\mathrm{ACF} =
\sum_{k=1}^K w_k R[k]
\end{equation}
Here, we simply adopt $K=4$ and uniform weights. 
The similarity $\mathrm{sim}(u_t, u_{t-k})$ is computed using two approaches: cosine similarity based on sentence embeddings and sentence-level chrF~\citep{popovic-2015-chrf}. 
For the former, we employ the jina-embeddings-v3 model~\citep{sturua2024jinaembeddingsv3multilingualembeddingstask} for encoding. 
Our analysis is conducted on the English side\footnote{
Although the original languages of some datasets are not English,
both the chrF metric and the performance of the jina-embedding-v3 model exhibit obvious variation across different languages. Therefore, to maintain relative fairness, 
we conduct the evaluation on the common English side.
} of IWSLT2017, BWB, GuoFeng, 
and GlobVDoc, considering two types of units: sentences and FRC pairs.

\begin{figure}[t]
    \centering
    \includegraphics[width=1\columnwidth]{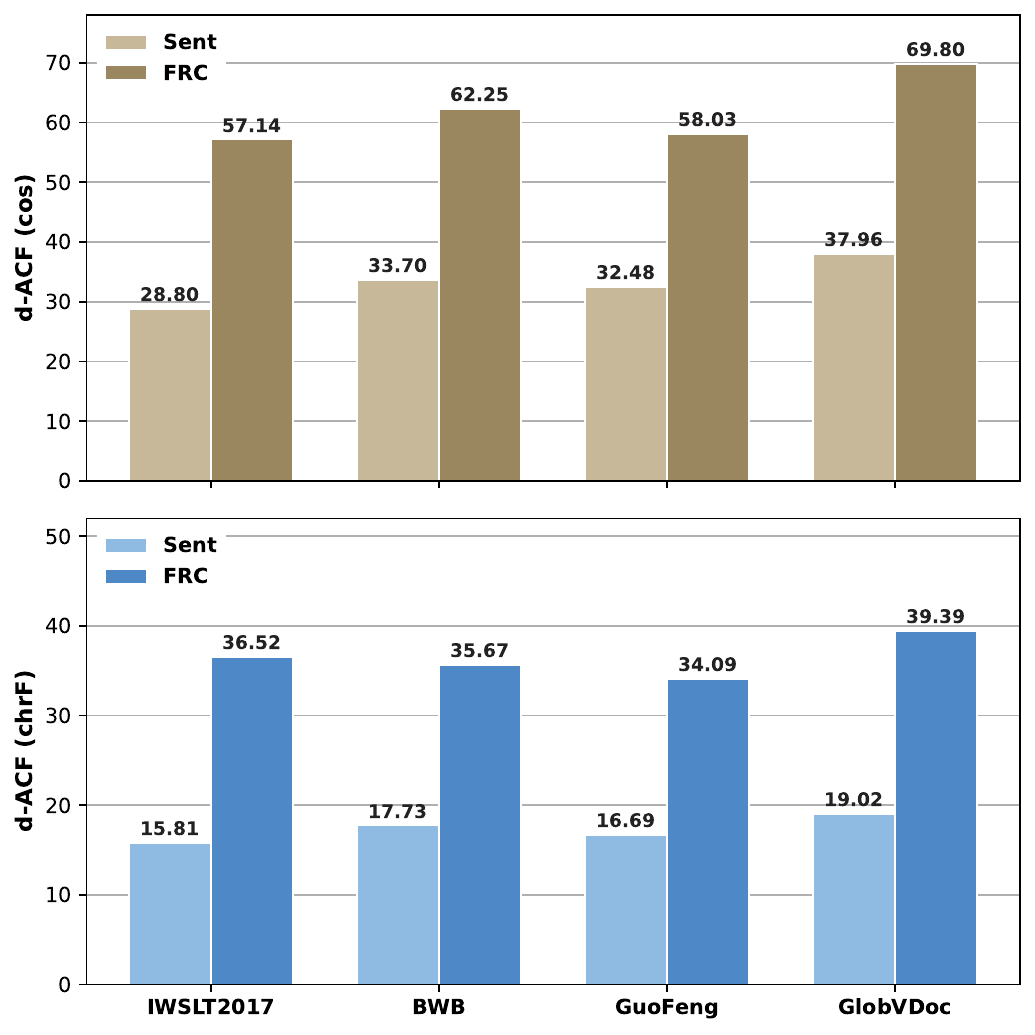}
    \caption{The results of d-ACF~($\times100$) on four datasets. }
    \label{fig:acf_results}
    \vspace{-1em}
\end{figure}

As shown in Figure~\ref{fig:acf_results},
on the cosine-based d-ACF metric, 
GlobVDoc outperforms all other datasets at both the sentence level and the chunk level, 
indicating stronger semantic autocorrelation and greater document-level topical coherence.
From a lexical perspective, 
the d-ACF value gap between GlobVDoc and the other datasets is smaller than that observed under the cosine-based setting, 
but GlobVDoc still maintains a slight advantage.

\begin{figure}[t]
    \centering
    \includegraphics[width=\columnwidth]{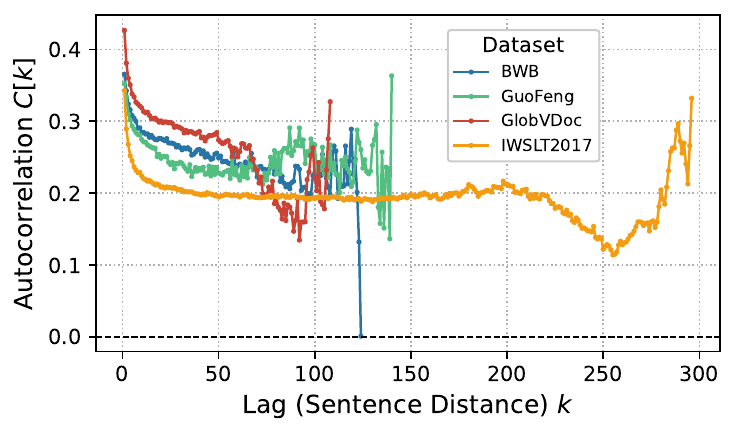}
    \caption{Autocorrelation distributions of all datasets across all lag values $k$.}
    \label{fig:acf_cos_all}
\end{figure}

Due to the above discussion, which considers only autocorrelation values at relatively close intra-document positions 
(with the lag $k$ limited to 4),
we further examine the overall document-level behavior by analyzing the autocorrelation distributions of all datasets across all lag values $k$.
As presented in Figure~\ref{fig:acf_cos_all}, these distributions are computed at the sentence level using cosine similarity.

We observe that, at the initial stage of the curves, 
all datasets exhibit a sharp decline followed by a more gradual decrease, 
indicating that strong semantic relatedness is primarily concentrated among sentences that are close to each other.
Furthermore, both GlobVDoc and IWSLT2017 show another significant drop followed by an increase toward the end of the curves,
resulting in an overall U-shaped pattern. 
In particular, for IWSLT2017, the notable rise after around lag $k = 290$ suggests that certain documents in the corpus may exhibit strong correspondences between their beginnings and endings.
In contrast, BWB and GuoFeng do not display a clear U-shaped trend in the later stages.
Instead, after the steady decline, their curves gradually increase. 
This indicates that although semantic correlation at medium distances continues to weaken, these datasets still demonstrate meaningful long-range coherence at larger sentence separations.

\section{Experimental Setups}
\label{sec:train_setup}

\subsection{Hyperparameters for FRC and Chunk Alignment}

For Fixed-Range Chunking, we set the maximum number of chunks allowed to fall outside the predefined range during the relaxed optimization stage to $K = 8$.
This setting is intended to improve the robustness of the algorithm. 
Under this configuration, no invalid samples were encountered during either training data construction or test-time chunk generation.

For the Dual-Boundary Matching-based Chunk Alignment algorithm, we set $\lambda = 0.5$ and $\sigma = 0.05$.
In addition, the number $k$ of target-unit candidates considered for each source boundary is set to 256.

\subsection{Training Data Filtering and Statistics}
\label{sec:data_stat}

Given that the DocBlocks dataset comprises high-quality data filtered via quality-aware methods~\citep{rei-etal-2023-scaling, ramirez-sanchez-etal-2020-bifixer}, 
we adopt a simple strategy to process chunk pairs:
\begin{enumerate}
\item Exclude chunk pairs where the source chunk length exceeds the upper threshold of the predefined range.
\item Determine the median length ratio for each language pair, 
and then filter the parallel chunk pairs by retaining only those within the interval $[\text{median}/1.5,\text{median}\times1.5]$.
\end{enumerate}
We utilize these outlier samples as breakpoints to further segment documents into subdocuments.
Table~\ref{tab:frc_outliers} presents the statistics for outlier samples, 
which can also characterize the data attrition rate of our proposed Dual-Boundary Matching-based Chunk Alignment Algorithm.

\begin{table*}[t]
    \centering
    \begin{adjustbox}{width=\textwidth}
    \begin{tabular}{lccccccccccc}
        \toprule
        \multirow{2}{*}{\textbf{\makecell{Lang \\ Pair}}} & \textbf{Median Ratio} & \textbf{Outlier Count} & \textbf{Outlier\%$\downarrow$} & \textbf{Outlier\%$\downarrow$} & \textbf{Delta$\downarrow$} & & \textbf{Median Ratio} & \textbf{Outlier Count} & \textbf{Outlier\%$\downarrow$} & \textbf{Outlier\%$\downarrow$} & \textbf{Delta$\downarrow$} \\
         & \textbf{[L $\to$ B]} & \textbf{[L $\to$ B]} & \textbf{(L)} & \textbf{(B)} & \textbf{[L $\to$ B]} & & \textbf{[Q $\to$ T]} & \textbf{[Q $\to$ T]} & \textbf{(Q)} & \textbf{(T)} & \textbf{[Q $\to$ T]}\\
        \midrule
        en$\leftrightarrow$de & 1.56 $\to$ 1.57 & 3,847 $\to$ 22,038 & 1.40\% & 8.02\%  & +6.62\% &|& 1.56 $\to$ 1.62 & 3,847 $\to$ 4,058 & 1.40\% & 1.35\% & -0.05\%\\
        en$\leftrightarrow$es & 1.41 $\to$ 1.41 & 2,007 $\to$ 5,923  & 0.71\% & 2.10\%  & +1.39\% &|& 1.41 $\to$ 1.50 & 2,007 $\to$ 2,217 & 0.71\% & 0.70\% & -0.01\%\\
        en$\leftrightarrow$fr & 1.50 $\to$ 1.50 & 3,061 $\to$ 9,609  & 1.14\% & 3.57\%  & +2.43\% &|& 1.50 $\to$ 1.56 & 3,061 $\to$ 3,342 & 1.14\% & 1.13\% & -0.01\%\\
        en$\leftrightarrow$it & 1.53 $\to$ 1.53 & 2,235 $\to$ 10,135 & 1.05\% & 4.75\%  & +3.70\% &|& 1.53 $\to$ 1.56 & 2,235 $\to$ 2,463 & 1.05\% & 1.07\% & +0.02\%\\
        \textbf{en$\leftrightarrow$ko} & 1.51 $\to$ 1.51 & 557 $\to$ 2,339    & 3.31\% & 13.91\% & +10.60\%&|& \textbf{1.51 $\to$ 2.51} & 557 $\to$ 1,087   & 3.31\% & 4.34\% & +1.03\%\\
        en$\leftrightarrow$nl & 1.61 $\to$ 1.61 & 5,587 $\to$ 25,912 & 2.58\% & 11.99\% & +9.41\% &|& 1.61 $\to$ 1.66 & 5,587 $\to$ 5,991 & 2.58\% & 2.54\% & -0.04\%\\
        en$\leftrightarrow$pt & 1.45 $\to$ 1.45 & 2,921 $\to$ 9,643  & 1.45\% & 4.79\%  & +3.34\% &|& 1.45 $\to$ 1.59 & 2,921 $\to$ 3,269 & 1.45\% & 1.45\% & 0.00\%\\
        en$\leftrightarrow$ru & 1.75 $\to$ 1.75 & 746 $\to$ 1,793    & 0.64\% & 1.55\%  & +0.91\% &|& 1.75 $\to$ 1.89 & 746 $\to$ 632     & 0.64\% & 0.48\% & -0.16\%\\
        \textbf{en$\leftrightarrow$zh} & 0.88 $\to$ 0.81 & \textbf{20,396 $\to$ 232,791} & \textbf{4.48\%} & \textbf{51.13\%} & \textbf{+46.65\%} &|& \textbf{0.88 $\to$ 1.34} & 20,396 $\to$ 40,353 & 4.48\% & 6.37\% & +1.89\%\\
        \bottomrule
    \end{tabular}
    \end{adjustbox}
    \caption{Outlier sample statistics across different similarity approaches and tokenizers.
    \textbf{L} denotes chunk alignment methods using the \textsc{Qwen2.5-7B-Instruct} tokenizer paired with either LaBSE embedding cosine similarity. And \textbf{B} denotes nllb-200-distilled-600M translation based sentence-BLEU as the similarity score,
    respectively. 
    \textbf{Q} and \textbf{T} represent the use of \textsc{Qwen2.5-7B-Instruct} and \textsc{Tower-7B-Instruct} tokenizers while maintaining LaBSE-based cosine similarity as the alignment criterion.
    ``en$\leftrightarrow$xx'' encompasses bidirectional data, 
     the length ratio is uniformly calculated for all pairs as the length of the non-English language relative to the English text.}
    \label{tab:frc_outliers}
\end{table*}

\begin{table*}[t]
    \centering
    \begin{adjustbox}{width=\textwidth}
    \begin{tabular}{lcccc}
        \toprule
        \multirow{2}{*}{\textbf{\makecell{Lang \\ Pair}}} & \textbf{Total Samples} & \textbf{Median Ratio} & \textbf{Outlier Count} & \textbf{Outlier\%$\downarrow$} \\
        & \multicolumn{4}{c}{\textbf{$[0,256]$ $\to$ $[256,512]$ $\to$ $[512,768]$}} \\
        \midrule
        en$\leftrightarrow$de & 894,029 $\to$ 301,176 $\to$ 179,509 & 1.62 $\to$ 1.62 $\to$ 1.62 & 32,799 $\to$ 4,058 $\to$ 2,124 & 3.67\% $\to$ 1.35\% $\to$ 1.18\%\\
        en$\leftrightarrow$es & 934,705 $\to$ 314,982 $\to$ 187,526 & 1.50 $\to$ 1.50 $\to$ 1.50 & 14,699 $\to$ 2,217 $\to$ 1,304 & 1.57\% $\to$ 0.70\% $\to$ 0.70\%\\
        en$\leftrightarrow$fr & 877,691$\to$ 295,207 $\to$ 176,007 & 1.56 $\to$ 1.56 $\to$ 1.56 & 25,941 $\to$ 3,342 $\to$ 1,859 & 2.96\% $\to$ 1.13\% $\to$ 1.06\%\\
        en$\leftrightarrow$it & 683,886 $\to$ 230,570 $\to$ 137,893 & 1.56 $\to$ 1.56 $\to$ 1.56 & 17,819 $\to$ 2,463 $\to$ 1,376 & 2.61\% $\to$ 1.07\% $\to$ 1.00\%\\
        en$\leftrightarrow$ko & 73,353$\to$ 25,073 $\to$ 15,034 & 2.51 $\to$ 2.51 $\to$ 2.51 & 9,061 $\to$ 1,087 $\to$ 466   & 12.4\% $\to$ 4.34\% $\to$ 3.10\%\\
        en$\leftrightarrow$nl & 700,761 $\to$ 235,593 $\to$ 140,879 & 1.66 $\to$ 1.66 $\to$ 1.67 & 45,464 $\to$ 5,991 $\to$ 3,226 & 6.49\% $\to$ 2.54\% $\to$ 2.29\%\\
        en$\leftrightarrow$pt & 669,903$\to$ 225,892 $\to$ 135,099 & 1.59 $\to$ 1.59 $\to$ 1.59 & 18,179 $\to$ 3,269 $\to$ 1,859 & 2.71\% $\to$ 1.45\% $\to$ 1.38\%\\
        en$\leftrightarrow$ru & 393,606 $\to$ 132,690 $\to$ 79,083 & 1.88 $\to$ 1.89 $\to$ 1.89 & 9,271 $\to$ 632 $\to$ 197     & 2.36\% $\to$ 0.48\% $\to$ 0.25\%\\
        en$\leftrightarrow$zh & 1,862,572$\to$ 633,248 $\to$ 381,331 & 1.35 $\to$ 1.34 $\to$ 1.33 & 289,068 $\to$ 40,353 $\to$ 13,312 & 15.51\% $\to$ 6.37\% $\to$ 3.49\%\\
        \bottomrule
    \end{tabular}
    \end{adjustbox}
    \caption{Sample statistics across different fixed ranges.
    Data samples were curated using LaBSE-based cosine similarity for chunk alignment and processed with the \textsc{Tower-7B-Instruct} tokenizer.
    The sequence $\ast\to\ast\to\ast$ denotes the data corresponding to fixed-range intervals of $[0,256]$, $[256,512]$, and $[512,768]$, respectively.}
    \label{tab:frc_out-_stat}
\end{table*}

In addition to the LaBSE-based cosine similarity utilized for chunk alignment in Section~\ref{sec:main_setup},
we introduce a comparative strategy using sentence-BLEU scores derived from the nllb-200-distilled-600M model\footnote{\url{https://huggingface.co/facebook/nllb-200-distilled-600M}}~\citep{nllbteam2022languageleftbehindscaling} translations. 
Both methods employ the \textsc{Qwen2.5-7B-Instruct} tokenizer for length measurement during the initial FRC phase.
Our results indicate that while both approaches yield nearly identical median ratios due to the shared tokenizer, 
the translation-based method leads to a substantial increase in the outlier rate,
particularly for zh.
This discrepancy likely stems from the high proportion of GuoFeng and BWB data within the DocBlocks dataset.
As these materials belong to the web novel domain, 
they pose relative translation challenges.

\begin{figure}[t]
  \centering
  \includegraphics[width=\linewidth]{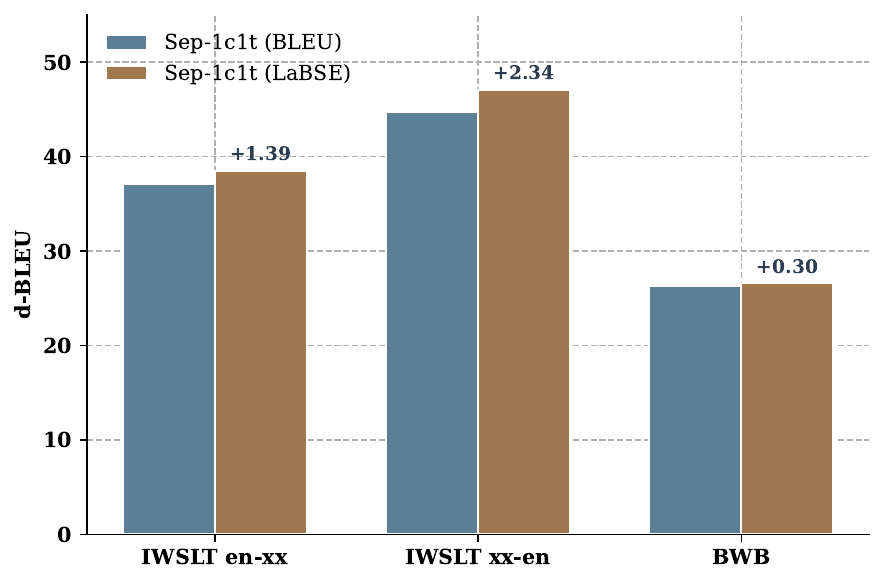}
  \caption{Comparison of \textsc{Qwen2.5-7B-Instruct} trained with \textsc{Sep}-1c1t on samples derived from chunk alignment using SentenceBLEU scores v.s. LaBSE-based cosine similarity.}
  \label{fig:bleu_labse}
  \vspace{-1em}
\end{figure}

To quantitatively assess the difference in alignment quality 
between chunk alignment based on SentenceBLEU scores 
and that based on LaBSE-based cosine similarity, 
we also trained \textsc{Qwen2.5-7B-Instruct} model
using the \textsc{Sep}-1c1t strategy on samples
obtained with the former method. 
A simple comparison of the results is shown in Figure~\ref{fig:bleu_labse}.
After data filtering, the LaBSE-based method, 
which retained a larger number of samples, 
unsurprisingly achieves consistently better performance on the test datasets than the SentenceBLEU-based approach. 
Notably, although the SentenceBLEU-based approach discarded a substantial number of outlier Chinese data points, 
its performance on the BWB dataset is not far behind that of the LaBSE-based approach. 
This suggests that the performance gains brought by large-scale Chinese training data exhibit a clear pattern of diminishing marginal returns.

Provided that LaBSE cosine similarity is used consistently, 
the sample attrition rates for FRC using either the \textsc{Qwen2.5-7B-Instruct} or \textsc{Tower-7B-Instruct} tokenizers do not differ significantly. 
However, inherent discrepancies between these tokenizers lead to substantial variations in cross-lingual length ratios, 
a phenomenon particularly prominent in ko and zh.

Subsequently, based on these subdocuments and their corresponding FRCs, 
we construct training samples tailored to the formatting requirements of the \textsc{d2dFT}, 
\textsc{Stair}, \textsc{Sep}, and \textsc{Uni} models. 
Table~\ref{tab:num_sample} reports the number of final training samples under the fixed range of $[256,512]$, 
while Table~\ref{tab:frc_out-_stat} presents the statistics for the three fixed-range datasets.
In addition,
Table~\ref{tab:frc_out-_stat} reports only the number of samples for en$\leftrightarrow$xx language pairs.
However, there are also a very small number of samples corresponding to language pairs that do not include English. 
For example, under the fixed-range setting of $[256,512]$, there are 63,403 such samples. 
These samples were also included in the final training dataset.

\begin{figure*}[htbp]
    \centering
    \includegraphics[width=1.0\textwidth]{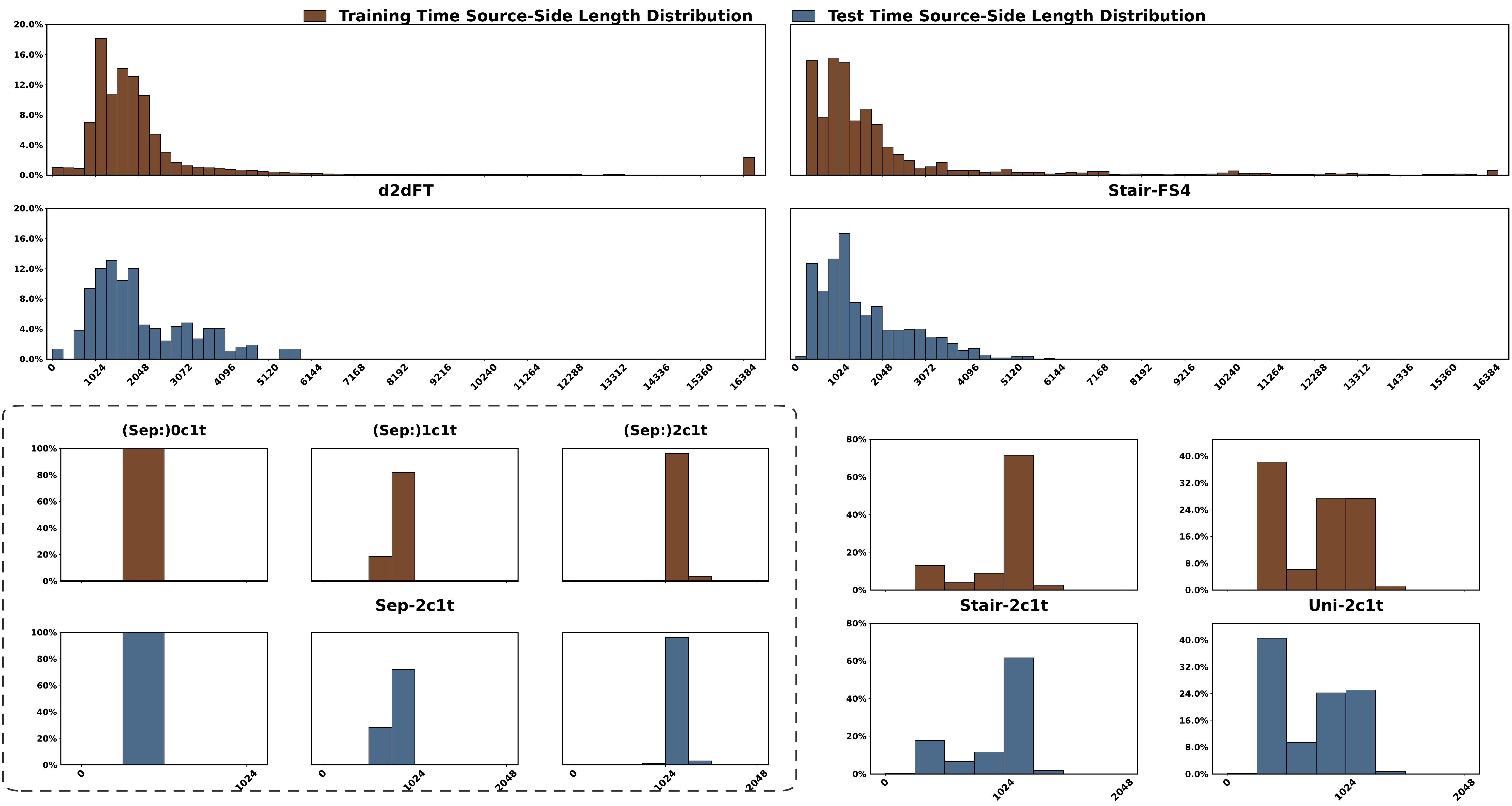}
    \caption{Training- and test-time (source-side) input length distributions. 
    The x-axis denotes input length, 
    and the y-axis denotes the proportion of documents or chunks in each length bin. 
    Lengths are measured using the \textsc{Tower}'s tokenizer. 
    Training statistics are computed on DocBlocks, 
    and test statistics are computed on IWSLT2017 en-xx.}
    \label{fig:length_distribution}
    \vspace{-1em}
\end{figure*}

To further illustrate the effect of FRC on length control, 
Figure~\ref{fig:length_distribution} visualizes the training- and test-time source-side input length distributions of different training formats under the \textsc{Tower} backbone
with the fixed range $[256,512]$. 
Overall, \textsc{d2dFT} and \textsc{Stair-FS4} exhibit broader or less-aligned train-test length distributions, 
whereas \textsc{Sep}-2c1t, \textsc{Stair}-2c1t, and \textsc{Uni}-2c1t, 
show more compact and better-aligned distributions.
Notably, the three sub-models of \textsc{Sep}-2c1t exhibit particularly sharp and narrowly bounded length distributions.

\begin{table}[h]
\centering
\begin{tabular}{lrr} 
\toprule
\multirow{2}{*}{\textbf{Model}} & \multicolumn{2}{c}{\textbf{Tokenizer}} \\ 
\cmidrule(lr){2-3}
 & \textbf{\textsc{Qwen2.5}} & \textbf{\textsc{Tower}} \\ 
\midrule
\textsc{d2dFT}       & 241,828   & 241,828 \\
\textsc{Stair}-FS4   & 953,631   & 1,016,814 \\
\textsc{Stair}-1c1t  & 2,004,084 & 2,331,028 \\
\textsc{Stair}-2c1t  & 2,004,084 & 2,331,028 \\
(\textsc{Sep}:) 0c1t    & 2,004,084 & 2,331,028 \\
(\textsc{Sep}:) 1c1t  & 1,743,169 & 2,056,597 \\
(\textsc{Sep}:) 2c1t  & 1,494,053 & 1,797,915 \\
\textsc{Uni}-2c1t  & --        & 6,185,540 \\
\bottomrule
\end{tabular}
\caption{The number of training data utilized for each model configuration.}
\label{tab:num_sample}
\vspace{-1em}
\end{table}

\subsection{Training and Decoding Setups}
We employ supervised fine-tuning~(SFT) to train our models,
where the loss is computed only on the target-side tokens.
Given a document pair $d = (d^s,d^t) \in \mathcal{D}$ after FRC-based chunk alignment,
we denote the aligned chunk pairs as
$\{(c_{d,i}^s, c_{d,i}^t)\}_{i=1}^{|d|}$.
$C_{d,i}^{*}$ is the set of contextual
source chunks used for predicting the $i$-th target chunk.

\paragraph{\textsc{d2dFT}}
\begin{equation}
\mathcal{L}_{\text{\textsc{d2dFT}}}
=
- \sum_{d \in \mathcal{D}}
\log p_\theta
\big(
d^t \mid d^s
\big)
\end{equation}
\begin{equation}
\theta_{\text{\textsc{d2dFT}}}
=
\arg\min_{\theta}
\mathcal{L}_{\text{\textsc{d2dFT}}}(\theta)
\end{equation}

\paragraph{\textsc{Stair}-2c1t}
\begin{equation}
C_{d,i}^{\text{\textsc{Stair}-2c1t}} =
\begin{cases}
\varnothing, & i = 1, \\
\{c^s_{d,1}\}, & i = 2, \\
\{c^s_{d,i-2}, c^s_{d,i-1}\}, & i \ge 3 .
\end{cases}
\end{equation}

{\small
\begin{equation}
\mathcal{L}_{\text{\textsc{Stair}-2c1t}}
=
- \sum_{d \in \mathcal{D}}
\sum_{i=1}^{|d|}
\log p_\theta
\big(
c^t_{d,i} \mid c^s_{d,i}, C_{d,i}^{\text{\textsc{Stair}-2c1t}}
\big)
\end{equation}
}

\begin{equation}
\theta_{\text{\textsc{Stair}-2c1t}}
=
\arg\min_{\theta}
\mathcal{L}_{\text{\textsc{Stair}-2c1t}}(\theta)
\end{equation}

\paragraph{\textsc{Stair}-FS4}
We merge the aligned chunk pairs into four consecutive chunks within one document,
defined as $\{(z^s_{d,i}, z^t_{d,i})\}_{i=1}^{4}$.
\begin{equation}
C_{d,i}^{\text{\textsc{Stair}-FS4}} =
\begin{cases}
\varnothing, & i = 1, \\
\{z^s_{d,1}\}, & i = 2, \\
\{z^s_{d,1}, z^s_{d,2}\}, & i = 3 \\
\{z^s_{d,1}, z^s_{d,2}, z^s_{d,3}\}, & i = 4.
\end{cases}
\end{equation}

{\small
\begin{equation}
\mathcal{L}_{\text{\textsc{Stair}-FS4}}
=
- \sum_{d \in \mathcal{D}}
\sum_{i=1}^{4}
\log p_\theta
\big(
z^t_{d,i} \mid z^s_{d,i}, C_{d,i}^{\text{\textsc{Stair}-FS4}}
\big)
\end{equation}
}

\begin{equation}
\theta_{\text{\textsc{Stair}-FS4}}
=
\arg\min_{\theta}
\mathcal{L}_{\text{\textsc{Stair}-FS4}}(\theta)
\end{equation}

\paragraph{\textsc{Sep / Uni}-2c1t}
\begin{equation}
C_{d,i}^{(k)} =
\begin{cases}
\varnothing, & k = \text{0c1t}, \\
\{c^s_{d,i-1}\}, & k = \text{1c1t}, \\
\{c^s_{d,i-2}, c^s_{d,i-1}\}, & k = \text{2c1t} .
\end{cases}
\end{equation}

\begin{equation}
\mathcal{L}^{(k)}
=
- \sum_{d \in \mathcal{D}}
\sum_{i=1}^{|d|}
\log p_\theta
\big(
c^t_{d,i} \mid c^s_{d,i}, C_{d,i}^{(k)}
\big)
\end{equation}

{\small
\begin{equation}
\theta^{(k)}_{\text{\textsc{Sep}-2c1t}}
=
\arg\min_{\theta}
\mathcal{L}^{(k)}(\theta),
k \in \{\text{0c1t}, \text{1c1t}, \text{2c1t}\}
\end{equation}
}

\begin{equation}
\theta_{\text{\textsc{Uni}-2c1t}}
=
\arg\min_{\theta}
\sum_k
\mathcal{L}^{(k)}(\theta)
\end{equation}

The training and inference prompts are presented in the Appendix~\ref{sec:prompts}.

The training hyperparameters almost follow the configurations established by~\citet{ramos2025multilingual}.
However, we adjust the batch size to 16 for \textit{Doc2Doc} fine-tuning 
while setting it to 64 for all other models. 
All training processes were conducted using 8 NVIDIA H100 GPUs,
optimized by AdamW~\citep{loshchilov2018decoupled}.
Table~\ref{tab:detail_hyper} contains the detailed hyperparameter configuration for the training step.

\begin{table}[h]
\centering
\begin{adjustbox}{width=\columnwidth}
\begin{tabular}{ll}
\hline
Batch size                 & 16~(\textsc{d2dFT}) / 64~(others) \\
Number of Epochs           & 2 \\
Learning rate              & $7 \times 10^{-6}$ \\
LR Scheduler               & cosine \\
Warmup Steps               & 125 \\
Weight Decay               & 0.01 \\
Optimizer                  & AdamW\\
Adam $\beta_1$             & 0.9 \\
Adam $\beta_2$             & 0.999 \\
Adam $\epsilon$            & $1 \times 10^{-8}$ \\
Maximum Sequence Length    & 32,768 \\
\hline
\end{tabular}
\end{adjustbox}
\caption{Hyperparameter configuration to fine-tune models.}
\label{tab:detail_hyper}
\vspace{-1em}
\end{table}

Both inference and evaluation were conducted on a single H100 GPU. For the decoding process, 
we utilized vLLM~\citep{kwon2023efficient} and employed a greedy decoding strategy.

During inference, d2d refers to direct document-to-document translation. 
The $*$c1t models utilize a staged decoding approach: 
0c1t translates all fixed-range chunks without context; 
1c1t initiates the first chunk of each document in 0c1t mode,
while subsequent chunks follow the 1c1t format; 
2c1t decoding begins with 0c1t for the first chunk and 1c1t for the second,
with all remaining chunks processed using 2c1t context. 
FS4 follows a similar staged procedure. 
All inference prompts are kept consistent with the training prompts shown in Appendix~\ref{sec:prompts}.

\begin{figure*}[htbp]
    \centering
    \includegraphics[width=0.9\textwidth]{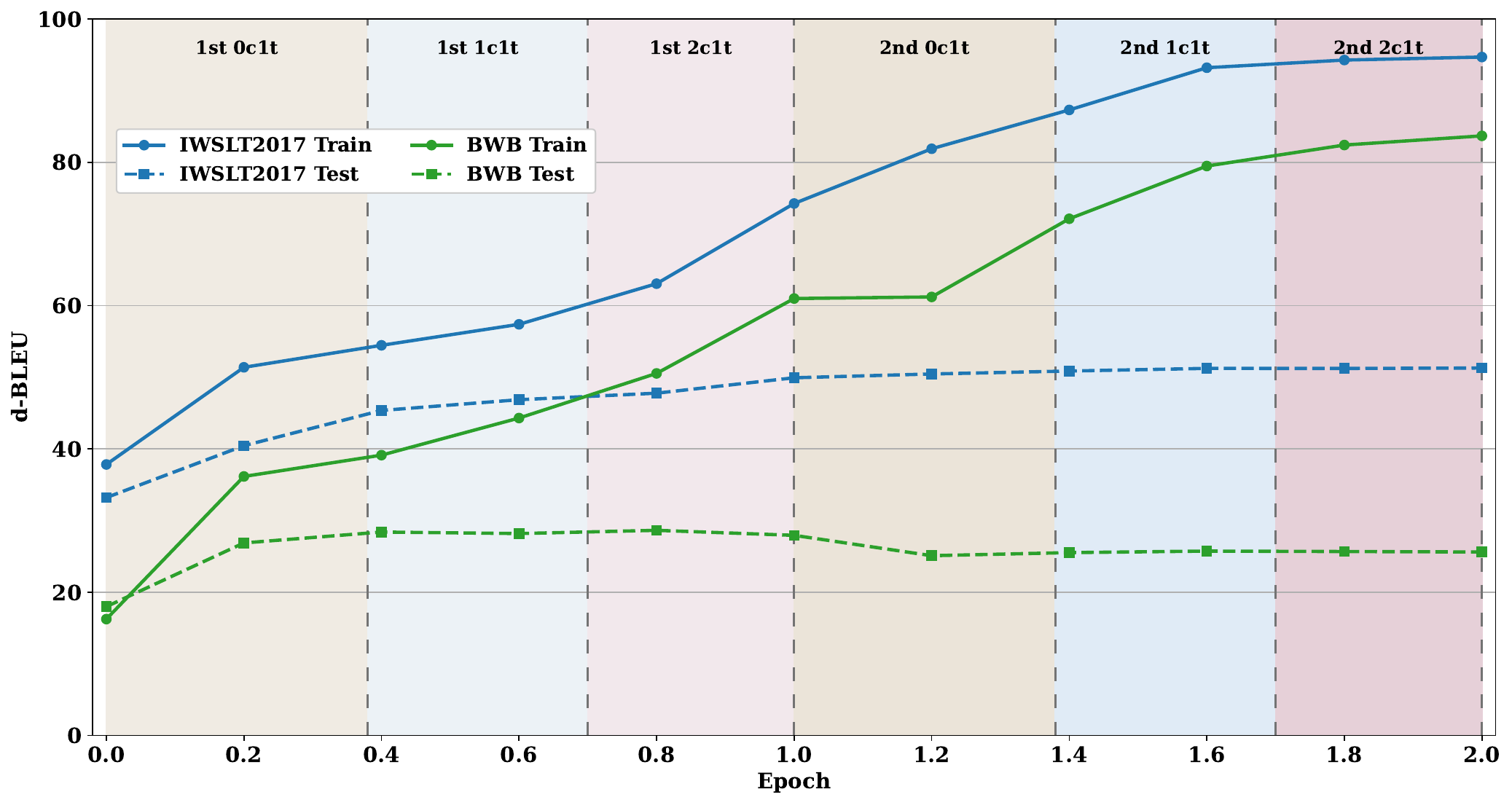}
    \caption{Training Process of \textsc{Tower}-based \textsc{Uni}-2c1t model.}
    \label{fig:uni-2c1t_training}
\end{figure*}

\section{Analysis of Joint Training}
\label{sec:uni_training}

\subsection{Training Process of \textsc{Uni}-2c1t}

Given that the training data for \textsc{Uni}-2c1t 
reaches 6M samples~(see Table~\ref{tab:num_sample})
and that the model repeatedly encounters each source chunk three times through the 0c1t, 1c1t, and 2c1t formats, 
we sought to investigate the potential risk of overfitting. 
We monitored the performance of the \textsc{Uni}-2c1t model
on both the test set and a subset of the training data throughout the training process.
The training subset was constructed via random sampling,
consisting of 12 document pairs selected from each of the IWSLT2017 and BWB datasets.

The results are illustrated in Figure~\ref{fig:uni-2c1t_training}.
Since the IWSLT2017 training data is essentially bi-directional~(xx-xx),
we report the averaged performance across en-xx and xx-en directions on the test set.
During the training process,
model checkpoints were saved every 0.2 epochs. 
Since the training data is in a fixed sequence of
0c1t, 1c1t, and 2c1t formats within each epoch,
we have explicitly marked the specific time points
at which the learning phase for each data format concludes.

Regarding the training data,
both the IWSLT2017 and BWB subsets exhibit a natural and consistent improvement 
in fitting as training progresses. 
For the test data, 
the performance on IWSLT2017 demonstrates a gradual increase before eventually plateauing, 
with no discernible signs of overfitting. 
However, a distinct trend emerges for the BWB data during the transition from epoch 1.0 to 1.2: 
we observe a marginal gain on the training set accompanied by a performance dip on the test set.
This suggests that the transition in data format,
reverting from 2c1t back to 0c1t at the start of a new epoch,
may introduce temporary optimization confusion for the BWB domain. 
In the subsequent phased learning stages, 
while the model continues to fit the training data rapidly,
its generalization performance on the test data eventually stabilizes.
However, given the model’s exceptionally high degree of fitting to the training data,
its generalization performance may already have been limited.

\subsection{Analysis of Context Sensitivity}

To better understand the behavior of \textsc{Uni}-2c1t under joint training, 
we conduct additional analyses to investigate whether the model effectively utilizes contextual information 
or instead primarily relies on the source chunk itself.

\paragraph{Context replacement: Test.}
We randomly replace the context associated with each source chunk in the in-distribution test set while keeping the source chunk and target chunk unchanged. 
To maximize the perturbation,
the replacement context is sampled from a different document written in a different language.
During evaluation, we remove the first chunk of each document since it has no preceding context. 
We denote the original context as Orig-C and the replaced context as RDM-C. 
All results are averaged over five independent runs.

\begin{table*}[t]
\centering
\small
\begin{tabular}{llcc|cc|cc}
\toprule
\multirow{2}{*}{Model} & \multirow{2}{*}{Decoding}
& \multicolumn{2}{c|}{IWSLT en-xx}
& \multicolumn{2}{c|}{IWSLT xx-en}
& \multicolumn{2}{c}{BWB xx-en} \\
\cmidrule(lr){3-4}
\cmidrule(lr){5-6}
\cmidrule(l){7-8}
& & Orig-C & RDM-C & Orig-C & RDM-C & Orig-C & RDM-C \\
\midrule

\multirow{2}{*}{\textsc{Sep}-2c1t}
& 1c1t
& 39.26 & 37.86 (-1.40)
& 67.04 & 59.85 (-7.19)
& 26.55 & 23.37 (-3.18) \\

& 2c1t
& 38.40 & 37.02 (-1.38)
& 67.26 & 57.70 (-9.56)
& 26.15 & 21.96 (-4.19) \\

\midrule

\multirow{2}{*}{\textsc{Stair}-2c1t}
& 1c1t
& 39.12 & 37.43 (-1.69)
& 65.79 & 57.72 (-8.07)
& 27.18 & 23.12 (-4.06) \\

& 2c1t
& 39.16 & 37.60 (-1.56)
& 66.97 & 56.33 (-10.64)
& 27.18 & 22.16 (-5.02) \\

\midrule

\multirow{2}{*}{\textsc{Uni}-2c1t}
& 1c1t
& 36.36 & 36.13 (\textbf{-0.23})
& 66.28 & 64.25 (\textbf{-2.03})
& 25.22 & 24.72 (\textbf{-0.50}) \\

& 2c1t
& 36.55 & 35.84 (\textbf{-0.71})
& 66.43 & 63.66 (\textbf{-2.77})
& 25.28 & 24.28 (\textbf{-1.00}) \\

\bottomrule
\end{tabular}
\caption{d-BLEU scores under original context (Orig-C) and randomly replaced context (RDM-C). Numbers in parentheses denote the absolute performance drop after context replacement. Results are averaged over five runs.}
\label{tab:context_replace}
\end{table*}

\begin{table*}[t]
\centering
\small
\begin{tabular}{lccc|cc|cc|cc}
\toprule

\multirow{2}{*}{Model}
& \multicolumn{3}{c|}{Original}
& \multicolumn{2}{c|}{RDM1}
& \multicolumn{2}{c|}{RDM2}
& \multicolumn{2}{c}{RDM3} \\

\cmidrule(lr){2-4}
\cmidrule(lr){5-6}
\cmidrule(lr){7-8}
\cmidrule(l){9-10}

& 0c1t & 1c1t & 2c1t
& 1c1t & 2c1t
& 1c1t & 2c1t
& 1c1t & 2c1t \\

\midrule

\multirow{2}{*}{\textsc{Stair}-2c1t}

& 49.29 & 52.36 & 56.99
& 47.95 & 46.63
& 47.08 & 45.29
& 46.52 & 44.56 \\

& -- & -- & --
& -8.42\% & -18.18\%
& -10.08\% & -20.53\%
& -11.15\% & -21.81\% \\

\midrule

\multirow{2}{*}{\textsc{Uni}-2c1t}

& 97.35 & 97.50 & 97.35
& 95.40 & 95.27
& 96.62 & 95.78
& 96.14 & 95.22 \\

& -- & -- & --
& \textbf{-2.15\%} & \textbf{-2.14\%}
& \textbf{-0.90\%} & \textbf{-1.61\%}
& \textbf{-1.39\%} & \textbf{-2.19\%} \\

\bottomrule
\end{tabular}
\caption{d-BLEU on sampled training instances under different decoding formats and after random context replacement. Percentages denote the relative performance drop compared with the corresponding original score.}
\label{tab:training_context}
\end{table*}

As shown in Table~\ref{tab:context_replace}, 
replacing the context causes substantial performance degradation for both \textsc{Sep}-2c1t and \textsc{Stair}-2c1t across all test sets. 
In contrast, \textsc{Uni}-2c1t exhibits only marginal performance drops under the same perturbation. 
This observation suggests that \textsc{Uni}-2c1t is considerably less sensitive to contextual information, 
even when the original context is replaced with unrelated content.

\paragraph{Context replacement: Train.}
To further investigate this phenomenon, 
we sample 320 source chunks from the training set. 
These chunks appear only in the 2c1t format for \textsc{Stair}-2c1t,
while they appear in all three formats (0c1t, 1c1t, and 2c1t) for \textsc{Uni}-2c1t.
We first evaluate how well each model fits these training instances under different decoding formats and then perform the same context replacement experiment.

Table~\ref{tab:training_context} reveals two notable observations. 
First, \textsc{Stair}-2c1t achieves its highest fitting accuracy under the 2c1t decoding format, 
whereas \textsc{Uni}-2c1t attains nearly perfect d-BLEU scores across all three decoding formats, 
indicating that it fits the training data well regardless of the decoding format.
Second, replacing the context results in approximately a 20\% performance drop for \textsc{Stair}-2c1t,
while \textsc{Uni}-2c1t shows only negligible degradation. 
These results suggest that,
although \textsc{Uni}-2c1t is trained with contextual information, 
the learned representations become largely insensitive to context.
Instead, the model appears to rely primarily on the source chunk itself when generating translations.
This behavior may partially explain why jointly training all three data formats does not perform as well as other models.

\subsection{Training Cost}
Regarding computational efficiency,
the \textsc{(Sep:)} 0c1t, 1c1t, and 2c1t sub-models required 39h 48m, 37h 21m, and 36h 42m respectively on 8$\times$ H100 GPUs, 
totaling 113h 51m. 
In contrast, the \textsc{Uni}-2c1t model required 119h 35m under the same hardware configuration. 
Given its shorter total training time and the capacity for simultaneous parallel training of its sub-models, 
the \textsc{Sep} approach can offer a higher degree of training efficiency.

\section{Source-Only v.s. Source-MT Context}
\label{sec:con_src_vs_mt}
\subsection{Translation Quality–Efficiency Trade-off}
We employ the same FRC dataset that was constructed previously. 
However, we introduce reference translations into the context parts of the 1c1t and 2c1t data, 
naming these augmented formats as $\text{1c}^{+}\text{1t}$ and $\text{2c}^{+}\text{1t}$. 
Notably, due to cross-lingual differences in lexical representation, 
appending target-language text to the context precludes a unified input-side length distribution. 
As a result, the length distribution transitions to a language-wise paradigm.

\begin{figure*}[htbp]
    \centering
    \begin{subfigure}[b]{0.48\textwidth}
        \centering
        \includegraphics[width=\linewidth]{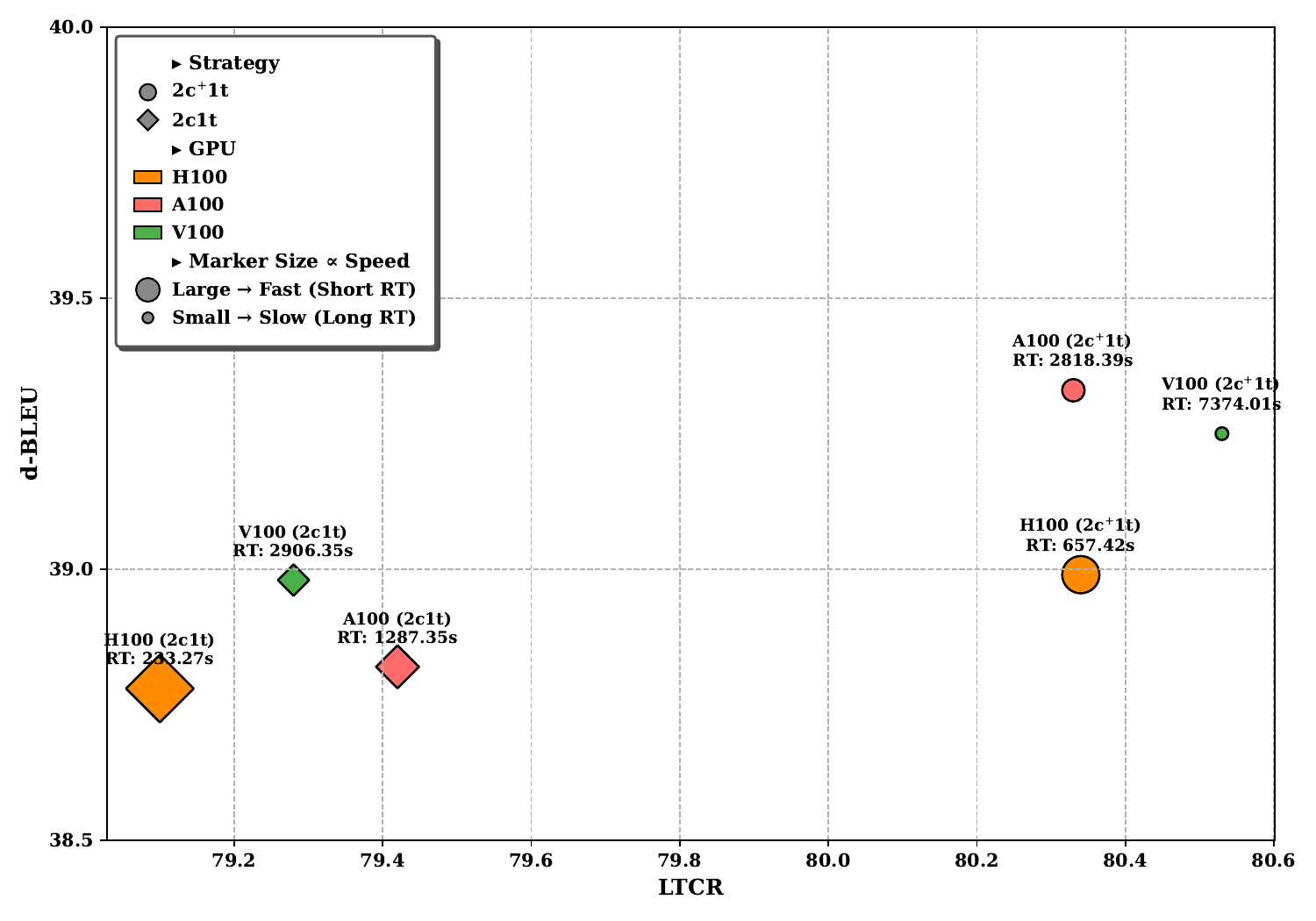}
        \caption{Results on IWSLT2017 en-xx.}
        \label{fig:iwslt_en-xx}
    \end{subfigure}
    \hfill
    \begin{subfigure}[b]{0.48\textwidth}
        \centering
        \includegraphics[width=\linewidth]{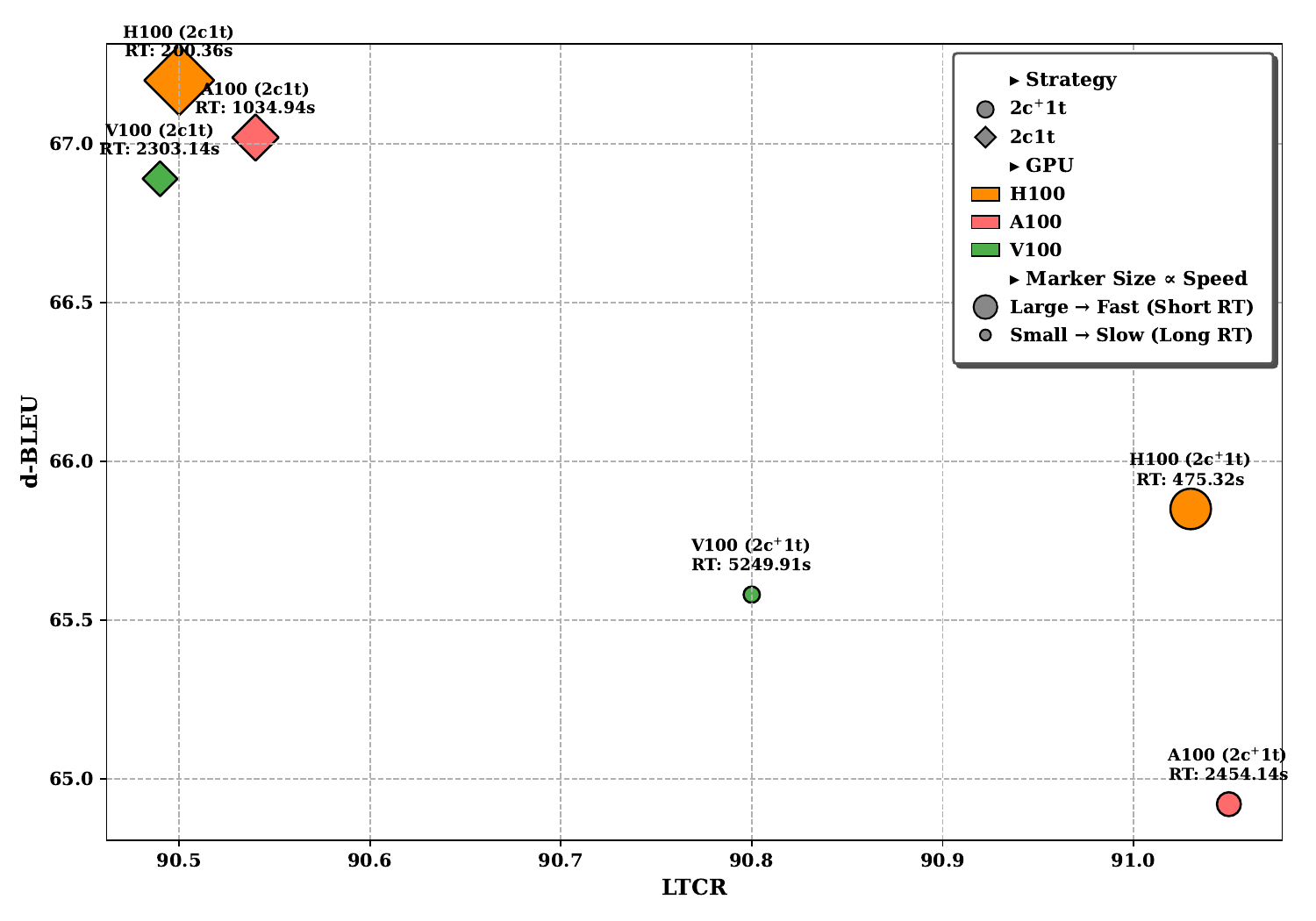}
        \caption{Results on IWSLT2017 xx-en.}
        \label{fig:iwslt_xx-en}
    \end{subfigure}
    \caption{Results of Source-Only Context v.s. Source-MT Context on the IWSLT2017 dataset.
    ``RT'' denotes runtime.
    The marker size is directly proportional to the decoding speed and inversely proportional to the runtime.}
    \label{fig:iwslt_both}
\end{figure*}

The results are presented in Figure~\ref{fig:iwslt_both}.
Utilizing the \textsc{Tower}-based model,
we evaluated the two strategies in terms of accuracy, term consistency,
and efficiency across various hardware configurations on the IWSLT2017 dataset.
However, it is worth noting that,
although the hyperparameter configurations remained strictly consistent for the comparative experiments conducted across different GPUs, 
we cannot guarantee completely identical execution environments. 
Specifically, the H100 trials were deployed on an AArch64 architecture, 
whereas the A100 and V100 experiments were executed on an x86-64 architecture.
We utilize the total runtime~(RT) 
required to execute the complete decoding script 
as the primary metric for evaluating decoding efficiency.

\begin{table*}[t]
\centering
\small
\begin{adjustbox}{width=\textwidth}
\setlength{\tabcolsep}{3pt}
\begin{tabular}{llcccccccccccccccccc}
\toprule
\multirow{3}{*}{Dataset} & \multirow{3}{*}{Lang}
& \multicolumn{6}{c}{d-BLEU $\uparrow$}
& \multicolumn{6}{c}{LTCR $\uparrow$}
& \multicolumn{6}{c}{NRR (\%) $\downarrow$} \\
\cmidrule(lr){3-8} \cmidrule(lr){9-14} \cmidrule(lr){15-20}
& &
\multicolumn{3}{c}{1c1t}
& \multicolumn{3}{c}{2c1t}
& \multicolumn{3}{c}{1c1t}
& \multicolumn{3}{c}{2c1t}
& \multicolumn{3}{c}{1c1t}
& \multicolumn{3}{c}{2c1t} \\
\cmidrule(lr){3-5} \cmidrule(lr){6-8}
\cmidrule(lr){9-11} \cmidrule(lr){12-14}
\cmidrule(lr){15-17} \cmidrule(lr){18-20}
& &
Base & $\text{c}^{+}$ & $\Delta_{1}$
& Base & $\text{c}^{+}$ & $\Delta_{2}$
& Base & $\text{c}^{+}$ & $\Delta_{1}$
& Base & $\text{c}^{+}$ & $\Delta_{2}$
& Base & $\text{c}^{+}$ & $\Delta_{1}$
& Base & $\text{c}^{+}$ & $\Delta_{2}$ \\
\midrule
\multirow{2}{*}{IWSLT2017} & en-xx
& \textbf{39.51} & 39.35 & $-0.16$
& 38.78 & \textbf{38.99} & $+0.21$
& 79.12 & \textbf{80.06} & $+0.94$
& 79.10 & \textbf{80.34} & $+1.24$
& \textbf{1.34} & 1.87 & $+0.53$
& 1.60 & 1.60 & $0.00$ \\

& xx-en
& \textbf{66.84} & 65.88 & $-0.96$
& \textbf{67.20} & 65.85 & $-1.35$
& 90.59 & \textbf{90.95} & $+0.36$
& 90.52 & \textbf{91.03} & $+0.51$
& \textbf{0.80} & 1.87 & $+1.07$
& \textbf{0.53} & 1.07 & $+0.54$ \\

BWB & xx-en
& \textbf{27.16} & 26.91 & $-0.25$
& \textbf{26.82} & 26.26 & $-0.56$
& 79.81 & \textbf{83.73} & $+3.92$
& 79.79 & \textbf{83.08} & $+3.29$
& 0.00 & 0.00 & $0.00$
& 0.00 & 0.00 & $0.00$ \\

GlobVDoc & en-xx
& 41.87 & \textbf{42.24} & $+0.37$
& \textbf{41.98} & 41.54 & $-0.44$
& 84.52 & \textbf{85.70} & $+1.18$
& 84.60 & \textbf{85.59} & $+0.99$
& 0.00 & 0.00 & $0.00$
& \textbf{0.00} & 1.11 & $+1.11$ \\

\bottomrule
\end{tabular}
\end{adjustbox}
\caption{
Comparison among 1c1t, 1$\text{c}^{+}$1t, 2c1t, and 2$\text{c}^{+}$1t across datasets.
``Base'' and ``$\text{c}^+$'' represent 1c1t and 1$\text{c}^{+}$1t,
respectively.
$\Delta_{1}$ denotes 1$\text{c}^{+}$1t $-$ 1c1t, and $\Delta_{2}$ denotes 2$\text{c}^{+}$1t $-$ 2c1t.
The better result for each metric on each dataset in the comparison is highlighted in bold.
}
\label{tab:compare_cplus_with_delta}
\vspace{-1em}
\end{table*}

As a consistent trend across all three hardware environments, 
under the 2$\text{c}^{(+)}$ configuration, 
incorporating previous translations naturally yields superior terminological consistency~\citep{choudhary-etal-2025-exploring}, 
alongside marginally higher d-BLEU scores in the IWSLT2017 en-xx direction
compared to the Source-Only method. 
Conversely,
the Source-Only approach exhibits a highly significant edge in decoding efficiency, 
paired with a distinct accuracy advantage in the xx-en direction.
Specifically, the decoding latency under the Source-MT context is, 
on average, approximately 2.5$×$ that of the Source-Only baseline. 
This efficiency gap is readily apparent even on high-speed GPUs such as the H100, 
and becomes especially pronounced on comparatively slower hardware like the V100. 
This is because leveraging previous translations as context inherently requires strictly sequential decoding,
in contrast to the Source-Only approach, 
which permits all chunks to be preprocessed and decoded in parallel.
Given that document-level machine translation is fundamentally a long-sequence generation task, 
decoding efficiency remains an indispensable consideration for practical, real-world deployment.

To further evaluate the translation quality of both approaches,
we conducted comprehensive experiments across all datasets using a single H100 GPU for inference.
The results are summarized in Table~\ref{tab:compare_cplus_with_delta}.
In terms of LTCR metric, the 1$\text{c}^{+}$1t configuration demonstrates comprehensive superiority. 
This advantage is particularly pronounced on the BWB dataset, 
where it approaches the peak performance of the \textsc{Stair}-FS4 method~(achieved under the Source-Only setting) 
and surpasses large-scale LLMs~(see Figure~\ref{fig:ltcr_results}).
Given the characteristics of the BWB dataset's web novel domain, 
domain-specific terminology~(e.g., character names) likely appears at a higher frequency than in other evaluated domains.
Consequently, incorporating the previous translation as context yields the most substantial improvement in terminological consistency. 
However, regarding translation accuracy measured by d-BLEU, 
the Source-Only approach maintains the advantage in the majority of cases. 
This performance is strongly negatively correlated with the NRR; a lower NRR indicates a more stable generation output.

\subsection{Exposure Bias: A Case Study}
Furthermore, for the $\text{c}^{+}$ strategy, 
conditioning on previous translations is vulnerable to exposure bias and error propagation~\citep{wu-etal-2018-beyond,arora-etal-2022-exposure}:
a suboptimal translation used as context can lead to cascading deterioration in the quality of subsequent chunk translations within the same document~\citep{mino-etal-2020-effective,zhang-etal-2021-multi}.
To illustrate this point, 
we provide a concise analytical overview alongside a qualitative case study.
We define ``broken chunks'' as outputs with log probabilities~(lp) below $−5$,
and ``repetition chunks'' that contain n-gram repetitions.
The distribution of these broken chunks across various languages under the 1$\text{c}^{+}$1t setting on the IWSLT2017 en-xx data is presented in Table~\ref{tab:broken_chunks}.
Notably, the number of repetition chunks exceeds that of broken chunks. 
This discrepancy arises because, 
in some instances,
the repetition initiates near the tail end of the chunk. 
Consequently, the overall log probabilities for the chunk may not drop below the $-5$ threshold~(typically hovering between $-3$ and $-5$).
Importantly, these specific chunks frequently act as the initial trigger for cascading deterioration,
as illustrated in Figure~\ref{fig:case_study}.

\begin{figure*}[t] 
\begin{tcolorbox}[
    title={Error Propagation and Recovery Across Chunks},
    colback=white,
    colframe=gray!70,
    colbacktitle=gray!70,
    coltitle=white,
    fonttitle=\bfseries\large,
    arc=4pt,
    boxrule=0.8pt,
    left=3mm, right=3mm, top=2mm, bottom=2mm
]

\newcolumntype{K}{>{\raggedright\arraybackslash}p{1.4cm}}
\newcolumntype{P}{>{\raggedright\arraybackslash}p{2.6cm}}
\newcolumntype{S}{>{\raggedright\arraybackslash}p{2.8cm}}

\textbf{\faLanguage \ En $\rightarrow$ Ko ($n=6$ chunks)} \hfill \textit{Status: Partial Recovery}
\vspace{1mm}

{\small
\begin{tabularx}{\linewidth}{P P S X}
    \hline
    \textbf{Chunk No.} & \textbf{Log-Prob} & \textbf{State} & \textbf{Description} \\
    \hline
    0, 1 & $\approx -0.71$ & \textcolor{statusNormal}{\faCheckCircle \ Normal} & Standard Korean output. \\
    2 & $-3.43$ & \textcolor{statusTrigger}{\faExclamationTriangle \ Trigger} & \textbf{[REP]} Initial repetition trigger. \\
    3 & $-368.97$ & \textcolor{statusBroken}{\faTimesCircle \ Broken} & Severely collapsed: ``\ko{의 의 의 의 \dots}'' \\
    4, 5 & $-1.08 \pm 0.3$ & \textcolor{statusRecover}{\faUndo \ Recovered} & Returns to normal flow. \\
    \hline
\end{tabularx}
}

\vspace{5mm} 

\textbf{\faLanguage \ En $\rightarrow$ Zh ($n=11$ chunks)} \hfill \textit{Status: Failed to Recover}
\vspace{1mm}

{\small
\begin{tabularx}{\linewidth}{P P S X}
    \hline
    \textbf{Chunk No.} & \textbf{Log-Prob} & \textbf{State} & \textbf{Description} \\
    \hline
    0, 1 & $\approx -1.00$ & \textcolor{statusNormal}{\faCheckCircle \ Normal} & Standard Chinese output. \\
    2 & $-6.63$ & \textcolor{statusTrigger}{\faExclamationTriangle \ Trigger} & \textbf{[REP]} Initial repetition trigger. \\
    3 & $-518.00$ & \textcolor{statusBroken}{\faTimesCircle \ Broken} & Deep failure: ``\zh{计 计 计 计 \dots}'' \\
    4 $\sim$ 10 & $-32 \sim -212$ & \textcolor{statusBroken}{\faBan \ Persistent} & Never recovers; remains in error state. \\
    \hline
\end{tabularx}
}

\vspace{5mm} 

\textbf{\faLanguage \ En $\rightarrow$ De ($n=7$ chunks)} \hfill \textit{Status: Full Recovery}
\vspace{1mm}

{\small
\begin{tabularx}{\linewidth}{P P S X}
    \hline
    \textbf{Chunk No.} & \textbf{Log-Prob} & \textbf{State} & \textbf{Description} \\
    \hline
    0 & $-1.48$ & \textcolor{statusTrigger}{\faExclamationTriangle \ Trigger} & \textbf{[REP]} Early trigger. \\
    1 & $-36.54$ & \textcolor{statusBroken}{\faTimesCircle \ Broken} & Broken; inherits errors from No.0 Chunk. \\
    2 $\sim$ 6 & $\approx -0.25$ & \textcolor{statusRecover}{\faCheckDouble \ Full Rec.} & Rapid and stable recovery. \\
    \hline
\end{tabularx}
}

\end{tcolorbox}
\caption{A Case Study: Potential Exposure issues in the Source-MT Context Setting.}
\label{fig:case_study}
\end{figure*}

We categorize the model's capacity to return to standard output following a generation collapse into three distinct groups:
Full Recovery indicates complete restoration after a brief degradation; 
Partial Recovery denotes a return to normal generation but with a sustained drop in output quality; 
and Failed to Recover signifies a continuous collapse until the end. 
In our analysis, we identified a total of 22 repetition chunks. 
Excluding those situated at the end of a document, 
19 chunks remained. 
Strikingly, only two of them~(the specific instances detailed in our case study) managed to recover when conditioned on a degraded translation context.
This yields an error propagation rate of 89\% (17/19), 
demonstrating that a mid-translation quality collapse is highly likely to trigger a subsequent cascading failure. 
In contrast to the independent decoding mechanism of the Source-Only context approach,
this susceptibility to error propagation represents a critical latent issue 
that must be addressed when utilizing previous target translations as context.

\begin{table}[t]
\centering
\begin{adjustbox}{width=\columnwidth}
\begin{tabular}{lrrrr}
\toprule
Lang & Chunks & Broken & Repetition & Mean Log-Prob \\
\midrule
ko & 519 & 8 & 13 & -3.496 \\
zh & 520 & 9 & 7 & -3.354 \\
de & 460 & 1 & 2 & -0.349 \\
fr & 521 & 0 & 0 & -0.234 \\
it & 92  & 0 & 0 & -0.238 \\
nl & 94  & 0 & 0 & -0.285 \\
\bottomrule
\end{tabular}
\end{adjustbox}
\caption{Distribution of broken and repetition chunks across languages on the IWSLT en-xx dataset~(\textsc{Tower}-based \textsc{Sep}-1$\text{c}^{+}$1t model.}
\label{tab:broken_chunks}
\vspace{-1em}
\end{table}

\section{Test-Time Chunking}
\label{sec:ttc}

\subsection{Length Analysis at Test Time}
\begin{figure}[t]
  \centering

  \begin{subfigure}{\linewidth}
    \centering
    \includegraphics[width=0.9\linewidth]{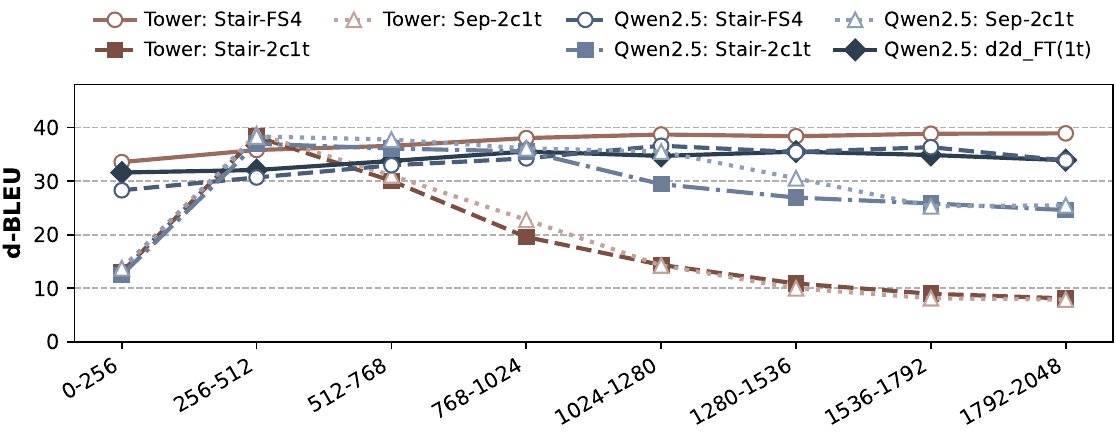}
    \caption{d-BLEU results on IWSLT2017 en–xx with increasing FRC length intervals.}
  \end{subfigure}

  \vspace{0.2cm} 

  \begin{subfigure}{\linewidth}
    \centering
    \includegraphics[width=0.9\linewidth]{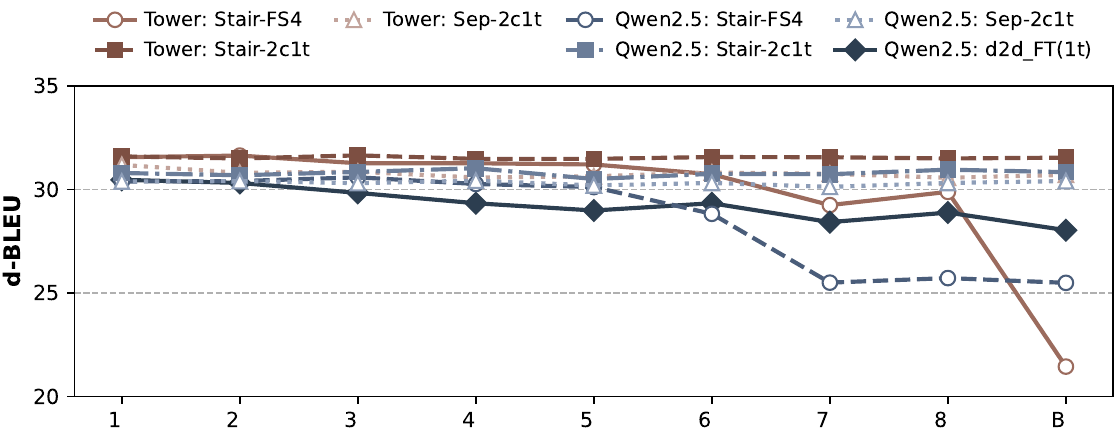}
    \caption{d-BLEU results on BWB zh–en under progressive connection of adjacent chapters.}
  \end{subfigure}

  \caption{Test-time analysis of length variation.}
  \label{fig:tta_length}
  \vspace{-1em}
\end{figure}

As shown in Figure~\ref{fig:tta_length}, 
we perform a two-type length-centric analysis. 
First, we investigate test-time performance across varying FRC length intervals.
Second, we increase test data lengths by concatenating adjacent chapters 
from the same book in the BWB dataset until full books are formed.
Evaluation is conducted at the book level by concatenating all translated chapters for each of the six books.

For the \textsc{Stair}-FS4 and \textsc{d2dFT},
d-BLEU scales positively as the FRC increases.
However, when the length of the test data itself exceeds a certain threshold, 
performance begins to degrade. 
This indicates that while both models generalize effectively within their operational range, 
their capacity is strictly bounded by the training data distribution.

Both the \textsc{Stair}-2c1t and \textsc{Sep}-2c1t models achieve peak accuracy only when the test FRC aligns with the training range~($[256,512]$). 
Meanwhile, \textsc{Tower}-based models exhibit higher sensitivity to FRC than \textsc{Qwen2.5}-based models.
In summary, as long as the FRC is set to match the training configuration,
variations in test data length have little impact on their performance.

\subsection{Impact of Unit Selection on FRC}
\label{sec:unit_exp}

Beyond the baseline approach based on line breaks, 
we also adopt embedding-based semantic chunking and Meta-chunking~\citep{zhao2025metachunkinglearningtextsegmentation}.
For the former, we use the mE5-large model~\citep{wang2024multilinguale5textembeddings} to compute embeddings for all sentences, 
measure the similarity between each pair of adjacent sentences, 
and retain only the top 20\% of connection points with the highest similarity. 
For the latter, we directly employ the Meta-chunking method, 
which identifies appropriate breakpoints based on perplexity,
using the original \textsc{Tower-7B-Instruct-Mistral} model for perplexity~(PPL) computation.
These generated units are subsequently used to construct fixed-range chunks.
The experimental results are summarized in Table~\ref{tab:unit}.

\begin{table}[h!]
\centering
\small
\renewcommand{\arraystretch}{1.2} 
\begin{adjustbox}{width=\columnwidth}
\begin{tabular}{llcccccc}
\toprule
\multirow{2}{*}{\bf Method} &
\multirow{2}{*}{\bf Unit} &
\multicolumn{2}{c}{\bf IWSLT2017} & 
\multicolumn{2}{c}{\bf BWB} &
\multicolumn{2}{c}{\bf GlobVDoc} \\
\cmidrule(lr){3-4}\cmidrule(lr){5-6}\cmidrule(lr){7-8}
& & 
\makecell{\bf d-B.} & \makecell{\bf d-C.} &
\makecell{\bf d-B.} & \makecell{\bf d-C.} &
\makecell{\bf d-B.} & \makecell{\bf d-C.}   \\
\midrule
\addlinespace[1pt]
\multirow{3}{*}{\bf \textsc{Stair}-FS4}
& line-break & 35.81 & 82.29 & 27.61 & 80.56 & 41.96 & 88.31 \\
& emb-sim    & \textcolor{gold}{-0.08} & \textbf{\textcolor{gold}{-0.33}} & \textcolor{gold}{-0.13} & \textcolor{gold}{-0.05} & \textcolor{gold}{-0.07} & +0.01 \\
& meta-ppl   & \textbf{\textcolor{gold}{-0.65}} & \textbf{\textcolor{gold}{-0.35}} & \textcolor{gold}{-0.17} & \textcolor{gold}{-0.13} & \textbf{+0.31} & +0.01 \\
\hdashline
\multirow{3}{*}{\bf \textsc{Stair}-2c1t}
& line-break & 38.29 & 84.14 & 27.53 & 81.10 & 42.06 & 89.35 \\
& emb-sim    & \textcolor{gold}{-0.09} & \textbf{\textcolor{gold}{-0.85}} & \textbf{+0.30} & +0.11 & +0.07 & \textcolor{gold}{-0.01} \\
& meta-ppl   & \textbf{+0.33} & \textbf{\textcolor{gold}{-0.59}} & \textcolor{gold}{-0.12} & +0.09 & \textcolor{gold}{-0.08} & +0.02   \\
\hdashline
\multirow{3}{*}{\bf \textsc{Sep}-2c1t}
& line-break & 38.78 & 84.80 & 26.82 & 81.03 & 41.98 & 89.26 \\
& emb-sim    & \textbf{+0.44} & \textcolor{gold}{-0.02} & +0.14 & \textcolor{gold}{-0.08} & +0.05 & +0.01 \\
& meta-ppl   & \textbf{+0.56} & +0.04 & +0.01 & \textcolor{gold}{-0.10} & \textcolor{gold}{-0.10} & +0.06 \\
\bottomrule
\end{tabular}
\end{adjustbox}
\caption{The results obtained with different compositions of FRC units.
The translation directions for the three datasets are IWSLT2017 en-xx,
BWB zh-en, and GlobVDoc en-xx.
Increases and \textcolor{gold}{decreases} are calculated relative to the ``line-break'' within each respective model.
Significant changes in performance are emphasized in \textbf{bold}.
All the three models are \textsc{Tower}-based.}
\label{tab:unit}
\end{table}

Compared to the BWB and GlobVDoc datasets, 
unit selection induces significantly greater perturbations on IWSLT en-xx. 
This effect is predominantly detrimental to both \textsc{Stair} models, whereas \textsc{Sep}-2c1t exhibits a slight improvement in d-BLEU. 
To elucidate the underlying causes, 
we further calculated the ds-BLEU scores,
n-gram repetition rate~(NRR),
and the proportion of problematic document lengths within the inference results
for these three models on IWSLT2017 en-xx, as shown in Table~\ref{tab:unit_ana}.

\begin{table}[h!]
\centering
\small
\renewcommand{\arraystretch}{1.2} 
\begin{adjustbox}{width=\columnwidth}
\begin{tabular}{llccc}
\toprule
\multirow{2}{*}{\bf Method} &
\multirow{2}{*}{\bf Unit} &
\multicolumn{3}{c}{\bf IWSLT2017} \\
\cmidrule(lr){3-5}
& & 
\makecell{\bf ds-B.} &
\makecell{\bf NRR$\downarrow$} & 
\makecell{\bf Prop.$\downarrow$}  \\
\midrule
\addlinespace[1pt]
\multirow{3}{*}{\bf \textsc{Stair}-FS4}
& line-break & 35.32 & 9.89\%  & 28.25 \%\\
& emb-sim    & 35.22 & 10.70\% & 28.62 \%\\
& meta-ppl   & 34.41 & 14.17\% & 36.94 \%\\
\hdashline
\multirow{3}{*}{\bf \textsc{Stair}-2c1t}
& line-break & 36.40 & 9.09\%  & 16.37\% \\
& emb-sim    & 36.33 & 10.43\% & 18.79\% \\
& meta-ppl   & 36.67 & 8.02\%  & 14.16\%  \\
\hdashline
\multirow{3}{*}{\bf \textsc{Sep}-2c1t}
& line-break & 37.54 & 1.87\% & 4.56\% \\
& emb-sim    & 37.53 & 1.60\% & 3.45\%\\
& meta-ppl   & 37.57 & 1.87\% & 4.03\% \\
\bottomrule
\end{tabular}
\end{adjustbox}
\caption{Further investigation and analysis of unit selecetion results for IWSLT en-xx.
NRR denotes the n-gram repetition rate, 
and Prop. represents the proportion of problematic document lengths. }
\label{tab:unit_ana}
\vspace{-1em}
\end{table}

A comparison of the three models reveals that \textsc{Sep}-2c1t exhibits the most stable decoding performance, 
characterized by the lowest NRR.
In addition, this stability remains largely invariant to the FRC unit selection strategy.
In contrast, both \textsc{Stair} models demonstrate clear volatility in decoding results when unit configurations are varied.
According to the findings of \citet{jin2024chaptertochaptercontextawareliterarytranslation},
n-gram repetition tends to emerge when the input sequence length exceeds 1k tokens. 
However, while the inputs for \textsc{Sep}-2c1t~(including context) typically surpass this 1k threshold, 
they exhibit a relatively low NRR.
In contrast, \textsc{Stair}-2c1t shows obviously higher NRR levels. 
This discrepancy can likely be attributed to the training phase:
\textsc{(Sep)}-2c1t was trained on a single length distribution, 
whereas \textsc{Stair}-2c1t was exposed to three.
In addition, \textsc{Stair}-FS4 incorporated even greater distributional diversity, 
with input lengths substantially exceeding those of the other two models.

Comparing the results in Table~\ref{tab:unit} and Table~\ref{tab:unit_ana} indicates that increases in NRR and the proportion of problematic document lengths correlate strongly with declines in d-BLEU and vise versa.
While the NRR for \textsc{Sep}-2c1t remains relatively consistent across configurations,
its d-BLEU improvements appear more salient because the other two strategies yield a smaller proportion of problematic document lengths compared to the line-break baseline. 
However, this trend is less pronounced in ds-BLEU, which utilizes averaged weights.
The negligible variations in d-BLEU observed for the three models on the BWB and GlobVDoc datasets can be attributed to the fact that their NRR remains consistently at zero across all three unit selection settings.

\begin{figure}[t]
    \centering
    \includegraphics[width=\columnwidth]{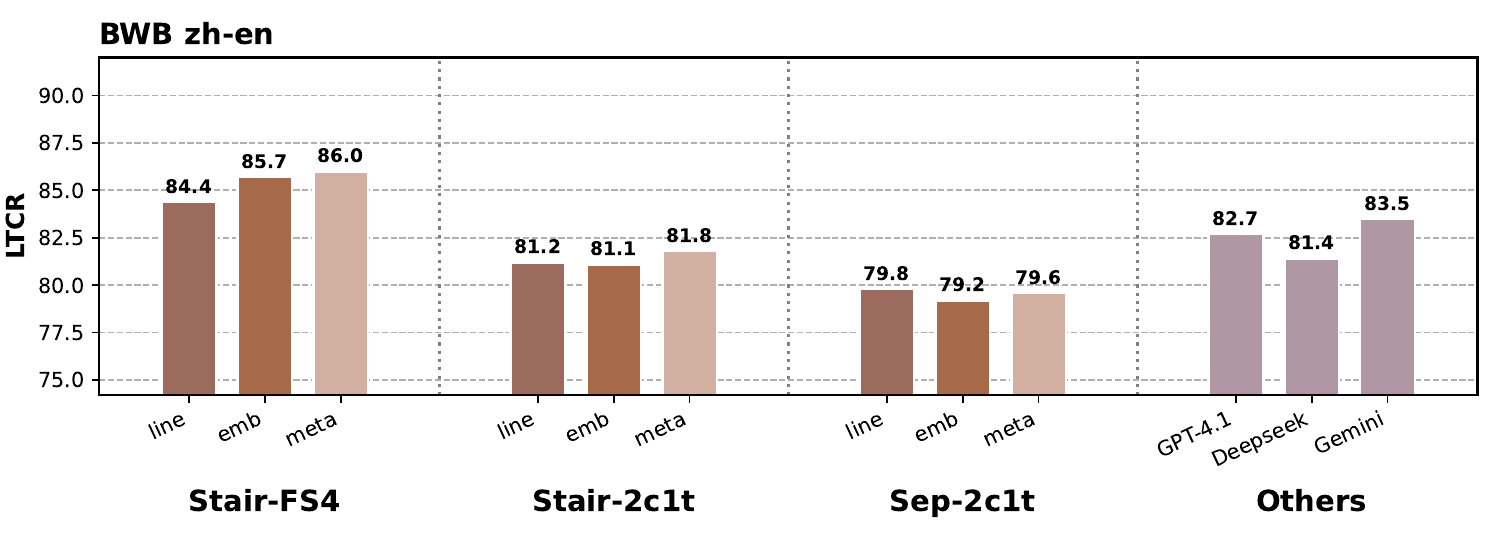}
    \caption{LTCR results of different unit selections across three models on BWB zh-en.}
    \label{fig:unit_ltcr}
    \vspace{-1em}
\end{figure}

In summary, utilizing various unit configurations for FRC fails to provide a stable boost in global accuracy.
One explanation is that the structural significance of the units becomes generalized during the formation of fixed-range chunks.
Additionally, d-BLEU and d-COMET may be insensitive to improvements in discourse coherence within each unit. 
Consequently, we conducted an LTCR analysis on the in-domain BWB dataset~(NRR=0) to better differentiate the three methods. 
The results are presented in Figure~\ref{fig:unit_ltcr}.

Regarding \textsc{Stair}-FS4 and \textsc{Stair-}2c1t, 
the Meta-chunking approach yields marginal improvements~(+1.6 and +0.6, respectively). 
Both unit strategies provide relatively notable gains for \textsc{Stair}-FS4,
even surpassing the performance of Gemini-2.5-Pro. 
These findings suggest that strategic unit partitioning can effectively enhance overall terminological consistency to a certain extent.

\section{Results of LoRA Fine-Tuning}
\label{sec:lora_results}

Table~\ref{tab:lora_result} presents the results of the \textsc{Tower}-based \textsc{Sep}-2c1t models under both full-parameter fine-tuning and LoRA fine-tuning.
Although LoRA fine-tuning does not achieve the same performance as full-parameter fine-tuning, 
it still provides substantial improvements over the original model. 
Increasing the LoRA rank from $r=16$ to $r=64$ introduces a larger number of trainable parameters; 
however, the performance gains on the in-distribution test sets are relatively limited. 

\begin{table*}[h!]
\centering
\small
\resizebox{\textwidth}{!}{
\begin{tabular}{cccccccccc}
\toprule
\multirow{2}{*}{\bf Models} &
\multirow{2}{*}{\makecell{\bf Decoding\\ \bf Format}} &
\multicolumn{2}{c}{\bf IWSLT2017 en-xx} &
\multicolumn{2}{c}{\bf IWSLT2017 xx-en} &
\multicolumn{2}{c}{\bf BWB zh-en} &
\multicolumn{2}{c}{\bf GlobVDoc en-xx} \\
\cmidrule(lr){3-4}\cmidrule(lr){5-6}\cmidrule(lr){7-8}\cmidrule(lr){9-10}
& & \makecell{\bf d-BLEU} & \makecell{\bf d-COM.}
& \makecell{\bf d-BLEU} & \makecell{\bf d-COM.}
& \makecell{\bf d-BLEU} & \makecell{\bf d-COM.}
& \makecell{\bf d-BLEU} & \makecell{\bf d-COM.} \\
\midrule
\multirow{3}{*}{\bf Orig Model}
& 0c1t & 30.00 & 81.83 & 35.42 & 85.03 & 19.38 & 80.72 & 37.58 & 87.49\\
& 1c1t & 29.80 & 82.05 & 35.29 & 84.94 & 19.07 & 80.47 & 36.45 & 85.87\\
& 2c1t & 30.07 & 82.42 & 35.37 & 84.88 & 18.75 & 80.59 & 37.51 & 86.18\\
\hline
\multirow{3}{*}{\makecell{\bf \textsc{Sep}-2c1t\\ \bf (Full)}}
& 0c1t & \bf 39.61 & 84.69 & 63.05 & 86.09 & \textbf{27.38} & \textbf{81.11} & 41.92 & 89.11\\
& 1c1t & 39.51 & \bf 84.83 & 66.84 & 86.49 & 27.16 & 81.03 & 41.87 & 89.25\\
& 2c1t & 38.78 & 84.80 & \bf 67.20 & \bf 86.53 & 26.82 & 81.03 & 41.98 & 89.26\\
\midrule
\multirow{3}{*}{\makecell{\bf \textsc{Sep}-2c1t\\ \bf (LoRA $r=16$)}} 
& 0c1t & 38.61 & 84.04 & 43.04 & 85.06 & 25.99 & 80.42 & 42.78 & 89.02\\
& 1c1t & 38.65 & 84.25 & 43.56 & 86.15 & 25.96 & 80.52 & \textbf{43.28} & \textbf{89.52}\\
& 2c1t & 37.85 & 84.26 & 43.88 & 86.30 & 25.74 & 80.52 & 42.76 & 89.39\\
\midrule
\multirow{3}{*}{\makecell{\bf \textsc{Sep}-2c1t\\ \bf (LoRA $r=64$)}} 
& 0c1t & 38.26 & 83.94 & 43.70 & 85.17 & 26.36 & 80.61 & 41.47 & 88.23\\
& 1c1t & 38.81 & 84.28 & 44.43 & 86.35 & 26.21 & 80.17 & 42.03 & 88.77\\
& 2c1t & 38.85 & 84.29 & 44.68 & 86.38 & 26.51 & 80.64 & 41.78 & 88.69\\
\bottomrule
\end{tabular}
}
\caption{
Comparison among original, full-parameter fine-tuned \textsc{Sep}-2c1t and LoRA fine-tuned \textsc{Sep}-2c1t models.
For each evaluation metric, the best performance is shown in \textbf{bold}.
}
\label{tab:lora_result}
\vspace{-1em}
\end{table*}

\section{AI Assistance Usage}
We used AI-assisted tools,
including ChatGPT\footnote{\url{https://chatgpt.com}}, 
Google Gemini\footnote{\url{https://gemini.google.com}}, 
and Claude\footnote{\url{https://claude.ai}},
to support code development and writing refinement (in accordance with ACL Policy on AI Writing Assistance).

\begin{figure*}[htbp]
    \centering
    \includegraphics[width=1.0\textwidth]{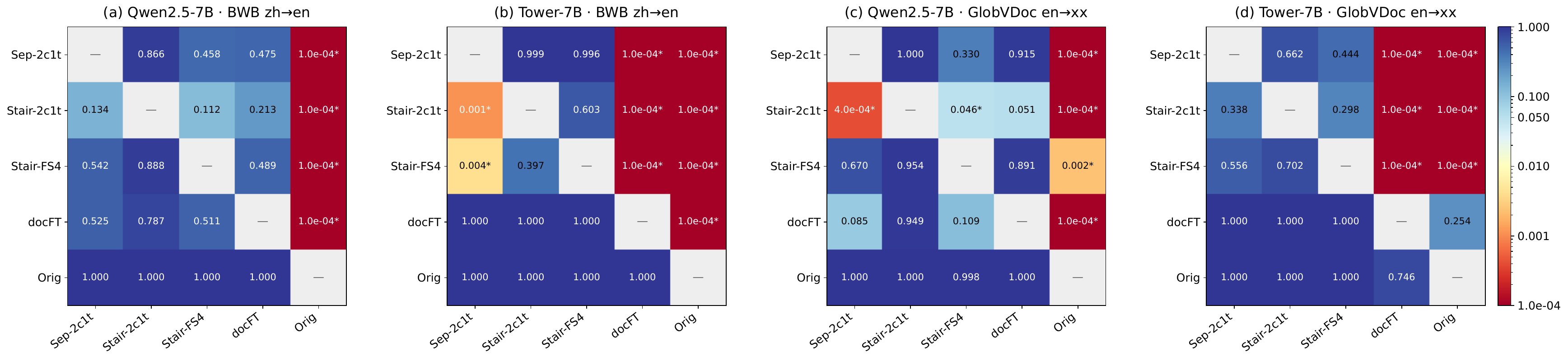}
    \caption{Pairwise one-sided significance tests on BWB zh-en and GlobVDoc en-xx for \textsc{Qwen2.5}-7B and \textsc{Tower}-7B.
    Each cell reports the paired document-bootstrap p-value for the hypothesis that the row system outperforms the column system in d-BLEU.
    Asterisks indicate \(p<0.05\).}
    \label{fig:bwb_globv_stat}
    \vspace{-1em}
\end{figure*}

\section{Prompts}
\label{sec:prompts}
\begin{tcolorbox}[
  colback=white,
  colframe=black,
  title=Prompt: D2D and 0c1t,
  colbacktitle=gray!20,
  coltitle=black,
  fonttitle=\bfseries,
  rounded corners,
  boxrule=0.6pt
]
\textcolor{blue}{<|im\_start|>user}

Translate the following source text from \{source\_lang\} into \{target\_lang\}.

\{source\_lang\}: \{source\_text\}.

\{target\_lang\}: \textcolor{blue}{<|im\_end|>}

\textcolor{blue}{<|im\_start|>assistant}

\{target\_text\}.\textcolor{blue}{<|im\_end|>}

\end{tcolorbox}

\begin{tcolorbox}[
  colback=white,
  colframe=black,
  title=Prompt: 1c1t,
  colbacktitle=gray!20,
  coltitle=black,
  fonttitle=\bfseries,
  rounded corners,
  boxrule=0.6pt
]
\textcolor{blue}{<|im\_start|>user}

Context:

\{source\_lang\}: \{source\_context\}

Translate the following source text from \{source\_lang\} into \{target\_lang\}.

\{source\_lang\}: \{source\_text\}.

\{target\_lang\}: \textcolor{blue}{<|im\_end|>}

\textcolor{blue}{<|im\_start|>assistant}

\{target\_text\}.\textcolor{blue}{<|im\_end|>}

\end{tcolorbox}

\begin{tcolorbox}[
  colback=white,
  colframe=black,
  title=Prompt: 2c1t,
  colbacktitle=gray!20,
  coltitle=black,
  fonttitle=\bfseries,
  rounded corners,
  boxrule=0.6pt
]
\textcolor{blue}{<|im\_start|>user}

Context1:

\{source\_lang\}: \{source\_context1\}

Context2:

\{source\_lang\}: \{source\_context2\}

Translate the following source text from \{source\_lang\} into \{target\_lang\}.

\{source\_lang\}: \{source\_text\}.

\{target\_lang\}: \textcolor{blue}{<|im\_end|>}

\textcolor{blue}{<|im\_start|>assistant}

\{target\_text\}.\textcolor{blue}{<|im\_end|>}

\end{tcolorbox}

\begin{tcolorbox}[
  colback=white,
  colframe=black,
  title=Prompt: FS4,
  colbacktitle=gray!20,
  coltitle=black,
  fonttitle=\bfseries,
  rounded corners,
  boxrule=0.6pt
]
\textcolor{blue}{<|im\_start|>user}

Context1:

\{source\_lang\}: \{source\_context1\}

Context2:

\{source\_lang\}: \{source\_context2\}

Context3:

\{source\_lang\}: \{source\_context3\}

Translate the following source text from \{source\_lang\} into \{target\_lang\}.

\{source\_lang\}: \{source\_text\}.

\{target\_lang\}: \textcolor{blue}{<|im\_end|>}

\textcolor{blue}{<|im\_start|>assistant}

\{target\_text\}.\textcolor{blue}{<|im\_end|>}

\end{tcolorbox}

\begin{tcolorbox}[
  colback=white,
  colframe=black,
  title=Gemba-DA,
  colbacktitle=gray!20,
  coltitle=black,
  fonttitle=\bfseries,
  rounded corners,
  boxrule=0.6pt
]

Score the following translation from \{src\_lang\} to \{tgt\_lang\} with respect to human reference on a continuous scale 0 to 100 where score of zero means "no meaning preserved" and score of one hundred means "perfect meaning and grammar".\\

\{src\_lang\} source: \{src\_text\}

\{tgt\_lang\} human reference: \{ref\_text\}

\{tgt\_lang\} machine translation: \{tgt\_text\}\\

Without any thinking, output only JSON in the following format: \{"score": <int>\}

\end{tcolorbox}

\section{Algorithm}

\begin{algorithm}[h]
\caption{Dual-Boundary Matching based Chunk Alignment}
\label{alg:chunk_align}

\SetKwInOut{KwIn}{Input}
\SetKwInOut{KwOut}{Output}
\SetKwFunction{FScore}{SimwRP}
\SetKwProg{Fn}{Function}{:}{}

\KwIn{Source chunks $\mathcal{C}$, Target units $\mathcal{T}$, $\lambda$, $\sigma$, Initial $k$, $\Delta k$}
\KwOut{Aligned lower bound indices $J = \{j_1, \dots, j_{|\mathcal{C}|}\}$, $j \in \{1,..,|\mathcal{T}|\}$}

\BlankLine

\Fn{\FScore{$u, v$}}{
    $s \gets \text{Sim}(u, v)$\;
    $w \gets \exp \left(- \left(\frac{\text{rp}(u) - \text{rp}(v)}{\sigma}\right)^2 \right)$\;
    \Return $s \times ((1-\lambda) + \lambda w)$\;
}

\BlankLine

Initialize score matrix $\Phi \in \mathbb{R}^{|\mathcal{C}| \times |\mathcal{T}|}$\;

\While{$J = \emptyset$}{
\For{$i \gets 1$ \KwTo $|\mathcal{C}|$}{
    \For{$j \gets 1$ \KwTo $|\mathcal{T}|$}{
        $\Phi[i, j] \gets \FScore(c_i^{\text{end}}, t_j)$\;
        Find top-$k$ candidates $\{j_{i,1},...,j_{i,k}\}$\;
            \For{$r \gets 1$ \KwTo $k$}{
            $\Phi[i, j] \gets \Phi[i, j] + \FScore(c_{i+1}^{\text{first}}, t_{j_{i,r}+1})$\;
            }
    }
}

Find $j_1, \dots, j_{|\mathcal{C}|}$ maximizing $\sum \Phi[i, j_i]$\;
\quad subject to $j_{i-1} \le j_i$ (Monotonicity)\;
\quad and $j_{|\mathcal{C}|} = |\mathcal{T}|$ (Boundary Condition)\;

}
\lIf{$J = \emptyset$}{$k \gets \min(k + \Delta k, |\mathcal{T}|)$}
\Return{$J$ \textbf{or} Failure}
\end{algorithm}

\begin{algorithm*}[h]
\caption{Two-Stage Dynamic Programming Algorithm for Fixed-Range Chunking}
\label{alg:detailed_frc}
\SetKwInput{Input}{Input}
\SetKwInput{Output}{Output}
\SetKwFunction{Len}{Len}

\Input{
    Source units $S = \{s_1, s_2, \dots, s_n\}$; \\
    Target length $L^* = (m+M)/2$ derived from $[m, M]$; \\
    Separator cost $\delta$; \\
    Max allowed violations $K$.
}
\Output{Packed segments $\mathcal{P}$.}

\BlankLine
Calculate $|s_i|$ for all $s_i \in S$\;
Initialize $\mathcal{P} \leftarrow \emptyset$\;
\tcc{Pre-processing: Isolate super-long sentences}
Divide $S$ into blocks $\{B_1, B_2, \dots\}$ by splitting at any $s_i$ where $|s_i| > M$\;

\BlankLine
\ForEach{block $B \in \{B_1, B_2, \dots\}$ with units $u_1 \dots u_{T}$}{
    \If{$|B| = 1$ \textbf{and} $|u_1| > M$}{
        Add $\{u_1\}$ to $\mathcal{P}$; \textbf{continue}\;
    }
    
    \tcc{Define length function with separator cost}
    Let $Len(i, j) = \sum_{k=i}^{j} |u_k| + (j-i) \cdot \delta$\;
    
    \BlankLine
    \tcc{Stage 1: Strict Constraint DP (Attempt to keep all segments $\in [m, M]$)}
    Initialize $dp[0] \leftarrow 0$, $dp[1 \dots T] \leftarrow \infty$\;
    \For{$i \leftarrow 1$ \KwTo $T$}{
        \For{$j \leftarrow i-1$ \KwTo $0$}{
            $l \leftarrow Len(j+1, i)$\;
            \If{$m \le l \le M$}{
                $dp[i] \leftarrow \min(dp[i], dp[j] + (l - L^*)^2)$\;
            }
        }
    }
    
    \BlankLine
    \If{$dp[T] < \infty$}{
        Backtrack $dp$ to recover partition, add to $\mathcal{P}$\;
    }
    \Else{
        \tcc{Stage 2: Relaxed DP with Violated Segment Constraints}
        Let $dp_{rel}[v][i]$ be min cost for first $i$ units with $v$ violations\;
        Initialize $dp_{rel}[0 \dots K][0] \leftarrow 0$, others $\infty$\;
        
        \For{$i \leftarrow 1$ \KwTo $T$}{
            \For{$j \leftarrow i-1$ \KwTo $0$}{
                $l \leftarrow Len(j+1, i)$\;
                \eIf{$m \le l \le M$}{
                    \For{$v \leftarrow 0$ \KwTo $K$}{
                         $dp_{rel}[v][i] \leftarrow \min(dp_{rel}[v][i], dp_{rel}[v][j] + (l - L^*)^2)$\;
                    }
                }{
                    $C_{out} \leftarrow (m-l)^2$ if $l < m$ else $(l-M)^2$\;
                    \For{$v \leftarrow 1$ \KwTo $K$}{
                         $dp_{rel}[v][i] \leftarrow \min(dp_{rel}[v][i], dp_{rel}[v-1][j] + C_{out})$\;
                    }
                }
            }
        }
        Find $v^*$ minimizing $dp_{rel}[v][T]$ for $v \in [1, K]$\;
        Backtrack $dp_{rel}[v^*]$ to recover partition, add to $\mathcal{P}$\;
    }
}

\Return{$\mathcal{P}$ ordered by original unit indices}\;

\end{algorithm*}

\begin{CJK}{UTF8}{mj}
\onecolumn
\renewcommand{\arraystretch}{1.05}
\setlength{\LTpre}{0pt}
\setlength{\LTpost}{0pt}
\small
\begin{longtable}{r p{3.8cm} c p{4.2cm} c p{2.2cm} p{2.2cm}}
\toprule
ID & Source Title & Src & Translated Title & Tgt & Authors & Translators \\
\midrule
\endfirsthead
\toprule
ID & Source Title & Src & Translated Title & Tgt & Authors & Translators \\
\midrule
\endhead
\midrule
\multicolumn{7}{r}{\small Continued on next page} \\
\endfoot
\endlastfoot
0 & The pros and cons of Chinese investment in Tajikistan's gold mining sector & En & \zh{发展与环保的两难：塔吉克斯坦金矿开采业的中国投资} & Zh & Shahida Yakub & Global Climate Justice Fellowship \\
1 & The complex role China plays in Africa's energy transition & En & \zh{中国在非洲能源转型中，扮演何种角色？} & Zh & Ruohan Xie, Desire Nimubona & Sicong Zou \\
2 & Boycotting Xinjiang cotton: What does it mean for environmental and labor justice in Central Asia? & En & \zh{抵制新疆棉，对中亚的环境和劳工权利影响几何？} & Zh & Shahida Yakub, Global Climate Justice Fellowship & Global Climate Justice Fellowship \\
3 & As electric vehicles gain momentum in Brazil, China's influence shines through & En & \zh{巴西电动汽车蓬勃发展背后的中国影响力} & Zh & Laís Martins, Luo Jieqi & Sicong Zou \\
4 & China increases gas imports from Turkmenistan for green energy transition. It's impact is unclear & En & \zh{中国增加土库曼斯坦天然气进口，绿色转型与甲烷泄露矛盾突显} & Zh & Global Climate Justice Fellowship, Shahida Yakub & Global Climate Justice Fellowship \\
5 & Will Ecuador lift Amazon oil block despite a historic referendum? & En & \zh{历史性公投后，厄瓜多尔会停止在亚马逊地区开采新油田吗？} & Zh & Gabriela Mesones Rojo, Alicia Chen & Global Climate Justice Fellowship \\
6 & With the reintroduction of import taxes on Chinese solar panels, Brazil hopes to develop its own industry & En & \zh{对中国光伏板重新征税，巴西希望发展本土产业} & Zh & Laís Martins, Luo Jieqi & Luo Jieqi \\
7 & Is China partly responsible for the destruction of Africa's Miombo woodlands? & En & \zh{非洲米欧波林地遭破坏，中国是否需要承担责任？} & Zh & Ruohan Xie, Desire Nimubona & Ruohan \\
8 & China helped Cameroon build drinking water infrastructure. Is it a debt crisis or developmental aid? & En & \zh{中国助喀麦隆饮建设用水基础设施：是援助发展，还是债务危机？} & Zh & Desire Nimubona, Ruohan Xie & Sicong Zou \\
9 & In Guadeloupe, ‘Creole gardens’ give climate lessons in a spirit of solidarity & En & Em Guadalupe, as “hortas crioulas” ensinam sobre o clima em um ambiente solidário & Pt & Olivia Losbar & Rogério de Sá \\
10 & May Day or Detention Day? Turkey marks Labor Day & En & Primeiro de Maio ou Dia da Detenção? A Turquia celebra o Dia do Trabalho & Pt & Arzu Geybullayeva & Ronaldo Santos \\
11 & In Russia, dozens of volunteers are working to save bat populations & En & Na Rússia, dezenas de voluntários trabalham para salvar populações de morcegos & Pt & Daria Dergacheva & Isabela Torezan \\
12 & When the ICE agent at the airport echoes Trump’s motto: A Brazilian journalist at the US border & En & Quando o agente da ICE no aeroporto ecoa o slogan de Trump: uma jornalista brasileira na fronteira dos EUA & Pt & Agencia Publica & Pública - Agência de jornalismo investigativo \\
13 & Saydnaya: In Syria, a legacy of pain looking for an honorable cleansing & En & Sednaya: na Síria, um legado de dor em busca de purificação honrosa & Pt & Rami Alhames & Denise Andréia Stange \\
14 & ‘Let’s talk about something else’: China’s AI chatbot DeepSeek censors sensitive topics & En & “Vamos mudar de assunto”: chatbot de IA chinês DeepSeek censura temas sensíveis & Pt & Hong Kong Free Press & Joao Pedro Ribeiro \\
15 & AI’s bitter truth: It has biases, too & En & A amarga verdade da IA: ela não é neutra e pode falhar com seus vieses & Pt & Tactical Tech & Fernando Baumgarten \\
16 & The Golem: A mythical protector inspiring US comics but also a metaphor for AI & En & O Golem: Um protetor mítico que inspira quadrinhos dos EUA, mas também uma metáfora para a IA & Pt & Fred Petrossian & Fernando Baumgarten \\
17 & The Svrzo House: A window into the past of everyday life in Sarajevo & En & A Casa Svrzo: Uma janela ao passado da vida diária em Sarajevo & Pt & Balkan Diskurs & Gabriel Ferreira Wittaker \\
18 & ‘Embracing imperfection is key to artistic evolution': An interview with Iranian artist Sadegh Adham & En & ‘Aceitar a imperfeição é a chave para a evolução artística': Entrevista com o artista iraniano Sadegh Adham & Pt & Omid Memarian & Isabela Torezan \\
19 & The Conscience under attack: A story of aid, silence, and starvation & En & Das Gewissen unter Beschuss: Eine Geschichte von Hilfe, Schweigen und Hungersnot & De & Walid El Houri & GV Deutsch \\
20 & A new aid regime for Gaza: Humanitarian facade, military core & En & Ein neues Hilfssystem für Gaza: Humanitäre Fassade, militärischer Kern & De & Saher & GV Deutsch \\
21 & The murder of a young girl in Ethiopia reveals TikTok’s content moderation failures & En & Die Ermordung eines jungen Mädchens in Äthiopien offenbart TikToks Versagen bei der Inhaltsmoderation & De & Endalkachew Chala & FTSK \\
22 & How climate change is affecting farmers in Tobago & En & Wie der Klimawandel die Landwirt*innen in Tobago beeinflusst & De & Cari-Bois News & FTSK \\
23 & ‘I don’t feel safe': Reactions to Germany’s suppression of pro-Palestine solidarity & En & „Ich fühle mich nicht sicher“ – Reaktionen auf Deutschlands Unterdrückung der pro-Palästina Bewegung & De & Safa & Salma \\
24 & Heatwave highlights climate vulnerabilities in Southeast Asia & En & Hitzewelle macht Klimaanfälligkeit Südostasiens deutlich & De & Sydney Allen & Mirjam Stiegel \\
25 & 43 years in Syria's prisons for refusing to bomb a city & En & 43 Jahre in Syriens Gefängnissen, weil er sich weigerte, eine Stadt zu bombardieren & De & Rami Alhames & Anne Hemeda \\
26 & Nepal’s journey to electric public transport & En & Nepals Weg hin zum elektrischen Nahverkehr & De & Nepali Times & Julia K., FTSK \\
27 & Blood and democracy: the fight for animal rights in Azerbaijan & En & Der Kampf für Tierrechte und Demokratie in Aserbaidschan & De & Arzu Geybullayeva, OC Media & Mirjam Stiegel \\
28 & Why I might not go back to El Salvador & En & Warum ich vielleicht nie wieder nach El Salvador zurückkehre & De & Melissa Vida & Monica Seemann \\
29 & How the Caribbean views the Trump administration's mass deportations & En & Caraïbes : le prisme des déportations massives de l'administration Trump & Fr & Janine Mendes-Franco, Candice Stewart & Corinne Molton \\
30 & Chinese social media users call this age “The Garbage Time of History” & En & Chine : les utilisateurs des réseaux sociaux appellent cette ère « L'âge des ordures » & Fr & Oiwan Lam & Rodrigue Macao \\
31 & World Environment Day: In Jamaica, the battle against plastics continues & En & Journée mondiale de l'environnement : en Jamaïque, la lutte contre les déchets plastiques se poursuit & Fr & Emma Lewis & Corinne Molton \\
32 & Across war zones, targeting healthcare has become a strategy, not an accident & En & Les attaques délibérées contre les systèmes de santé deviennent une caractéristique de la guerre moderne - et un test pour le droit international. & Fr & Walid El Houri & Helene Senecot \\
33 & This Ghanaian software engineer is working to raise women's financial literacy & En & Un ingénieur logiciel ghanéen à pied d'œuvre pour améliorer l'éducation financière des femmes africaines & Fr & Zita Zage & Fierté Mbadzi \\
34 & Monoliths of Nartiang: The remnants of India's Jaintia tribal kingdom through photos & En & Monolithes de Nartiang : les vestiges du royaume tribal indien de Jaintia en images & Fr & Arpita Das Choudhury & Corinne Molton \\
35 & Power, myth, and the personal: A conversation with Iranian-American artist Shiva Ahmadi & En & Pouvoir, mythe et personnalité : entretien avec l'artiste irano-américaine Shiva Ahmadi & Fr & Omid Memarian & Pierre-Emmanuel Farret \\
36 & Integrating Indigenous peoples and local communities (IPLCs) in Nepal's conservation efforts & En & Népal : intégrer les populations autochtones et les communautés locales (IPLC) dans les efforts de conservation du pays & Fr & Biswash Chepang & Pierre-Emmanuel Farret \\
37 & Tobago's coral reefs brace for ‘imminent threat’ & En & Les récifs coralliens de Tobago se préparent à une « menace imminente » & Fr & Janine Mendes-Franco & Stanislav Kibalnyk \\
38 & Redefining belonging: From statelessness to collective strength & En & Redéfinir l'appartenance : de l'apatridie à la force collective & Fr & Guest Contributor & Ridley Mortensen \\
39 & China’s warning against cross-border marriage scams reveals the pitfalls of human trafficking & En & Advertencia de China contra las estafas matrimoniales transfronterizas revela trampas de la trata de personas & Es & Rezwan, Oiwan Lam & Karen Lopez \\
40 & El Salvador's soaring property costs are hurting locals & En & Elevados precios de viviendas en El Salvador perjudican a lugareños & Es & Eddie Galdamez & Valeria Malavolta \\
41 & Podcast: Daria on carrying a Russian identity and setting her children free of it & En & Podcast: Daria sobre la carga de identidad rusa y su decisión de criar hijas sin esa carga & Es & Akwe Amosu,  Daria Dergacheva & Georgina Loscalzo \\
42 & No lambs for Eid: Drought, deforestation, and decline in Morocco & En & No hay corderos para el Eid: Sequía, deforestación y deterioro en Marruecos & Es & Mohamed Belkasen & Eva González \\
43 & Painting as a shelter: Spanish artist Bárbara Alegre on art, grief, and emotional healing & En & La pintura como refugio: Artista española Bárbara Alegre sobre el arte, el duelo y la sanación emocional & Es & Omid Memarian & Mariela Arnst \\
44 & Kenya outlawed the sharing of Indigenous seeds. A village choir is fighting back & En & Kenia prohibió intercambiar semillas autóctonas. Un coro comunitario contraataca & Es & Minority Africa & Mariela Arnst \\
45 & Ahead of municipal elections, Lebanese voter data is at risk & En & Antes de elecciones municipales, datos de los votantes libaneses corren peligro & Es & SMEX & Gabriela García Calderón Orbe \\
46 & The Congo Basin, a vital home for global biodiversity is at risk & En & Cuenca del Congo, vital para la biodiversidad global, está en peligro & Es & Guest Contributor & Mariela Arnst \\
47 & How the California gold rush continues to shape Africa and other global majority countries & En & Cómo la fiebre del oro en California sigue dando forma a África y a otros países de la mayoría global & Es & Abdallah Abdallah & Mariela Arnst \\
48 & Gains and gaps in gender equality in North Macedonia & En & Avances y brechas de igualdad de género en Macedonia del Norte & Es & Kristina Hadzi-Vasileva & Mariela Arnst \\
49 & In Turkey, a controversial law on cybersecurity is widely seen as yet another censorship tool & En & In Turkije wordt een omstreden wet over cyberbeveiliging gezien als een nieuw middel tot censuur & Nl & Arzu Geybullayeva & Mijke Luttikholt \\
50 & Bringing ‘Pateh’ to the world: Sara Qashghai’s artistic reinterpretation of Iranian needlework & En & ‘Pateh’ naar de wereld toe brengen: Sara Qashghai’s artistieke herinterpretatie van Iraans handwerk & Nl & Omid Memarian & Els Whittle \\
51 & Indigenous People defend traditional farming in northern Thailand & En & In Noord-Thailand zet de de bevolking zich in voor traditionele landbouwmethoden & Nl & Prachatai & Eric Crena Uiterwijk \\
52 & From Myanmar to Thailand: Displaced journalists tell their stories & En & Van Myanmar tot Thailand: Ontheemde journalisten vertellen hun verhaal & Nl & Prachatai & Eric Crena Uiterwijk \\
53 & Iran sees 80\% spike in executions two years after protests & En & Iran: 80\% meer executies twee jaar na protesten & Nl & Iran Open Data Center & Thomas De Backer \\
54 & Women are paying the ultimate price in Cameroon’s armed conflict & En & Vrouwen betalen de hoogste prijs in het gewapende conflict van Kameroen & Nl & Minority Africa & Jasmijn Dagevos \\
55 & Pouring concrete on rice fields in Nepal & En & In Nepal veranderen de rijstvelden in een betonnen woestenij & Nl & Nepali Times & Eric Crena Uiterwijk \\
56 & Where are the unusual swarms of bees in Chișinău, Moldova, coming from? & En & Waar komen de ongebruikelijke bijenzwermen uit Chișinău in Moldavïe vandaan? & Nl & Daria Dergacheva & Eric Crena Uiterwijk \\
57 & Cat lovers boost tourism in Taiwan village as feline residents revive once-flourishing mining town & En & Kattenliefhebbers stimuleren toerisme in een eens bloeiend, door kattenbewoners nieuw leven ingeblazen, Taiwanees mijnstadje & Nl & Hong Kong Free Press & Els Whittle \\
58 & When it comes to bullying, small actions can help stop big problems & En & Kleine daden kunnen groot verschil maken tegen pesten & Nl & Guest Contributor & Mijke Luttikholt \\
59 & Koryo-saram: The long and tragic story of Koreans in Russia & En & \ko{고려사람 : 러시아 한인의 길고 비극적인 역사} & Ko & Daria Dergacheva & June Lee \\
60 & Vaping loopholes are endangering children and the environment in Nigeria and Burkina Faso & En & \ko{나이지리아 및 부르키나파소의 어린이 그리고 환경, 전자담배의 허점으로 위험에 처해} & Ko & The Colonist Report & June Lee \\
61 & China strives to go green in South America's ‘Lithium Triangle’ & En & \ko{중국, 남미의 ‘리튬 삼각지대'에서 친환경화를 위해 분투하다} & Ko & Alicia Chen, Gabriela Mesones Rojo & June Lee \\
62 & Hong Kong secondary students may soon be schooled in ‘Xi Jinping Thought’ & En & \ko{홍콩 중학생들, 곧 ‘시진핑 사상’ 교육 받을 수도} & Ko & Hong Kong Free Press & June Lee \\
63 & Coup and resistance in Myanmar: A timeline of the first month under the 2021 military junta & En & \ko{미얀마에서의 쿠데타와 저항 : 2021년 군부 독재 하의 첫 달 타임라인} & Ko & Global Voices South East Asia & Seojin Lim \\
64 & Masculinity in my genes/jeans & En & \ko{청바지(진)에 들어있는 사내다움} & Ko & Amilcar Sanatan & May Cho \\
65 & The Caribbean's case for reparations: Part I & En & \ko{카리브해 지역 배상 문제: 제 1부} & Ko & Janine Mendes-Franco & May Cho \\
66 & The Caribbean's case for reparations: Part II & En & \ko{카리브 해 지역의 배상 문제: 제 2 부} & Ko & Janine Mendes-Franco & May Cho \\
67 & The Caribbean's case for reparations: Part III & En & \ko{카리브 해 지역의 배상 문제: 제 3부} & Ko & Janine Mendes-Franco & May Cho \\
68 & Women ‘don’t have to fit themselves into someone else's perception,’ says Turkish aerospace engineer & En & \ko{터키의 항공 우주 기술자: 여성은 타인의 인식에 맞출 필요가 없어요} & Ko & Sevgi Yagmur Bulut & Keun Jeong \\
69 & From barren land to thriving forest: The story of photographer Sebastião Salgado’s Instituto Terra in Brazil & En & Dalla terra arida alla foresta florida: la storia dell'Instituto Terra del fotografo Sebastião Salgado in Brasile & It & Ana Cavalcanti & Margherita Lista \\
70 & Global digital rights report reveals unexpected boost in transparency from Chinese tech giants & En & Rapporto globale sui diritti digitali rivela un inaspettato aumento della trasparenza da parte dei giganti tecnologici cinesi & It & Marisa Petricca & Marisa Petricca \\
71 & Two years on, Turkey earthquake survivors continue to live in limbo & En & Due anni dopo, i sopravvissuti al terremoto in Turchia continuano a vivere nell'incertezza & It & Arzu Geybullayeva & Martina Cesarano \\
72 & Argentine resistance hinders Milei’s forest and glacier destruction & En & La resistenza argentina ostacola la distruzione di foreste e ghiacciai operata da Milei & It & Climate Home News & Vivi \\
73 & A ballerina battles for European Georgia & En & Una ballerina che lotta per una Georgia europea & It & Giorgi Ninos & Martina Cesarano \\
74 & Women's rights under threat in Uganda as conservative groups push disinformation campaign & En & Diritti delle donne sotto attacco in Uganda mentre gruppi conservatori promuovono una campagna di disinformazione & It & Prudence Nyamishan & Laura Carlevero \\
75 & How China’s investment in Indonesia's nickel industry is impacting local communities & En & Gli investimenti cinesi nell'industria del nichel hanno un'impatto sulle comunità locali in Indonesia & It & Hasya Nindita, Zhaoyin Feng & Barbara Foggiato \\
76 & Pacific communities seek to protect kava as it gains global popularity & En & Le comunità del Pacifico cercano di proteggere la kava, una bevanda la cui popolarità continua a crescere & It & Mong Palatino & Barbara Foggiato \\
77 & On International Women's Day, a call for Jamaica to combat systemic barriers that hold women back & En & Nella Giornata internazionale della donna, la Giamaica deve combattere le barriere sistemiche che frenano le donne & It & Janine Mendes-Franco & Martina Cesarano \\
78 & Money from trees: What of Guyana's Indigenous people and their rights — and do they benefit from the carbon trade? & En & Denaro dagli alberi: cosa rimane agli indigeni della Guyana e i loro diritti? Beneficiano del commercio del carbonio? & It & Guest Contributor & Barbara Foggiato \\
79 & For how long? Aramaic language and its enduring legacy in Syria & En & \foreignlanguage{russian}{Арамейский язык в Сирии: бесценное наследие, которое мы можем потерять?} & Ru & Rami Alhames & \foreignlanguage{russian}{Елена Кузминцева} \\
80 & One man is trying to save a language in Bangladesh with only six native speakers & En & \foreignlanguage{russian}{Одиночка, пытающийся спасти язык, на котором говорят всего шесть человек} & Ru & Pantha & Rezwan, Laura Sabit \\
81 & Peru adopts controversial ‘anti-NGO’ law & En & \foreignlanguage{russian}{Перу принимает спорный закон против НПО} & Ru & IFEX, Laura Vidal & Dalia Tarek, Anastasia Pestova \\
82 & A decade of digital rights: Where has Big Tech's progress gone? & En & \foreignlanguage{russian}{Десятилетие цифровых прав: куда исчез прогресс Big Tech?} & Ru & Giovana Fleck & Elise Patterson \\
83 & A new aid regime for Gaza: Humanitarian facade, military core & En & \foreignlanguage{russian}{Новый режим помощи Газе: гуманитарный фасад, но военная основа} & Ru & Saher & Marina Bovkalo \\
84 & Chinese social media users are outraged by the mysterious death of an internet celebrity cat & En & \foreignlanguage{russian}{Гибель звезды соцсетей: в Китае гадают, что произошло с котом Укуном} & Ru & Oiwan Lam & Elise Patterson \\
85 & In the Llanos Orientales, seduction is another weapon of war & En & \foreignlanguage{russian}{Колумбия: соблазнение как ещё одно оружие войны} & Ru & Mi Historia & Hebe Powell, Anastasia Pestova \\
86 & The environmental impact of Chinese cement plants in Tajikistan remains hidden & En & \foreignlanguage{russian}{Что скрывают китайские цементные заводы в Таджикистане?} & Ru & Nurbek Bekmurzaev, Brian Hioe & Elise Patterson \\
87 & Nelly Gesare turns trash into treasure, highlighting the potential of sustainability-driven businesses in Africa & En & \foreignlanguage{russian}{Как в Африке бывшая журналистка Нелли Джесаре создала процветающий бизнес из отходов} & Ru & Bird & Marina Bovkalo \\
88 & Chechen opposition activist fears deportation to Russia from Kazakhstan & En & \foreignlanguage{russian}{Чеченский оппозиционер опасается депортации в Россию из Казахстана} & Ru & Guest Contributor, Daria Dergacheva & Anastasia Pestova \\
89 & Rebels and Rebellion in Classic Chinese Literature & En & \zh{中国古典文学里的反骨精神与叛逆分子} & Zh & I-fan Lin & Yanne C \\
\bottomrule
\caption{The metadata for the documents within the GlobVDoc dataset includes the document ID, 
source and translated titles, 
source and target languages, 
and the names of authors and translators.}\\
\label{tab:globvdoc_docs}
\end{longtable}
\twocolumn
\end{CJK}

\end{document}